\documentclass[11pt]{article}

\usepackage{amsmath,amsbsy,amsfonts,amssymb,amsthm,dsfont,fullpage,units}
\usepackage{graphicx,psfrag,epsfig,epsf}
\usepackage{algorithm,algorithmic}
\usepackage{mathtools}
\usepackage{color,cases}
\usepackage{tikz}
\usetikzlibrary{fit}
\usetikzlibrary{calc,shapes}
\usepackage{enumitem}
\usepackage[round]{natbib}
\usepackage{hyperref}
\usepackage{sublabel}
\usepackage{appendix}
\usepackage{caption,subcaption}

\renewcommand{\algorithmicrequire}{\textbf{Input:}}
\renewcommand{\algorithmicensure}{\textbf{Output:}}

 \newcommand{\IGNORE}[1]{}

\def\nn{\nonumber}

\newcommand\E{\mathbb{E}}
\newcommand\R{\mathbb{R}}

\def\Pc{\mathcal{P}}

\DeclareMathOperator{\diag}{Diag}
\DeclareMathOperator{\Const}{Const.}

\DeclareMathOperator{\Stopping}{S}
\def\Astar{A}
\def\Bstar{B}
\def\Cstar{C}

\def\wstar{w}
\def\astar{a}
\def\bstar{b}
\def\cstar{c}
\def\lambdastar{\lambda}

\newcommand\h{\widehat}

\newcommand\tl{\tilde}

\newcommand\mat{\operatorname{mat}}

\newcommand\poly{\operatorname{poly}}

\newcommand\dt[1]{\ensuremath{\Delta#1}}

\def\tl{\tilde}

\newcommand\inner[1]{\ensuremath{\langle #1 \rangle}}

\DeclareMathOperator{\polylog}{polylog}

 \DeclareMathOperator*{\argmin}{arg\,min}
 \DeclareMathOperator*{\argmax}{arg\,max}

\DeclareMathOperator{\dist}{dist}

\def\tha{{\mbox{\tiny th}}}

\DeclareMathOperator{\Diag}{Diag}

 \def\0{{\bf 0}}

\def\viz{{viz.,\ \/}}

\def\for{\,\,\mbox{for}\quad}

\def\nn{\nonumber}
\def\ni{\noindent}

\def\qed{\hfill\hbox{${\vcenter{\vbox{
    \hrule height 0.4pt\hbox{\vrule width 0.4pt height 6pt
    \kern5pt\vrule width 0.4pt}\hrule height 0.4pt}}}$}}

\definecolor{myred}{rgb}{0.3,0.0,0.7}
\definecolor{dkg}{rgb}{0.1,0.7,0.2}
\definecolor{dkb}{rgb}{0.0,0.2,0.8}

 \def\ha{\hat{a}}
 \def\hb{\hat{b}}
 \def\hc{\hat{c}}

 \def\hw{\hat{w}}

 \def\hA{\hat{A}}
 \def\hB{\hat{B}}
 \def\hC{\hat{C}}

\def\hT{\hat{T}}

\def\bfi{{\mathbf i}}

\def\Pc{{\cal P}}

\def\Sc{{\cal S}}

\def\Ebb{{\mathbb E}}

\def\Rbb{{\mathbb R}}

\newcommand{\bprfof}{\begin{proof_of}}
\newcommand{\eprfof}{\end{proof_of}}
\newcommand{\bprf}{\begin{myproof}}
\newcommand{\eprf}{\end{myproof}}
\newcommand{\bp}{\begin{psfrags}}
\newcommand{\ep}{\end{psfrags}}
\newcommand{\bl}{\begin{lemma}}
\newcommand{\el}{\end{lemma}}
\newcommand{\bt}{\begin{theorem}}
\newcommand{\et}{\end{theorem}}
\newcommand{\bc}{\begin{center}}
\newcommand{\ec}{\end{center}}
\newcommand{\bi}{\begin{itemize}}
\newcommand{\ei}{\end{itemize}}
\newcommand{\ben}{\begin{enumerate}}
\newcommand{\een}{\end{enumerate}}
\newcommand{\bd}{\begin{definition}}
\newcommand{\ed}{\end{definition}}
\def\beq{\begin{equation}}
\def\eeq{\end{equation}\noindent}
\def\beqn{\begin{eqnarray}}
\def\eeqn{\end{eqnarray} \noindent}
\def\beqnn{  \begin{eqnarray*}}
\def\eeqnn{\end{eqnarray*}  \noindent}
\def\bcase{  \begin{numcases}}
\def\ecase{\end{numcases}   \noindent}
\def\bsbcase{  \begin{subnumcases}}
\def\esbcase{\end{subnumcases}   \noindent}

\newtheorem{theorem}{Theorem}
\newtheorem{corollary}{Corollary}
\newtheorem{lemma}{Lemma}

\newtheorem{claim}{Claim}

\newtheorem{fact}{Fact}

\newtheorem{definition}{Definition}
\newtheorem{remark}{Remark}

\newenvironment{myproof}{\noindent{\bf Proof:} \hspace*{1em}}{
    \hspace*{\fill} $\Box$ }
\newenvironment{proof_of}[1]{\noindent {\bf Proof of #1: }}{\hspace*{\fill} $\Box$ }

\newcommand{\matplottc}[1]{               % single matlab plot twocolumn
        \unitlength .45truein
        \begin{center}
        \includegraphics{#1.ps}
        \end{picture}
        \end{center}
}

\def\psfancypar#1#2{\begingroup\def\par{\endgraf\endgroup\lineskiplimit=0pt}
               \setbox2=\hbox{\large\sc #2}
               \newdimen\tmpht \tmpht \ht2 \advance\tmpht by \baselineskip
               \font\hhuge=Times-Bold at \tmpht
               \setbox1=\hbox{{\hhuge #1}}
               \count7=\tmpht \count8=\ht1
               \divide\count8 by 1000 \divide\count7 by \count8
               \tmpht=.001\tmpht\multiply\tmpht by \count7
               \font\hhuge=Times-Bold at \tmpht
               \setbox1=\hbox{{\hhuge #1}}
               \noindent
                \hangindent1.05\wd1
               \hangafter=-2 {\hskip-\hangindent
               \lower1\ht1\hbox{\raise1.0\ht2\copy1}%
                \kern-0\wd1}\copy2\lineskiplimit=-1000pt}

\def\Kout{\setbox1=\hbox{\Huge\bf K}\hbox to
1.05\wd1{\hspace{.05\wd1}
}}
\def\Sout{\setbox1=\hbox{\Huge\bf S}\hbox to 1.05\wd1{\hspace{.05\wd1}
}}

\author{Anima Anandkumar\footnote{University of California, Irvine. Email: a.anandkumar@uci.edu} \and Rong Ge\footnote{Microsoft Research, New England. Email: rongge@microsoft.com} \and Majid Janzamin\footnote{University of California, Irvine. Email: mjanzami@uci.edu}}

\title{Guaranteed Non-Orthogonal Tensor Decomposition \\via Alternating Rank-$1$ Updates}

\begin{document}
\maketitle

\begin{abstract}
In this paper, we provide local and global convergence guarantees for recovering CP (Candecomp/Parafac) tensor decomposition. The main step of the proposed algorithm is a simple alternating rank-$1$ update which is the alternating version of the tensor power iteration adapted for asymmetric tensors. Local convergence guarantees  are established for third order tensors of rank $k$ in $d$ dimensions, when $k=o \bigl( d^{1.5} \bigr)$  and the tensor components are  incoherent. Thus, we can recover overcomplete tensor decomposition. We also strengthen the results to global convergence guarantees under stricter rank condition $k \le \beta d$ (for arbitrary constant $\beta > 1$) through a simple initialization procedure where the algorithm is initialized by top singular vectors of random tensor slices. Furthermore, the approximate local convergence guarantees for $p$-th order tensors are also provided under rank condition $k=o \bigl( d^{p/2} \bigr)$.
The guarantees also include tight perturbation analysis given noisy tensor.

%A simple alternating rank-$1$ update procedure is considered for CP tensor decomposition. Local convergence guarantees  are established for third order tensors of rank $k$ in $d$ dimensions, when $k=o \bigl( d^{1.5} \bigr)$  and the tensor components are  incoherent. We strengthen the results to global convergence guarantees when $k \le \beta d$ (for arbitrary constant $\beta > 1$) through a simple initialization procedure based on rank-$1$ singular value decomposition of random tensor slices. Furthermore, the local convergence guarantees for $p$-th order tensors are also provided under rank condition $k=o \bigl( d^{p/2} \bigr)$.
%The guarantees also include tight perturbation analysis given noisy tensor.

%Our tight perturbation analysis leads to  efficient sample guarantees for unsupervised learning of several latent variable models when $k=O(d)$, where $k$ is the number of mixture components and $d$ is the observed dimension. For learning overcomplete decompositions $(k=\omega(d))$, we prove that having an extremely small number of labeled samples, scaling as $\polylog(k)$ for each label, under the semi-supervised setting  (where the label corresponds to the choice variable in the mixture model) leads to global convergence guarantees for learning mixture models.
\end{abstract}

\paragraph{Keywords:} Tensor decomposition, alternating minimization, overcomplete representation, latent variable models. %unsupervised and semi-supervised learning, latent variable models.
\section{Introduction}

Tensor decompositions  have been recently popular for unsupervised learning of a wide range of latent variable models   such as independent component analysis~\citep{de2007fourth}, topic models, Gaussian mixtures, hidden Markov models~\citep{AnandkumarEtal:tensor12}, network community models~\citep{AnandkumarEtal:community12},  and so on.
The decomposition of a certain low order multivariate moment tensor (typically up to fourth order) in these models is guaranteed to provide a consistent estimate of the model parameters.   Moreover, the sample and computational requirements are only a low order  polynomial in the rank of the tensor~\citep{AnandkumarEtal:tensor12,SongEtal:NonparametricTensorDecomp}.
In practice, the tensor decomposition techniques have been shown to be effective in a number of applications such as blind source separation~\citep{comon2002tensor}, computer vision~\citep{vasilescu2003multilinear}, contrastive topic modeling~\citep{zou2013contrastive}, and community detection~\citep{AnandkumarEtal:communityimplementation13}. In many cases, the tensor approach is shown to be orders of magnitude faster than existing techniques such as the stochastic variational approach.

The state of art for guaranteed tensor decomposition involves
two steps: converting the input tensor to an orthogonal symmetric form, and then solving the orthogonal decomposition through tensor eigen decomposition~\citep{Comon94,SIMAX-080148-Tensor-Eigenvalues,ZG01,AnandkumarEtal:tensor12}. The first step of converting the input tensor to an orthogonal symmetric form is  known as {\em whitening}. For the second step, the tensor eigen pairs can be found through a  simple tensor {\em power} iteration procedure.

While having efficient guarantees, the above procedure suffers from a number of theoretical and practical  limitations.
For instance, in practice, the     learning performance is  especially sensitive to  whitening~\citep{le2011ica}. Moreover, whitening is  computationally the most expensive step in deployments~\citep{AnandkumarEtal:communityimplementation13}, and it   can suffer from numerical instability   in high-dimensions   due to  ill-conditioning. Lastly, the above approach  is unable to learn {\em overcomplete representations} (this is the case when number of features/components is much larger than the dimension) due to the orthogonality constraint, which is especially limiting, given the recent popularity of overcomplete feature learning in many domains~\citep{bengio2012unsupervised,lewicki2000learning}.

The current practice for  tensor decomposition is the {\em alternating least squares} (ALS) procedure, which has been described as the ``workhorse'' of tensor decomposition~\citep{kolda_survey}. This involves solving the least squares problem on a {\em mode} of the tensor, while keeping the other modes fixed, and alternating between the tensor modes. The method is extremely fast since it involves calculating linear updates, but is not    guaranteed to converge to the global optimum in general~\citep{kolda_survey}.

In this paper, we provide local and global convergence guarantees for a modified alternating method, for which the main step is making rank-$1$ updates along different modes of the tensor. This update is basically a rank-1 ALS update. This method is extremely fast to deploy, trivially parallelizable, and does not suffer from  ill-conditioning issues faced by both ALS~\citep{kolda_survey} and whitening approaches~\citep{le2011ica}. Our analysis assumes the presence of {\em incoherent} tensor components, which can be viewed as a {\em soft-orthogonality} constraint. Incoherent representations have been extensively considered in literature in a number of contexts, e.g., compressed sensing~\citep{donoho2006compressed} and sparse coding~\citep{Arora2013,AgarwalEtal:SparseCoding2013}. Incoherent representations provide flexible modeling,   can handle   overcomplete   signals, and are robust to noise~\citep{lewicki2000learning}. Moreover, when  the latent variable model parameters are {\em generic} or  when we have randomly constructed (multiview)  features~\citep{mcwilliams2013correlated}, the moment   tensors have  incoherent components, as assumed here.  In this work, we establish that incoherence leads to efficient guarantees for tensor decomposition. The guarantees also include a tight perturbation analysis. In a  subsequent work \citep{OvercompleteLVMs2014}, we apply the tensor decomposition guarantees of this paper to various learning settings, and derive sample complexity bounds through novel covering arguments.

\subsection{Summary of results}

In this paper, we propose and analyze an algorithm for non-orthogonal CP (Candecomp/Parafac) tensor decomposition; see Figure~\ref{fig:TensorDecomposition} for the details of the algorithm. The main step of the algorithm is a simple alternating rank-$1$ update which is the alternating version of the tensor power iteration adapted for asymmetric tensors. In each iteration, one of the tensor modes is updated by projecting the other modes along their estimated directions, and the process is alternated between all the modes of the tensor; see~\eqref{eqn:asymmetric power update} for this update.

For the above update, we provide local convergence guarantees under incoherent tensor components for a rank-$k$  third order tensor in $d$ dimensions. We prove a linear rate of convergence   under appropriate initialization when $k = o(d^{3/2})$. Due to incoherence, the actual tensor components are not the stationary points of the update (even in the noiseless setting), and thus, there is an approximation error in the estimate after this update. The approximation error depends on the extent of overcompleteness, and scales as\,\footnote{$\tilde{O}$ is $O$ up to $\polylog$ factors.} $\tilde{O} (\sqrt{k}/d )$, which is small since $k = o(d^{3/2})$.
%Here, $\gamma := \frac{w_{\max}}{w_{\min}}$ where $w_{\max}$ and $w_{\min}$ are bounds on the norm of rank-1 components of the tensor, see \eqref{eqn:tensordecomp}.
The generalization to higher order tensors is also provided.
To the best of our knowledge, we give the first guarantees for overcomplete tensor decomposition under mild incoherence conditions.

In order to remove the approximation error $\tilde{O} (\sqrt{k}/d )$ after the above rank-1 updates, we propose an additional update to the algorithm which is basically a type of coordinate descent update; see~\eqref{eqn:BiasRemoval}. We run this update after the main rank-1 updates and show that this removes the approximation error in a linear rate of convergence, and thus, we finally consistently recover the tensor decomposition.

In the undercomplete or mildly overcomplete settings $(k = O(d))$, a simple initialization procedure (see Procedure~\ref{algo:SVD init}) based on rank-$1$ SVD of random tensor slices is provided. This initialization procedure lands the estimate in the basin of attraction for the alternating update procedure in polynomial number of trials (in the tensor rank  $k$). This leads to global convergence guarantees for tensor decomposition. %: the algorithm returns a tensor whose components are $\tilde{O}(\sqrt{k}/d)$ close to the correct tensor.
%To the best of our knowledge, these are first guarantees for tensor decomposition under incoherent tensor components. %\aacomment{Rong: how does this read?}

%without using any preprocessing such as whitening (but assuming incoherent components).
%\rgcomment{I commented the above argument out because there are some approaches that do not do whitening explicitly, I guess we should more focus on the better dependency on condition number}

We then extend the global convergence guarantees to settings where two modes of the tensor are (sufficiently) undercomplete (the dimension $d_u$ is much larger than tensor rank $k$), and the third tensor mode is (highly) overcomplete (the dimension $d_o$ is much smaller than tensor rank $k$). For instance, consider tensors arising from multi-view mixture models such as $\Ebb[x_1\otimes x_2 \otimes y]$, where $x_i$ are multi-view high dimensional features and $y$ is a low dimensional label. Previous procedures in~\citep{AnandkumarEtal:tensor12} which rely on transforming the input tensor to an orthogonal symmetric form cannot handle this   setting. Algorithms based on simultaneous diagonalization \citep{harshman1994parafac} can handle this case, but is not as robust to noise.
We prove global convergence guarantees  by considering
rank-$1$ SVD of random tensor slices along the $y$-mode as initialization for the $x_i$-modes of the tensor, and then running the alternating update procedure.

\paragraph{Overview of techniques:} Greedy or rank-$1$ updates are perhaps the most natural procedure for CP tensor decomposition. For orthogonal tensors, they lead to guaranteed recovery~\citep{ZG01}. However, when the tensor is non-orthogonal, greedy procedure is not optimal in general~\citep{kolda2001orthogonal}. Finding tensor decomposition in general is NP-hard~\citep{TensorNPHard}. We circumvent this obstacle by limiting ourselves to tensors with incoherent components. We exploit incoherence  to prove error contraction under each step of the alternating update procedure with an approximation error, which is decaying, when $k = o(d^{1.5})$. To this end, we require tools from random matrix theory, bounds on $2 \to p$ norm for random matrices~\citep{guedon2007lp, adamczak2011chevet} for some $p<3$, and matrix perturbation results to provide tight bounds on error contraction.

\subsection{Related work}
CP tensor decomposition~\citep{carroll1970analysis}, also known as PARAFAC decomposition~\citep{harshman1970foundations, harshman1994parafac} is a classical definition for tensor decomposition with many applications. The most commonly used algorithm for CP decomposition is Alternating Least Squares (ALS)~\citep{comon2009tensor}, which has  no   convergence guarantees in general.
%A guaranteed approach for CP decomposition consists of two steps, \viz whitening the input tensor to obtain an orthogonal symmetric form, and then employing the tensor power update procedure to find the orthogonal decomposition.
 \citet{kolda2001orthogonal} and \citet{ZG01} analyze the greedy or the rank-1 updates in the orthogonal setting. In the noisy setting, \citet{AnandkumarEtal:tensor12} analyze deflation procedure for orthogonal decomposition,  and~\citet{SongEtal:NonparametricTensorDecomp} extend the analysis to the nonparametric setting. 
For the non-orthogonal tensors, a common strategy is to first apply a procedure called {\em whitening} to reduce it to the orthogonal case. But as discussed earlier, the whitening procedure can lead to poor performance and bad sample complexity. Moreover, it requires the tensor factors to have full column rank, which rules out overcomplete tensors.

Learning overcomplete tensors is challenging, and they may not even be identifiable in general. \citet{Kruskal:76,Kruskal:77} provided an identifiability result based on the {\em Kruskal} rank of the factor matrices of the tensor. However, this result is limiting since it requires $k=O(d)$, where $k$ is the tensor rank and $d$ is the dimension. The FOOBI procedure by~\citet{de2007fourth} overcomes this limitation by assuming {\em generic} factors, and shows that a polynomial-time procedure can recover the tensor components when $k =O(d^2)$, and the tensor is fourth order. However, the procedure does not work for third-order overcomplete tensors, and has no polynomial sample complexity bounds. Simple procedures can recover overcomplete tensors for higher order tensors (five or higher). For instance, for the fifth order tensor, when $k=O(d^2)$, we can utilize random slices along a mode of the tensor, and perform simultaneous diagonalization on the matricized versions. Note that this  procedure cannot handle the same level of overcompleteness as FOOBI, since an additional dimension is required for obtaining two (or more) fourth order tensor slices. The simultaneous diagonalization  procedure   entails careful perturbation analysis, carried out by~\citep{fourierpca,bhaskara2013smoothed}. In addition, \citet{fourierpca} provide stronger results for independent components analysis (ICA), where the tensor slices can be obtained in the Fourier domain. %\tcr{\citet{GMMICA2013} convert the problem of learning Gaussian mixtures to an ICA problem and exploit the Fourier PCA method in \citet{fourierpca} to learn that. More precisely, they prove that a Gaussian mixture with known identical covariance matrices whose number of components is a polynomial of any fixed degree in the dimension $d$ is polynomially learnable as long as a certain non-degeneracy condition on the means is satisfied.}

There are other recent works which can learn overcomplete models, but under different settings than the ones considered in this paper. For instance,~\citet{Arora2013,AgarwalEtal:SparseCoding2013} provide guarantees for the sparse coding problem.~\citet{AnandkumarEtal:NIPS13} learn overcomplete sparse topic models, and provide guarantees for {\em Tucker} tensor decomposition under sparsity constraints. Specifically, the model is identifiable using $(2n)^{\tha}$ order moments when the latent dimension  $k=O(d^n)$ and   the sparsity level of the factor matrix is   $O(d^{1/n})$, where $d$ is the observed dimension. The Tucker decomposition is different from the CP decomposition considered here (it has weaker assumptions and guarantees), and the techniques in~\citep{AnandkumarEtal:NIPS13} differ significantly from the ones considered here.

The algorithm employed here falls under the general framework of alternating minimization. There are many recent works which provide guarantees on local/global convergence for alternating minimization, e.g., for matrix completion~\citep{jain2013low, hardt2013provable}, phase retrieval~\citep{netrapalli2013phase} and sparse coding~\citep{AgarwalEtal:SparseCoding2013}. However, the techniques in this paper are significantly different, since they involve tensors, while the previous works only required matrix analysis.

%\subsection*{Motivation}
%Unsupervised learning: mixture and latent models. \\
%Supervised learning: Let tensor $T$ corresponds to $\Ebb[x \otimes x \otimes y]$, where $x$ is the feature and $y$ is the label. This setting can be also represented by a multi-view model (Unsupervised setting), where $y$ is one of the views.
%Semi-supervised learning: We can use training set to provide good initialization vectors, which is not possible in the unsupervised setting. This is discussed at the end of Section \ref{sec:initialization}.
%We also need to motivate the incoherence assumption: 1) Generic parameters have incoherence property. 2) random projection of features to lower dimension.
%Comparable with whitening plus orthogonal decomposition: need to compare the method with tensor power approach after whitening (and symmetrization). Our alternating approach is more efficient since it can be implemented in parallel and each step only involves matrix-vector products. need to have a detailed comparison. Whitening requires assumptions on condition numbers so it ties everything together but here that is not the issue. Moreover, symmetrization requires additional SVDs that can be avoided.
%For the overcomplete setting, it is sometimes useful to think of only recovering the top few components of the decomposition (corresponding to strong latent components).. so even if we establish guarantees for this setting, it is good.
%
%\subsection*{Summary of contributions}
%\bi
%\item Local convergence
%\item Initialization
%\item Perturbation analysis
%\ei

\subsection{Notations and tensor preliminaries}
Let $[n]$ denote the set $\{1,2,\dotsc,n\}$. 

Notice that while the standard asymptotic notation is to write $f(d) = O(g(d))$ and $g(d) = \Omega(f(d))$, we sometimes use $f(d) \leq O(g(d))$ and $g(d) \geq \Omega(f(d))$ for additional clarity.
We also use the asymptotic notation $f(d) = \tl{O}(g(d))$ if and only if $f(d) \leq \alpha g(d)$ for all $d \geq d_0$, for some $d_0 >0$ and $\alpha = \polylog(d)$, i.e., $\tilde{O}$ hides $\polylog$ factors.

\subsubsection*{Tensor preliminaries}
A real \emph{$p$-th order tensor} $T \in \bigotimes_{i=1}^p \R^{d_i}$ is a member of the outer product of Euclidean spaces $\R^{d_i}$, $i \in [p]$.
For convenience, we restrict to the case where $d_1 = d_2 = \dotsb = d_p = d$, and simply write $T \in \bigotimes^p \R^d$.
%For vector $v \in \R^d$, we use $v^{\otimes p} := v \otimes v \otimes
%\dotsb \otimes v \in \bigotimes^p \R^d$ to denote its $p$-th tensor power.
As is the case for vectors (where $p=1$) and matrices (where $p=2$), we may
identify a $p$-th order tensor with the $p$-way array of real numbers $[
T_{i_1,i_2,\dotsc,i_p} \colon i_1,i_2,\dotsc,i_p \in [d] ]$, where
$T_{i_1,i_2,\dotsc,i_p}$ is the $(i_1,i_2,\dotsc,i_p)$-th coordinate of $T$
with respect to a canonical basis. For convenience, we limit to third order tensors $(p=3)$ in our analysis, while the results for higher order tensors are also provided.

The different dimensions of the tensor are referred to as {\em modes}. For instance, for a matrix, the first mode refers to columns and the second mode refers to rows.
In addition,￼ {\em fibers} are higher order analogues of matrix rows and columns. A fiber is obtained by fixing all but one of the indices of the tensor (and is arranged as a column vector). For instance, for a matrix, its mode-$1$ fiber is any matrix column while a mode-$2$ fiber is any row. For a
third order tensor $T\in \R^{d \times d \times d}$, the mode-$1$ fiber is given by $T(:, j, l)$, mode-$2$ by $T(i, :, l)$ and mode-$3$ by $T(i, j, :)$.
Similarly, {\em slices} are obtained by fixing all but two of the indices of the tensor. For example, for the third order tensor $T$, the slices along $3$rd mode are given by $T(:, :, l)$.
For $r \in \{1,2,3\}$, the mode-$r$ matricization of a third order tensor $T\in \R^{d \times d \times d}$, denoted by $\mat(T,r) \in \Rbb^{d \times d^2}$, consists of all mode-$r$ fibers arranged as column vectors.

We view a tensor $T \in \Rbb^{d \times d \times d}$ as a multilinear form. %For simplicity let us consider order-3 tensors.
Consider matrices $M_r \in \R^{d\times d_r}, r \in \{1,2,3\}$. Then tensor $T(M_1,M_2,M_3) \in \R^{d_1}\otimes \R^{d_2}\otimes \R^{d_3}$ is defined as
\begin{align} \label{eqn:multilinear form def}
T(M_1,M_2,M_3)_{i_1,i_2,i_3} := \sum_{j_1, j_2,j_3\in[d]} T_{j_1,j_2,j_3} \cdot M_1(j_1, i_1) \cdot M_2(j_2, i_2) \cdot M_3(j_3, i_3).
\end{align}
%In particular, if $u$, $v$ and $w$ are vectors and $T$ is a $3$rd order tensor, then $T(u,v,w)$ is a scalar, $T(I,v,w)$ is a vector, and $T(I, I, w)$ is a matrix.
In particular, for vectors $u,v,w \in \R^d$, we have\,\footnote{Compare with the matrix case where for $M \in \R^{d \times d}$, we have $ M(I,u) = Mu := \sum_{j \in [d]} u_j M(:,j) \in \R^d$.}
\begin{equation} \label{eqn:rank-1 update}
 T(I,v,w) = \sum_{j,l \in [d]} v_j w_l T(:,j,l) \ \in \R^d,
\end{equation}
which is a multilinear combination of the tensor mode-$1$ fibers.
Similarly $T(u,v,w) \in \R$ is a multilinear combination of the tensor entries,  and $T(I, I, w) \in \R^{d \times d}$ is a linear combination of the tensor slices.

A $3$rd order tensor $T \in \Rbb^{d \times d \times d}$ is said to be rank-$1$ if it can be written in the form
\begin{align} \label{eqn:rank-1 tensor}
T= w \cdot a \otimes b\otimes c \Leftrightarrow T(i,j,l) = w \cdot a(i) \cdot b(j) \cdot c(l),
\end{align}
where notation $\otimes$  represents the {\em outer product} and $a \in \Rbb^d$, $b \in \Rbb^d$, $c \in \Rbb^d$ are unit vectors (without loss of generality).
A tensor $T  \in \Rbb^{d \times d \times d}$ is said to have a CP rank $k\geq 1$ if it can be written as the sum of $k$ rank-$1$ tensors
\begin{equation}\label{eqn:tensordecomp}
T = \sum_{i\in [k]} w_i a_i \otimes b_i \otimes c_i, \quad w_i \in \Rbb, \ a_i,b_i,c_i \in \Rbb^d.
\end{equation}
This decomposition is closely related to the multilinear form. In particular, for vectors $\ha,\hb,\hc \in \Rbb^d$, we have
$$T(\ha,\hb,\hc) = \sum_{i\in [k]} w_i \langle a_i, \ha\rangle\langle b_i, \hb\rangle\langle c_i,\hc\rangle.$$
Consider the decomposition in equation~\eqref{eqn:tensordecomp},
denote matrix $A:=[a_1 \ a_2 \ \dotsb \ a_k] \in \R^{d \times k}$, and similarly $B$ and $C$.    Without loss of generality, we assume that the matrices have normalized columns (in $2$-norm), since we can always rescale them, and adjust the weights $w_i$ appropriately.

Throughout, $\|v\| := (\sum_i v_i^2)^{1/2}$ denotes the Euclidean ($\ell_2$) norm
of a vector $v$, and $\|M\|$ denotes the spectral (operator) norm of a matrix $M$.  
Furthermore, $\|T\|$ and $\|T\|_F$ denote the spectral (operator) norm and the Frobenius norm of a tensor, respectively. In particular, for a $3$rd order tensor, we have
$$
\|T\| := \sup_{\|u\| = \|v\| = \|w\| = 1} |T(u,v,w)|, \quad \|T\|_F := \sqrt{\sum_{i,j,l \in [d]} T_{i,j,l}^2}.
$$

\section{Tensor Decomposition Algorithm} \label{sec:algorithm}

%\begin{wrapfigure}{r}{3.5in}
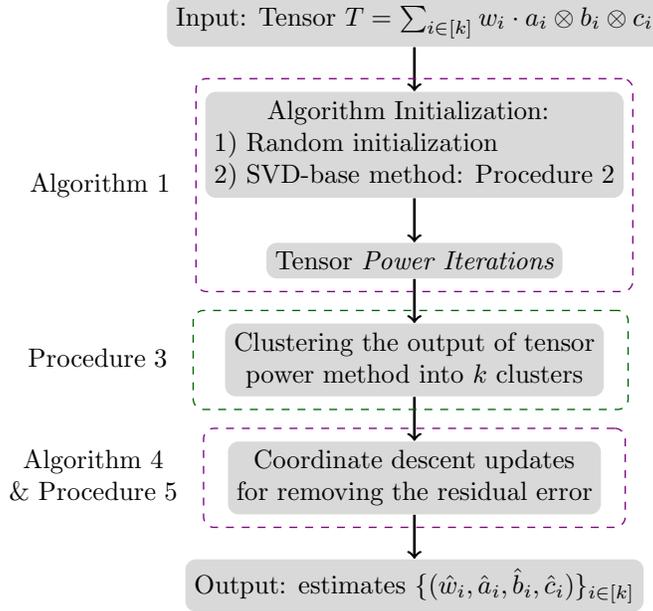
\begin{figure}%[t]
\bc
\begin{tikzpicture}
[
scale=1,
   nodestyle/.style={fill = gray!30, shape = rectangle, rounded corners, minimum width = 2cm},
]
\small
\matrix [column sep=2mm,row sep=6mm] {
\node[nodestyle](a1){Input: Tensor $T=\sum_{i \in [k]} w_i \cdot a_i \otimes b_i \otimes c_i$}; & \\
\node[nodestyle, align=left](a){\quad \quad Algorithm Initialization: \\ 1) Random initialization \\ 2) SVD-base method: Procedure~\ref{algo:SVD init}}; & \\
\node[nodestyle, align=center](b){Tensor {\em Power Iterations}}; & \\
\node[nodestyle, align=center](c){Clustering the output of tensor \\ power method into $k$ clusters}; &
%\node[nodestyle, align=center](c2){Labeled data: \\ $\{(x_i,y_i)\}$};
\\
\node[nodestyle, align=center](d){Coordinate descent updates \\ for removing the residual error}; & \\
\node[nodestyle](e){Output: estimates $\lbrace (\hw_i, \ha_i, \hb_i, \hc_i)  \rbrace_{i \in [k]}$}; & \\
%\node[nodestyle, align=center](e){Spectral/tensor method: \\ find $u_j$'s s.t.\ $\displaystyle\Ebb \left[ \nabla^{(m)} G(x) \right]= \sum_{j \in [k]} u_j^{\otimes m}$}; & \\
%\node[nodestyle, align=center](f){Extract discriminative features using $u_j$'s/ \\ do model-based prediction with $u_j$'s as parameters}; & \\
};
\draw [->, line width = 1pt] (a1) to (a);
\draw [->, line width = 1pt] (a) to (b);
\draw [->, line width = 1pt] (b) to (c);
\draw [->, line width = 1pt] (c) to (d);
\draw [->, line width = 1pt] (d) to (e);
%\draw [->, line width = 1pt] (e) to (f);
%\draw [->, line width = 1pt] (c2) to (c);

%\node [draw, dashed, rounded corners, violet, line width=0.5pt,
%	fit = {(a) ($(a.east)+(2mm,0)$) ($(a.west)-(2mm,0)$) ($(a.north)+(0,0.5mm)$) ($(a.south)-(0,0.5mm)$)},
%	label=left:{\begin{tabular}{c} Using lables \\ or Procedure~\ref{algo:SVD init}\end{tabular}}
%	] {};

\node [draw, dashed, rounded corners, violet, line width=0.5pt,
	fit = {(a) (b)  ($(b.east)+(2mm,0)$) ($(b.west)-(3mm,0)$) ($(a.north)+(0,0.5mm)$) ($(b.south)-(0,0.5mm)$)},
	label=left:{\begin{tabular}{c} Algorithm~\ref{algo:Power method form} \end{tabular}}
	] {};
	
\node [draw, dashed, rounded corners, green!40!black, line width=0.5pt,
	fit = {(c)  ($(c.east)+(2mm,0)$) ($(c.west)-(3mm,0)$) ($(c.north)+(0,0.5mm)$) ($(c.south)-(0,0.5mm)$)},
	label=left:{\begin{tabular}{c} Procedure~\ref{alg:cluster} \end{tabular}}
	] {};

\node [draw, dashed, rounded corners, violet, line width=0.5pt,
	fit = {(d) ($(d.east)+(2mm,0)$) ($(d.west)-(2mm,0)$) ($(d.north)+(0,0.5mm)$) ($(d.south)-(0,0.5mm)$)},
	label=left:{\begin{tabular}{c} Algorithm~\ref{algo:coordinate-descent}\\ \& Procedure~\ref{algo:fix-procedure} \end{tabular}}
	] {};
%\draw[<->] ($(-5,4)$) to ($(-5,2.6)$);
%\node [align=center] at ($(-7.3,3.3)$) {Unsupervised estimation of \\ score functions};
%\draw[<->] ($(-5,1.5)$) to ($(-5,-3.8)$);
%\node [align=center] at ($(-8,-1)$) {Supervised framework for \\ extracting discriminative features};
\end{tikzpicture}
\ec
\vspace{-0.2in}
\caption{\small Overview of tensor decomposition algorithm.}
\label{fig:TensorDecomposition}
\end{figure}
%\end{wrapfigure}

In this section, we introduce the alternating tensor decomposition algorithm, and the guarantees are provided in Section~\ref{sec:analysis}. The goal of tensor decomposition algorithm is to recover the rank-1 components of tensor; see~\eqref{eqn:tensordecomp} for the notion of tensor rank.
Figure~\ref{fig:TensorDecomposition} depicts the overview of our tensor decomposition method where the corresponding algorithms and procedures are also specified. Our algorithm includes two main steps as 1) alternating tensor power iteration, and 2) coordinate descent iteration for removing the residual error. The former one is performed in Algorithm~\ref{algo:Power method form} (see equation~\eqref{eqn:asymmetric power update}, and the latter one is done in Algorithm~\ref{algo:coordinate-descent} (see equation~\eqref{eqn:BiasRemoval}). We now describe these steps of the algorithm in more details as well as providing the auxiliary procedures required to complete the algorithm.

\subsection{Tensor power iteration in Algorithm~\ref{algo:Power method form}}

%The rank-$1$ alternating update method for tensor decomposition is given in Algorithm \ref{algo:Power method form}. Given an initial estimate of the vectors denoted by $\bigl( \ha^{(0)}, \hb^{(0)}, \hc^{(0)} \bigr)$, an  {\em asymmetric power update}\,\footnote{This is exactly the generalization of asymmetric matrix power update to $3$rd order tensors.} in \eqref{eqn:asymmetric power update} on the input tensor $T$ is performed in each iteration of the algorithm. Notice that the updates in \eqref{eqn:asymmetric power update} alternate among different modes of the tensor.

The main step of the algorithm is tensor power iteration which basically performs alternating {\em asymmetric power updates}\,\footnote{This is exactly the generalization of asymmetric matrix power update to $3$rd order tensors.}
on different modes of the tensor as %(see \eqref{eqn:rank-1 update} for the definition of multilinear form)
\begin{equation} \label{eqn:asymmetric power update}
\ha^{(t+1)} = \frac{T \left( I, \hb^{(t)}, \hc^{(t)} \right)}{\left\| T \left( I, \hb^{(t)}, \hc^{(t)} \right) \right\|}, \ \
\hb^{(t+1)} = \frac{T \left( \ha^{(t)}, I, \hc^{(t)} \right)}{\left\| T \left( \ha^{(t)}, I, \hc^{(t)} \right) \right\|}, \ \
\hc^{(t+1)} = \frac{T \left( \ha^{(t)}, \hb^{(t)},I \right)}{\left\| T \left( \ha^{(t)}, \hb^{(t)},I \right) \right\|},
\end{equation}
where $\{\ha^{(t)},\hb^{(t)},\hc^{(t)}\}$ denotes estimate in the $t$-th iteration.
Recall that for vectors $v,w \in \R^d$, the multilinear form $T(I,v,w) \in \R^d$ used in the above update formula is defined in \eqref{eqn:rank-1 update}, where $T(I,v,w)$ is a multilinear combination of the tensor mode-$1$ fibers.
Notice that the updates alternate among different modes of the tensor which can be viewed as a rank-$1$ form of the standard Alternating Least Squares (ALS) method. We later discuss this relation in more details.

\paragraph{Optimization viewpoint:}
Consider the problem of best rank-$1$ approximation of tensor $T$ as
\begin{equation} \label{eqn:power-opt}
\min_{\substack{a,b,c \in \Sc^{d-1} \\ w \in \Rbb}} \| T - w \cdot a \otimes b \otimes c \|_F,
\end{equation}
where $\Sc^{d-1}$ denotes the unit $d$-dimensional sphere. This optimization program is non-convex, and has multiple local optima. It can be shown that the updates in \eqref{eqn:asymmetric power update} are the alternating optimization for this program where in each update, optimization over one vector is performed while the other two vectors are assumed fixed. This alternating minimization approach does not converge to the true components of tensor $T$ in general, and in this paper we provide sufficient conditions for the convergence guarantees.

\begin{algorithm}[t]
\caption{Tensor decomposition via alternating asymmetric power updates}
\label{algo:Power method form}
\begin{algorithmic}[1]
%\REQUIRE Tensor $T \in \Rbb^{d \times d \times d}$, and stopping threshold $t_{\Stopping}$.
\REQUIRE Tensor $T \in \Rbb^{d \times d \times d}$, number of initializations $L$, number of iterations $N$.
\FOR{$\tau=1$ \TO $L$}
\STATE \textbf{Initialize} unit vectors $\ha_\tau^{(0)} \in \Rbb^d$, $\hb_\tau^{(0)} \in \Rbb^d$, and $\hc_\tau^{(0)} \in \Rbb^d$ as
\bi[itemsep=-1mm]
\vspace{-2mm}
\item Option 1: SVD-based method in Procedure~\ref{algo:SVD init} when  $k\leq \beta d$ for arbitrary constant $\beta$. 
\item Option 2: random initialization.
\ei
%\STATE $t \leftarrow 0$.
\FOR{$t=0$ \TO $N-1$}%\REPEAT%\WHILE{not converged}
\STATE Asymmetric power updates (see \eqref{eqn:rank-1 update} for the definition of the multilinear form):
\begin{align*} %\label{eqn:asymmetric power update}
\ha_\tau^{(t+1)} = \frac{T \left( I, \hb_\tau^{(t)}, \hc_\tau^{(t)} \right)}{\left\| T \left( I, \hb_\tau^{(t)}, \hc_\tau^{(t)} \right) \right\|}, \quad
\hb_\tau^{(t+1)} = \frac{T \left( \ha_\tau^{(t)}, I, \hc_\tau^{(t)} \right)}{\left\| T \left( \ha_\tau^{(t)}, I, \hc_\tau^{(t)} \right) \right\|}, \quad
\hc_\tau^{(t+1)} = \frac{T \left( \ha_\tau^{(t)}, \hb_\tau^{(t)},I \right)}{\left\| T \left( \ha_\tau^{(t)}, \hb_\tau^{(t)},I \right) \right\|}.
\end{align*}
%\STATE $t \leftarrow t+1.$
\ENDFOR%\UNTIL{$\max \left( \left\| \ha_\tau^{(t)} - \ha_\tau^{(t-1)} \right\|^2, \left\| \hb_\tau^{(t)} - \hb_\tau^{(t-1)} \right\|^2, \left\| \hc_\tau^{(t)} - \hc_\tau^{(t-1)} \right\|^2  \right) \leq t_{\Stopping}$}%\ENDWHILE
\STATE weight estimation:
\begin{align} \label{eqn:weight update}
\hw_\tau = T \left( \ha_\tau^{(N)}, \hb_\tau^{(N)}, \hc_\tau^{(N)} \right).
\end{align}
%\mjcomment{Earlier, the following update was proposed which is closely related to the above one.
%\begin{align*}
%\hw_\tau^{(t+1)} = \sqrt[3]{\left\| T \left( I, \hb_\tau^{(t)}, \hc_\tau^{(t)} \right) \right\| \cdot \left\| T \left( \ha_\tau^{(t)}, I, \hc_\tau^{(t)} \right) \right\| \cdot \left\| T \left( \ha_\tau^{(t)}, \hb_\tau^{(t)},I \right) \right\|}.
%\end{align*}}
\ENDFOR
\STATE Cluster set $\left\{ \left( \hw_\tau,\ha_\tau^{(N)},\hb_\tau^{(N)},\hc_\tau^{(N)} \right), \tau \in [L] \right\}$ into $k$ clusters as in Procedure~\ref{alg:cluster}.
\RETURN the center member of these $k$ clusters as estimates $(\hw_j,\ha_j,\hb_j,\hc_j), j \in [k]$.
\end{algorithmic}
\end{algorithm}

\paragraph{Intuition:} 
We now provide an intuitive argument on the functionality of power updates in~\eqref{eqn:asymmetric power update}.
Consider a rank-$k$ tensor $T$ as in  \eqref{eqn:tensordecomp}, and suppose we start at the correct vectors $\ha=a_j$ and $\hb=b_j$,  for some $j \in [k]$. Then for the numerator of update formula~\eqref{eqn:asymmetric power update}, we have
\begin{equation}\label{eqn:intuition}
T \left( \ha, \hb, I \right)
= T \left( a_j, b_j, I \right)
= w_j c_j + \sum_{i \neq j} w_i \langle a_j,a_i \rangle \langle b_j,b_i \rangle c_i,
%= w_j c_j + C \Diag(w) \left( \left(J_A\right)_{(j)} * \left(J_B\right)_{(j)} \right),
\end{equation}where the first term is along $c_j$ and the second term is an error term due to non-orthogonality. For orthogonal decomposition, the second term is zero, and the true vectors $a_j,b_j$ and $c_j$ are stationary points for the power update procedure. However, since we consider non-orthogonal tensors, this procedure cannot recover the decomposition exactly  leading to a residual error after running this step. 
Under incoherence conditions which encourages soft-orthogonality constraints\,\footnote{See Assumption~\ref{cond:incoherence} in Appendix~\ref{sec:assumptions} for precise description.} (and some other conditions), we show that the residual error is small (see Lemma~\ref{thm:local convergence-poweriteration} where the guarantees for the tensor power iteration step is provided), and thus, with the additional step we propose in Section~\ref{sec:coordinate-descent}, we can also remove this residual error.

\paragraph{Initialization and clustering procedures:}
We discussed that the tensor power updates in~\eqref{eqn:asymmetric power update} are the alternating iterations for the problem of rank-1 approximation of the tensor; see~\eqref{eqn:power-opt}. This is a non-convex problem and has many local optima. Thus, the power update requires careful initialization to ensure convergence to the  true rank-1 tensor components. 

For generating initialization vectors $\bigl( \ha^{(0)}, \hb^{(0)}, \hc^{(0)} \bigr)$, we introduce two possibilities. One is the simple  random initializations, where $\ha^{(0)}$ and $\hb^{(0)}$ are uniformly drawn from unit sphere $\Sc^{d-1}$. The other option is SVD-based technique in Procedure~\ref{algo:SVD init} where top left and right singular vectors of $T(I,I,\theta)$ (for some random $\theta \in \Rbb^d$) are respectively introduced as $\ha^{(0)}$ and $\hb^{(0)}$. Under both initialization procedures, vector $\hc^{(0)}$ is generated through update formula in \eqref{eqn:asymmetric power update}. We establish in Section \ref{sec:global convergence} that when $k=O(d)$, the SVD procedure leads to global convergence guarantees under polynomial number of trials. In practice random initialization also works well, however the analysis is still an open problem.%See Section \ref{sec:initialization} for more details.

%In the unsupervised setting, we also need to identify which initializations are successful in recovering the true rank-1 components of the tensor which is performed by the clustering Procedure~\ref{alg:cluster}.
Notice that the algorithm is run for $L$ different initialization vectors for which we do not know the good ones in prior. In order to identify which initializations are successful at the end, we also need a {\em clustering} step proposed in Procedure~\ref{alg:cluster} to obtain the final estimates of the vectors. The detailed analysis of clustering procedure is provided in Appendix~\ref{sec:clustering}.

\floatname{algorithm}{Procedure}
\begin{algorithm}[t]
\caption{SVD-based initialization when $k\leq \beta d$ for arbitrary constant $\beta$}
\label{algo:SVD init}
\begin{algorithmic}[1]
\REQUIRE Tensor $T \in \Rbb^{d \times d \times d}$.
\STATE Draw a random standard Gaussian vector $\theta \sim \mathcal{N}(0,I_d).$
\STATE Compute $u_1$ and $v_1$ as the top left and right singular vectors of  $T(I,I,\theta) \in \R^{d \times d}$.
\STATE $\ha^{(0)} \leftarrow u_1$, $\hb^{(0)} \leftarrow v_1$.
\STATE Initialize $\hc^{(0)}$ by update formula in \eqref{eqn:asymmetric power update}.
\RETURN $\bigl( \ha^{(0)}, \hb^{(0)}, \hc^{(0)} \bigr)$.
\end{algorithmic}
\end{algorithm}

\begin{algorithm}[t]
\caption{Clustering process}
\label{alg:cluster}
\begin{algorithmic}[1]
\REQUIRE Tensor $T \in \Rbb^{d \times d \times d}$, set of $4$-tuples %\,\footnote{Here we drop the superscript $(N)$.} 
$\left\{(\hw_\tau, \ha_\tau,\hb_\tau, \hc_\tau),\tau\in [L]\right\}$, parameter $\nu$.
\FOR{$i = 1$ \TO $k$}
\STATE Among the remaining 4-tuples, choose $\ha,\hb,\hc$ which correspond to the largest $|T(\ha,\hb,\hc)|$.
\STATE Do $N$ more iterations of alternating updates in \eqref{eqn:asymmetric power update} starting from $\ha,\hb,\hc$.
\STATE Let the output of iterations denoted by $(\ha,\hb,\hc)$ be the center of cluster $i$.
\STATE Remove all the tuples with $\max\{|\langle \ha_\tau,\ha\rangle|,|\langle \hb_\tau,\hb\rangle|,|\langle \hc_\tau,\hc\rangle|\} > \nu/2$.
\ENDFOR
\RETURN the $k$ cluster centers.
\end{algorithmic}
\end{algorithm}

\subsection{Coordinate descent iteration in Algorithm~\ref{algo:coordinate-descent}} \label{sec:coordinate-descent}
%\paragraph{Removing residual error}
%As briefly discussed above, we argue that the tensor power method recovers the true rank-1 components of the input tensor up to some residual error when the rank-1 components are not necessarily orthogonal to each other.
We discussed in the previous section that the tensor power iteration recovers  the tensor rank-1 components up to some residual error. We now propose Algorithm~\ref{algo:coordinate-descent} to remove this additional residual error. This algorithm mainly runs a coordinate descent iteration as
\begin{align}
%\tl{w}_i^{(t+1)} &= \biggl\| T \left(\h{A}_i^{(t)},\h{B}_i^{(t)},I \right) - \sum_{j\ne i} \h{w}_j^{(t)} \inner{\h{A}_i^{(t)},\h{A}_j^{(t)}} \inner{\h{B}_i^{(t)},\h{B}_j^{(t)}} \cdot \h{C}_j^{(t)} \biggr\|, \nn \\
\tl{c}_i^{(t+1)} = \operatorname{Norm} \biggl( T \left(\h{a}_i^{(t)},\h{b}_i^{(t)},I \right) - \sum_{j\ne i} \h{w}_j^{(t)} \inner{\h{a}_i^{(t)},\h{a}_j^{(t)}} \inner{\h{b}_i^{(t)},\h{b}_j^{(t)}} \cdot \h{c}_j^{(t)} \biggr), \quad i \in [k], \label{eqn:BiasRemoval}
\end{align}
where for vector $v$, we have $\operatorname{Norm} (v) := v/\|v\|$, i.e., it normalizes the vector. The above is similarly applied for updating $\tl{a}^{(t+1)}_i$ and $\tl{b}^{(t+1)}_i$.
Unlike the power iteration, it can be immediately seen that $a_i$, $b_i$ and $c_i$ are stationary points of the above update even if the components are not orthogonal to each other. Inspired by this intuition, we prove that when the residual error is small enough (as guaranteed in the analysis of tensor power iteration), this step removes it.

%combined with a fixing procedure to remove the residual error remaining after running power iteration.

The analysis of this algorithm requires that the estimate matrices $\hA, \hB, \hC$ satisfy some bound on the spectral norm and some column-wise error bounds; see Definition~\ref{def:nice-property} in Appendix~\ref{sec:convergence proof-coordinate descent} for the details. The optimization program in~\eqref{eqn:coordinate descent-opt} (which is only run in the first iteration) and projection Procedure~\ref{algo:fix-procedure} ensure that these conditions are satisfied.

\floatname{algorithm}{Algorithm}
\begin{algorithm}[h]
\caption{Coordinate descent algorithm for removing the residual error}
\label{algo:coordinate-descent}
\begin{algorithmic}[1]
\REQUIRE Tensor $T \in \Rbb^{d \times d \times d}$, initialization set $\left\{\h{A}, \h{B}, \h{C}, \h{w}^{(0)} \right\}$, number of iterations $N$.
\STATE Initialize $\h{A}^{(0)}$ as (similarly for $ \h{B}^{(0)}, \h{C}^{(0)}$)
\begin{equation} \label{eqn:coordinate descent-opt}
\h{A}^{(0)} := \argmin_{\tl{A}} \|\tilde{A}\| \quad
\operatorname{s.t.} \ \|\tilde{a}_i - \widehat{a}_i \| \le \tl{O} \left( \sqrt{k}/d \right), i \in[k].
\end{equation}
\FOR{$t=0$ \TO $N-1$}
\FOR{$i=1$ \TO $k$}
\STATE
\begin{align*}
\tl{w}_i^{(t+1)} &= \biggl\| T \left(\h{a}_i^{(t)},\h{b}_i^{(t)},I \right) - \sum_{j\ne i} \h{w}_j^{(t)} \inner{\h{a}_i^{(t)},\h{a}_j^{(t)}} \inner{\h{b}_i^{(t)},\h{b}_j^{(t)}} \cdot \h{c}_j^{(t)} \biggr\|, \nn \\
\tl{c}_i^{(t+1)} &= \frac{1}{\tl{w}_i^{(t+1)}} \biggl( T \left(\h{a}_i^{(t)},\h{b}_i^{(t)},I \right) - \sum_{j\ne i} \h{w}_j^{(t)} \inner{\h{a}_i^{(t)},\h{a}_j^{(t)}} \inner{\h{b}_i^{(t)},\h{b}_j^{(t)}} \cdot \h{c}_j^{(t)} \biggr). %\label{eqn:BiasRemoval}
\end{align*}
\ENDFOR
\STATE Update $\h{C}^{(t+1)}$ by applying Procedure~\ref{algo:fix-procedure} with inputs $\tl{C}^{(t+1)}$ and $\h{C}^{(t)}$.
\STATE Repeat the above steps (with appropriate changes) to update $\h{A}^{(t+1)}$ and $\h{B}^{(t+1)}$.
\STATE Update $\h{w}^{(t+1)}$: \\for any $i \in [k]$, 
$
\h{w}_i^{(t+1)} =
\left\{\begin{array}{ll}
\tl{w}_i^{(t+1)}, & \left| \tl{w}_i^{(t+1)} - \h{w}^{(t)}_i \right| \leq \eta_0 \frac{\sqrt{k}}{d}, \\
\h{w}^{(t)}_i + \operatorname{sgn} \left(\tl{w}_i^{(t+1)} - \h{w}^{(t)}_i \right) \cdot \eta_0 \frac{\sqrt{k}}{d}, & \operatorname{o.w.}
\end{array} \right.
$
%\begin{align*}
%\h{w}_i^{(t+1)} =
%\left\{\begin{array}{ll}
%\tl{w}_i^{(t+1)}, & \left| \tl{w}_i^{(t+1)} - \h{w}^{(t)}_i \right| \leq \eta_0 \frac{\sqrt{k}}{d}, \\
%\h{w}^{(t)}_i + \operatorname{sgn} \left(\tl{w}_i^{(t+1)} - \h{w}^{(t)}_i \right) \cdot \eta_0 \frac{\sqrt{k}}{d}, & \operatorname{o.w.}
%\end{array} \right.
%\end{align*}
\ENDFOR
\RETURN $\left\{\h{A}^{(N)}, \h{B}^{(N)}, \h{C}^{(N)}, \h{w}^{(N)} \right\}$.
\end{algorithmic}
\end{algorithm}

\floatname{algorithm}{Procedure}
\begin{algorithm}[h]
\caption{Projection procedure}
\label{algo:fix-procedure}
\begin{algorithmic}[1]
\renewcommand{\algorithmicrequire}{\textbf{input}}
\renewcommand{\algorithmicensure}{\textbf{output}}
\REQUIRE Matrices $\tl{C}^{(t+1)}$, $\h{C}^{(t)}$.
\STATE Compute the SVD of $\tl{C}^{(t+1)} = UDV^\top$.
\STATE Let $\h{D}$ be the truncated version of $D$ as $\h{D}_{i,i} := \min \left\{ D_{i,i},\eta_1\sqrt{\frac{k}{d}} \right\}.$
\STATE Let $Q := U\h{D}V^\top$.
\STATE Update $\h{C}^{(t+1)}$: for any $i \in [k]$, 
$
\h{c}_i^{(t+1)} =
\left\{\begin{array}{ll}
Q_i, & \left\|Q_i-\h{c}^{(t)}_i \right\| \le \eta_0 \frac{\sqrt{k}}{d}, \\
\h{c}^{(t)}_i + \eta_0 \frac{\sqrt{k}}{d} \frac{\left( Q_i-\h{c}^{(t)}_i \right)}{\left\|Q_i-\h{c}^{(t)}_i \right\|}, & \operatorname{o.w.}
\end{array} \right.
$
%\begin{align*}
%\h{c}_i^{(t+1)} =
%\left\{\begin{array}{ll}
%Q_i, & \left\|Q_i-\h{C}^{(t)}_i \right\| \le \eta_0 \frac{\sqrt{k}}{d}, \\
%\h{C}^{(t)}_i + \eta_0 \frac{\sqrt{k}}{d} \frac{1}{\left\|Q_i-\h{C}^{(t)}_i \right\|} \cdot \left( Q_i-\h{C}^{(t)}_i \right), & \operatorname{o.w.}
%\end{array} \right.
%\end{align*}
\RETURN $\h{C}^{(t+1)}$.
\end{algorithmic}
\end{algorithm}

\subsection{Discussions}

We now provide some further discussions and comparisons about the algorithm.

\paragraph{Implicit tensor operations:}
In many applications, the input tensor $T$ is not available in advance, and it is computed from samples. It is discussed in~\citep{OvercompleteLVMs2014} that the tensor is not needed to be computed and stored explicitly, where the multilinear tensor updates~\eqref{eqn:asymmetric power update}~and~\eqref{eqn:BiasRemoval} in the algorithm can be efficiently computed through multilinear operations on the samples directly.

\paragraph{Comparison with symmetric orthogonal tensor power method:}
Algorithm~\ref{algo:Power method form} is similar to the symmetric tensor power method
analyzed by~\citet{AnandkumarEtal:tensor12} with the following main differences, \viz
\bi[itemsep=-0.5mm]
\item Symmetric and non-symmetric tensors: Our algorithm can be applied to both symmetric and non-symmetric tensors, while tensor power method in~\citet{AnandkumarEtal:tensor12} is only for symmetric tensors.
%\item Parallelization: As discussed in Remark \ref{remark:parallel implementaion}, our algorithm can be parallelized for recovering different columns, while tensor power method is implemented in a serial deflated manner.

\item Linearity: The updates in Algorithm \ref{algo:Power method form} are linear in each variable, while the symmetric tensor power update is a   quadratic operator given a third order tensor.

\item Guarantees: In~\citet{AnandkumarEtal:tensor12}, guarantees for the symmetric tensor power update under orthogonality  are obtained, while here we consider non-orthogonal tensors under the alternating updates.
%\item Computational complexity:
%\item \mjcomment{Stabilizing is crucial for getting better convergence in power method, while our method does not need such stabilization.}
\ei

\paragraph{Comparison with Alternating Least Square(ALS):}
The updates in Algorithm \ref{algo:Power method form} can be viewed as a rank-$1$ form of the standard alternating least squares (ALS) procedure. This is because   the unnormalized update for $c$ in \eqref{eqn:asymmetric power update} can be rewritten as
\begin{align}
\tl{c}_\tau^{(t+1)}
 := T \left( \ha_\tau^{(t)}, \hb_\tau^{(t)},I \right)
%& = C \Diag(w) \left( A^\top \ha_\tau^{(t)} * B^\top \hb_\tau^{(t)} \right) \nn \\
 = \mat(T,3) \cdot \left( \hb_\tau^{(t)}\odot \ha_\tau^{(t)} \right), \label{eqn:approx update-C}
\end{align}
where $\odot$ denotes the {\em Khatri-Rao} product, and $\mat(T,3) \in \R^{d \times d^2}$ is the mode-$3$ matricization of tensor $T$. On the other hand, the ALS update has the form
%\begin{align*}
%\tl{C}^{(t+1)}&=\mat(T,3) \cdot \left( \hB^{(t)}\odot \hA^{(t)} \right),
%\end{align*}
$$
\tl{C}^{(t+1)} =\mat(T,3) \cdot \left( \left( \hB^{(t)}\odot \hA^{(t)} \right)^\top \right)^\dagger,
$$
where $k$ vectors (all columns of $\tl{C}^{(t+1)} \in \R^{d \times k}$) are simultaneously updated given the current estimates for the other two modes $\hA^{(t)}$ and $\hB^{(t)}$. In contrast, our procedure updates only one vector (with the target of recovering one column of $C$) in each iteration. In our update, we do not require finding matrix inverses. This leads to efficient computational complexity, and we also show that our update procedure is more robust to perturbations.

\section{Analysis} \label{sec:analysis}

In this section, we provide the local and global convergence guarantees for the tensor decomposition algorithm proposed in Section~\ref{sec:algorithm}.
Throughout the paper, we assume tensor $\hT \in \Rbb^{d \times d \times d}$ is of the form $\hT = T + \Psi$, where $\Psi$ is the error or perturbation tensor, and\footnote{For 4th and higher order tensors, same techniques we introduce in this paper, can be exploited to argue similar results.}
$$
T = \sum_{i\in [k]} w_i \cdot a_i\otimes b_i\otimes c_i,
$$
is a rank-$k$ tensor such that $a_i,b_i,c_i \in \Rbb^d, i \in [k],$ are unit vectors. Let $A := [a_1 \ a_2 \ \dotsb \ a_k] \in \Rbb^{d \times k}$, and $B$ and $C$ are similarly defined.
The goal of robust tensor decomposition algorithm is to recover the rank-1 components $\{(a_i, b_i, c_i), i \in [k]\}$ given noisy tensor $\hT$. Our analysis emphasizes on the challenging {\em overcomplete} regime where the tensor rank is larger than the dimension, i.e., $k>d$.
Without loss of generality we also assume $w_{\max} = w_1 \ge w_2\ge \cdots\ge w_k = w_{\min} > 0$. 

We require natural deterministic conditions on the tensor components to argue the convergence guarantees; see Appendix~\ref{sec:assumptions} for the details. We show that all of these conditions are satisfied if the true rank-1 components of the tensor are uniformly  i.i.d.\ drawn from the unit $d$-dimensional sphere $\Sc^{d-1}$. Thus, for simplicity we assume this random assumption in the main part, and state the deterministic assumptions in Appendix \ref{sec:assumptions}. Notice that it is also reasonable to assume these deterministic assumptions hold for some non-random matrices. Among the deterministic assumptions, the most important one is the {\em incoherence} condition which imposes a soft-orthogonality constraint between different rank-1 components of the tensor.

%\subsubsection*{Basic definitions}
The convergence guarantees are provided in terms of distance between the  estimated and the true vectors, defined below.
%\begin{definition}
%For any two vectors $u, v \in \R^d$, the distance between them is defined as
%the $\sin$ of their angle. Equivalently, for two unit vectors
%\begin{align} \label{eqn:dist function definition}
%%\dist(u,v) := \sup_{z \perp u} \frac{\langle z,v \rangle}{\| z \| \cdot \| v \|}
%%= \sup_{z \perp v} \frac{\langle z,u \rangle}{\| z \| \cdot \| u \|}.
%\dist(u,v) := \|u-\langle u,v\rangle v\| %\sup_{z \perp u} \frac{\langle z,v \rangle}{\| z \| \cdot \| v \|}
%= \|v-\langle u,v\rangle u\| %\sup_{z \perp v} \frac{\langle z,u \rangle}{\| z \| \cdot \| u \|}.
%\end{align}
%\end{definition}
\begin{definition}
For any two vectors $u, v \in \R^d$, the distance between them is defined as
\begin{align} \label{eqn:dist function definition}
\dist(u,v) := \sup_{z \perp u} \frac{\langle z,v \rangle}{\| z \| \cdot \| v \|}
= \sup_{z \perp v} \frac{\langle z,u \rangle}{\| z \| \cdot \| u \|}.
\end{align}
\end{definition}

Note that distance function $\dist(u,v)$ is invariant w.r.t. norm of input vectors $u$ and $v$. Distance also provides an upper bound on the error between unit vectors $u$ and $v$ as (see Lemma A.1 of \citet{AgarwalEtal:SparseCoding2013})
$$
\min_{z \in \{-1,1\}} \|zu-v \| \leq \sqrt{2} \dist(u,v).
$$
Incorporating distance notion resolves the sign ambiguity issue in recovering the components: note that  a third order tensor is unchanged if the sign of a vector along one of the modes is fixed and the signs of  the corresponding vectors in the other two modes are flipped.

%Let  $\psi := \|\Psi\|$ denote the spectral norm of error tensor $\Psi$, and
%\begin{align} \label{eqn:target error}
%\epsilon_R := \frac{\psi}{w_{\min}} + \tl{O} \left( \gamma \frac{\sqrt{k}}{d} \right),
%\end{align}
%denote the recovery error where $\gamma := \frac{w_{\max}}{w_{\min}}$. 

\subsection{Local convergence guarantee} \label{sec:local convergence}

%The local convergence result is provided in the following theorem which bounds the estimation error after $N$ iterations of the Algorithm. Note that a good initialization is assumed in the local convergence guarantee and the behavior of asymmetric power update in the inner loop of Algorithm~\ref{algo:Power method form} is analyzed.

In the local convergence guarantee, we analyze the convergence properties of the algorithm assuming we have good initialization vectors for the non-convex tensor decomposition algorithm.

\paragraph{Settings of Algorithm in Theorem~\ref{thm:local convergence}:}
\bi[itemsep=-1mm]
\item Number of iterations: $N = \Theta \left( \log \left( \frac{1}{\gamma \epsilon_R} \right) \right)$, where $\gamma := \frac{w_{\max}}{w_{\min}}$ and $\epsilon_R := \min \left\{ \frac{\psi}{w_{\min}}, \tl{O} \left( \gamma \frac{\sqrt{k}}{d} \right) \right\}$.
\ei

\paragraph{Conditions for Theorem~\ref{thm:local convergence}:}
\bi[itemsep=-1mm]
\item Rank-$k$ true tensor with random components: Let
$$
T= \sum_{i\in [k]} \wstar_i \cdot \astar_i \otimes \bstar_i \otimes \cstar_i,\quad w_i>0, \astar_i, \bstar_i, \cstar_i \in \Sc^{d-1},
$$
where $a_i,b_i,c_i, i \in [k],$ are uniformly  i.i.d.\ drawn from the unit $d$-dimensional sphere $\Sc^{d-1}$. We state the deterministic assumptions in Appendix \ref{sec:assumptions}, and show that random matrices satisfy these assumptions.
\item Rank condition: $k = o \left( d^{1.5} \right).$
\item Perturbation tensor $\Psi$ satisfies the bound
$$\psi := \|\Psi\| \le \frac{w_{\min}}{6}.$$
\item Weight ratio: The maximum ratio of weights $\gamma := \frac{w_{\max}}{w_{\min}}$ satisfies the bound
$$
\gamma = O \left( \min \left\{ \sqrt{d}, \frac{d^{1.5}}{k} \right\} \right).
$$
\item Initialization: Assume we have good initialization vectors $\ha^{(0)}_j, \hb^{(0)}_j , j \in [k]$ satisfying
%The following initialization bound holds for some $j \in [k]$ as
\begin{align} \label{eqn:good init}
\epsilon_0 := \max \left\{ \dist \left(\ha^{(0)}_j, a_j\right), \dist \left(\hb^{(0)}_j, b_j \right) \right\}
= O (1/\gamma), \quad \forall j \in [k],
\end{align}
where $\gamma := \frac{w_{\max}}{w_{\min}}$. In addition, given $\ha^{(0)}_j$ and $\hb^{(0)}_j$, suppose $\hc^{(0)}_j$ is also calculated by the update formula in \eqref{eqn:asymmetric power update}.
\ei

\begin{theorem}[Local convergence guarantee of the tensor decomposition algorithm]
 \label{thm:local convergence}
Consider noisy rank-$k$ tensor $\hT = T + \Psi$ as the input to the tensor decomposition algorithm, and assume the conditions and settings mentioned above hold.
Then the algorithm outputs estimates $\hA:= [\ha_1 \dotsb \ha_k] \in \R^{d \times k}$ and $\hw := [\hw_1 \dotsb \hw_k]^\top \in \R^k$, satisfying w.h.p.\
\begin{equation*}
\left\| \widehat{A} - A \right\|_F \leq \tl{O} \left( \frac{\sqrt{k} \cdot \psi}{w_{\min}} \right), \quad
\left\| \hw - w \right\| \leq %w_{\min} \epsilon_R, \quad
\tl{O} \left( \sqrt{k} \cdot \psi \right).
\end{equation*}
Same error bounds hold for other factor matrices $B := [b_1 \dotsb b_k]$ and $C := [c_1 \dotsb c_k]$. 
\end{theorem}
See the proof in Appendix \ref{sec:convergence proof}.

Thus, we can efficiently decompose the tensor in the highly overcomplete regime $k \leq o \left( d^{1.5} \right)$ under incoherent factors and some other assumptions mentioned above. The deterministic version of assumptions are stated in Appendix~\ref{sec:assumptions}. We show that these assumptions are true for random components which is assumed here for simplicity. If $k$ is significantly smaller than $d^{1.5}$ ($k\ll d^{1.25}$), then many of the assumptions can be derived from incoherence. See Appendix~\ref{sec:assumptions} for the details.
%Note that our recovery is in terms of distance between any true vector $a_j$ (or $b_j$, $c_j$) and the estimate $\hat{a}^{(N)}$ (or $\hb^{(N)}$, $\hc^{(N)}$).

The above local convergence result can be also interpreted as a local identifiability result for tensor decomposition under incoherent factors.

The $\sqrt{k}$ factor in the above theorem error bound is from the fact that the final recovery guarantee is on the Frobenius norm of the whole factor matrix $A$.
In the following, we provide stronger column-wise guarantees (where there is no $\sqrt{k}$ factor) with the expense of having an additional residual error term.
%{\bf Tensor power iteration guarantees:}
Recall that our algorithm includes two main update steps including tensor power iteration in~\eqref{eqn:asymmetric power update} and residual error removal in~\eqref{eqn:BiasRemoval}. %We discussed in the previous section that the tensor power iteration step recovers the components up to some residual error (in addition to the noise term which is unavoidable).
The guarantee for the first step --- tensor power iteration --- is provided in the following lemma.

\begin{lemma}[Local convergence guarantee of the tensor power updates, Algorithm~\ref{algo:Power method form}]
\label{thm:local convergence-poweriteration}
Consider the same settings as in Theorem~\ref{thm:local convergence}. Then, the outputs of tensor power iteration steps (output of Algorithm~\ref{algo:Power method form}) satisfy w.h.p.%\footnote{Note that recovery of components is up to sign. This is because a third order tensor is unchanged if the sign along one of the modes is fixed and the signs along the other two modes are flipped.}
\begin{equation*}
\dist(\widehat{a}_j, a_j) \leq \tl{O} \left( \frac{\psi}{w_{\min}} \right) + \tl{O} \left( \gamma \frac{\sqrt{k}}{d} \right), \quad
\left| \hw_j - w_j \right| \leq %w_{\min} \epsilon_R, \quad
\tl{O} \left( \psi \right) + \tl{O} \left( w_{\max} \frac{\sqrt{k}}{d} \right), \quad
j \in [k].
\end{equation*}
Same error bounds hold for other factor matrices $B$ and $C$.
\end{lemma}
The above result provides guarantees with the additional residual error $\tl{O} \left( \gamma \frac{\sqrt{k}}{d} \right)$, but we believe this result also has independent importance for the following reasons.
%The above local convergence result can be also interpreted as an approximate local identifiability result for tensor power iteration under incoherent factors.
The above result provides column-wise guarantees which is stronger than the guarantees on the whole factor matrix in Theorem~\ref{thm:local convergence}. Furthermore, we can only have recovery guarantees for a subset of rank-1 components of the tensor (the ones for which we have good initializations) without worrying about the rest of components. Finally, in the high-dimensional regime (large $d$), the residual error term goes to zero.

%Note that the recovery error $\epsilon_R$ arises due to perturbation tensor $\Psi$ (given by $\frac{\psi}{w_{\min}} $) which is unavoidable, and non-orthogonality (given by $\tl{O} \bigl( \gamma \sqrt{k}/d \bigr)$). Thus, there is an approximation error in recovery of the tensor components. The above local convergence result can be also interpreted as an approximate local identifiability result for tensor decomposition under incoherent factors.

The result in the above lemma is actually stated in the non-asymptotic form, where the details of constants are explicitly provided in  Appendix \ref{sec:assumptions}.

\paragraph{Symmetric tensor decomposition:} The above local convergence result also holds for recovering the components of a rank-$k$ {\em symmetric} tensor. Consider symmetric tensor $T$ with CP decomposition $T = \sum_{i \in [k]} w_i a_i \otimes a_i \otimes a_i$. The proposed algorithm can be also applied to recover the components $a_i, i \in [k],$ where the main updates are changed to adapt to the symmetric tensor. The tensor power iteration is changed to
\begin{align} \label{eqn:symmetric power update}
\ha^{(t+1)} = \frac{T \left( \ha^{(t)}, \ha^{(t)}, I \right)}{\left\| T \left( \ha^{(t)}, \ha^{(t)}, I \right) \right\|},
\end{align}
and the coordinate descent update is changed to the form stated in~\eqref{eqn:BiasRemoval-symmetric}.
Then, the same local convergence result as in Theorem \ref{thm:local convergence} holds for this algorithm. The proof is very similar to the proof of Theorem \ref{thm:local convergence} with some slight modifications considering the symmetric structure.

\paragraph{Extension to higher order tensors:} 
We also provide the generalization of the tensor decomposition guarantees to higher order tensors. We state and prove the result for the tensor power iteration part in details, while the generalization of coordinate descent part (for removing the residual error) to higher order tensors, can be argued by the same techniques we introduce in this paper

For brevity, Algorithm \ref{algo:Power method form} and local convergence guarantee in Lemma \ref{thm:local convergence-poweriteration} are provided for a $3$rd order tensor. The algorithm can be simply extended to higher order tensors to compute the corresponding CP decomposition. %Here, we provide the result for $4$th order tensors.
Consider $p$-th order tensor $T \in \bigotimes^p \Rbb^d$ with CP decomposition
\begin{align} \label{eqn:4th order CP decomp}
T = \sum_{i \in [k]} w_i \cdot a_{(1),i} \otimes a_{(2),i} \otimes \dotsb \otimes a_{(p),i},
\end{align}
where $a_{(r),i} \in \R^d$ is the $i$-th column of $r$-th component $A_{(r)} := \left[a_{(r),1} \ a_{(r),2} \ \dotsb \ a_{(r),k} \right] \in \Rbb^{d \times k},$ for $r \in [p]$.
Algorithm \ref{algo:Power method form} can be extended to recover the components of above decomposition where update formula for the $p$-th mode is modified as
\begin{align} \label{eqn:asymmetric power update 4th}
\ha_{(p)}^{(t+1)} = \frac{T \left( \ha_{(1)}^{(t)}, \ha_{(2)}^{(t)}, \dotsc, \ha_{(p-1)}^{(t)}, I \right)}{\left\| T \left( \ha_{(1)}^{(t)}, \ha_{(2)}^{(t)}, \dotsc, \ha_{(p-1)}^{(t)}, I \right) \right\|},
\end{align}
and similarly the other updates are changed. 
Then, we have the following generalization of Lemma~\ref{thm:local convergence-poweriteration} to higher order tensors.

%Define the recovery error as (generalization of $3$rd order case in \eqref{eqn:target error})
%\begin{align} \label{eqn:target error 4th}
%\tl{\epsilon}_R := \frac{\psi}{w_{\min}} + \tl{O} \left( \gamma \sqrt{\frac{k}{d^{p-1}}} \right).
%\end{align}

\begin{corollary}[Local convergence guarantee of the tensor power updates in Algorithm~\ref{algo:Power method form} for $p$-th order tensor]  \label{thm:local convergence 4th}
Consider the same conditions and settings as in Lemma~\ref{thm:local convergence-poweriteration}, unless tensor $T$ is $p$-th order with CP decomposition in \eqref{eqn:4th order CP decomp} where $p\geq 3$ is a constant. In addition, the bounds on $\gamma : =\frac{w_{\max}}{w_{\min}}$ and $k$ are modified as
\begin{equation*}
\gamma = O \left( \min \left\{ d^{\frac{p-2}{2}}, \frac{d^{p/2}}{k} \right\} \right), \quad k = o \left( d^{\frac{p}{2}} \right).
\end{equation*}
Then, the outputs of tensor power iteration steps (output of Algorithm~\ref{algo:Power method form}) satisfy w.h.p.
\begin{equation*}
\dist \left( \ha_{(r),j}, a_{(r),j} \right) \leq \tl{O} \left( \frac{\psi}{w_{\min}} \right) + \tl{O} \left( \gamma \sqrt{\frac{k}{d^{p-1}}} \right), \quad
\left| \hw_j - w_j \right| \leq \tl{O} \left( \psi \right) + \tl{O} \left( w_{\max} \sqrt{\frac{k}{d^{p-1}}} \right),
\end{equation*}
for $j \in [k]$ and $r \in [p]$.
The number of iterations is $N = \Theta \left( \log \left( \frac{1}{\gamma \tl{\epsilon}_R} \right) \right)$, where $\gamma := \frac{w_{\max}}{w_{\min}}$ and $\tl{\epsilon}_R := \min \left\{ \frac{\psi}{w_{\min}}, \tl{O} \left( \gamma \sqrt{k/d^{p-1}} \right) \right\}$.
\end{corollary}

\subsection{Global convergence guarantee when $k=O(d)$} \label{sec:global convergence}

Theorem \ref{thm:local convergence} provides local convergence guarantee given good initialization vectors.
In this section, we exploit SVD-based initialization method in Procedure~\ref{algo:SVD init} to provide good initialization vectors when $k = O(d)$. This method proposes the top singular vectors of random slices of the moment tensor as the initialization. Combining the theoretical guarantees of this initialization method (provided in Appendix \ref{sec:initialization}) with the local convergence guarantee in Theorem \ref{thm:local convergence}, we provide the following global convergence result.

\paragraph{Settings of Algorithm in Theorem~\ref{thm:global convergence}:}
\bi[itemsep=-1mm]
\item Number of iterations: $N = \Theta \left( \log \left( \frac{1}{\gamma \epsilon_R} \right) \right)$, where $\gamma := \frac{w_{\max}}{w_{\min}}$ and $\epsilon_R := \min \left\{ \frac{\psi}{w_{\min}}, \tl{O} \left( \gamma \frac{\sqrt{k}}{d} \right) \right\}$.
\item The initialization in each run of Algorithm\ \ref{algo:Power method form} is performed by SVD-based technique proposed in Procedure~\ref{algo:SVD init}, with the number of initializations as
$$L  \geq k^{\Omega \left( \gamma^4 \left( k/d \right)^2 \right)}.$$
\ei

\paragraph{Conditions for Theorem~\ref{thm:global convergence}:}
\bi[itemsep=-1mm]
\item Rank-$k$ decomposition and perturbation conditions as\,\footnote{Note that the perturbation condition is stricter than the corresponding condition in the local convergence guarantee (Theorem~\ref{thm:local convergence}).}
$$
T = \sum_{i\in[k]} w_i \cdot a_i\otimes b_i\otimes c_i, \quad \psi := \|\Psi\| \le \frac{w_{\min} \sqrt{\log k}}{\alpha_0 \sqrt{d}},
$$
where $a_i,b_i,c_i, i \in [k],$ are uniformly i.i.d.\ drawn from the unit $d$-dimensional sphere $\Sc^{d-1}$, and $\alpha_0 >1$ is a constant.
\item Rank condition: $k = O(d)$, i.e., $k\le \beta d$ for arbitrary constant $\beta>1$.
\ei

\begin{theorem}[Global convergence guarantee of tensor decomposition algorithm when $k=O(d)$] \label{thm:global convergence}
Consider noisy rank-$k$ tensor $\hT = T + \Psi$ as the input to the tensor decomposition algorithm, and assume the conditions and settings mentioned above hold.
Then, the same guarantees as in Theorem~\ref{thm:local convergence} hold.
\end{theorem}
See the proof in Appendix \ref{sec:convergence proof}.

Thus, we can efficiently recover the tensor decomposition, when the tensor is undercomplete or mildly overcomplete (i.e., $k\le \beta d$ for arbitrary constant $\beta>1$), by initializing the algorithm with a simple SVD-based technique.
The number of initialization trials $L$ is polynomial when $\gamma$ is a constant, and $k =O(d)$.

Note that the argument in Lemma~\ref{thm:local convergence-poweriteration} can be similarly adapted leading to global convergence guarantee of the tensor power iteration step.

%Here, we focus on initialization result. It is useful to compare the effect of noise on SVD initialization proposed in this paper, and the random initialization used in~\cite{AnandkumarEtal:tensor12} for orthogonal tensor decomposition. Here, from Assumption \ref{cond:perturbation bound}, we need $\| \Psi \|_F \leq \frac{\sqrt{\log k}}{\polylog(d)} w_{\min}$, and for random initialization, $\| \Psi \| \leq \frac{w_{\min}}{k}$ is required which results in $\| \Psi \|_F \leq \frac{w_{\min}}{\sqrt{k}}$. Comparing two bounds, we observe that SVD initialization method in Procedure~\ref{algo:SVD init} is more robust to noise with a factor of $\sqrt{k}$. This is basically from the fact that in the undercomplete setting, we can get constant error bound with polynomial number of initializations using SVD-based method, while for random initialization, we can not get constant error bound with polynomial number of initializations.
%\rgcomment{This paragraph is not correct, because $\|\Psi\|$ is a tensor, not a matrix. The spectral norm and Frobenius norm of a tensor can differ by at most $O(d)$ instead of $O(\sqrt{d})$ (in particular it has no correlation with $k$ unless you try to project the tensor, even then it is $O(k)$ instead of $O(\sqrt{k})$). From the fact that $\|\Psi\| \le w_{min}/k$ it only implies $\|\Psi\|_F \le w_{min}$. I suggest we simply remove this paragraph.}

\subsubsection*{Two undercomplete, and one overcomplete component}
Here, we apply the global convergence result to the regime of two undercomplete and one overcomplete components. This arises in supervised learning problems under a multiview mixtures model and employing moment tensor $\Ebb[x_1\otimes x_2\otimes y]$, where $x_i\in \Rbb^{d_u}$ are multi-view high-dimensional features and $y\in \Rbb^{d_o}$ is a low-dimensional label.

Since in the SVD initialization Procedure~\ref{algo:SVD init}, two components $\ha^{(0)}$ and $\hb^{(0)}$ are initialized through SVD, and the third component $\hc^{(0)}$ is initialized through update formula \eqref{eqn:asymmetric power update}, we can generalize the global convergence result in Theorem \ref{thm:global convergence} to the setting where $A$, $B$ are undercomplete, and $C$ is overcomplete. %See the following corollary.

\begin{corollary} \label{corollary:TwoUnder OneOver}
Consider the same setting as in Theorem \ref{thm:global convergence}. In addition, suppose the regime of undercomplete components $A \in \Rbb^{d_u \times k}$, $B \in \Rbb^{d_u \times k}$, and overcomplete component $C \in \Rbb^{d_o \times k}$ such that $d_u \geq k \geq d_o$. In addition, in this case the bound on $\gamma := \frac{w_{\max}}{w_{\min}}$ is
$$
\gamma = O \left( \min \left\{ \sqrt{d_o}, \frac{d_u \sqrt{d_o}}{k} \right\} \right).
$$
Then, if $k=O(d_u)$ and $d_o \geq \polylog(k)$, the same convergence guarantee as in Theorem \ref{thm:global convergence} holds. %with the new recovery error bound as $\epsilon_R := \frac{\psi}{w_{\min}} + \tl{O} \left( \gamma \sqrt{\frac{k}{d_o d_u}} \right)$.
\end{corollary}
See the proof in Appendix \ref{sec:convergence proof}.

We observe that given undercomplete modes $A$ and $B$, mode $C$ can be arbitrarily overcomplete,  and we can still provide global  recovery of $A,B$ and $C$ by employing  SVD initialization procedure along modes $A$ and $B$.

%\begin{remark}
%When the two undercomplete modes $A$ and $B$ have orthogonal columns, then the constraint $k =O(\sqrt{d_u d_o})$ in the above theorem can be further relaxed. It now suffices to have  $k = O(d_u)$, for any $d_o $. %\aacomment{does $d_o$ need to be polylog in this case?}.
%This is because under orthogonality, the SVD initialization provides a much better initialization than under non-orthogonal components; also, the iterations has less perturbation.
%\end{remark}

%\begin{remark}
%Consider the regime of undercomplete components $A \in \Rbb^{d_u \times k}$ and $B \in \Rbb^{d_u \times k}$, and overcomplete component $C \in \Rbb^{d_o \times k}$ such that $d_u \geq k \geq d_o$. We can still have contraction with polynomial number of draws for initialization as long as $k=O(\sqrt{d_u d_o})$. It can be shown as follows. From Lemma \ref{lem:local convergence}, we need to start with $\epsilon_0 = O (\sqrt{d_o/k})$ for contraction. Therefore, in initialization Lemma \ref{lem:initialization}, we need $\frac{\mu_E}{\tl{\mu}} = O (\sqrt{d_o/k})$, and since $\mu_E = \Theta(\sqrt{k/d_u})$, we need $\tl{\mu} = \Omega(k / \sqrt{d_u d_o})$. As long as $k=O(\sqrt{d_u d_o})$, we need $\tl{\mu}=\Omega(1)$ for which the initialization can be done by polynomial number of draws.
%\end{remark}

\subsection{Proof outline}
The global convergence guarantee in Theorem \ref{thm:global convergence} is established by combining the local convergence result in Theorem \ref{thm:local convergence} and the SVD initialization result   in Appendix \ref{sec:initialization}.

The local convergence result in Theorem~\ref{thm:local convergence} is derived by establishing error contraction in each iteration of the tensor power iteration and the coordinate descent for removing the residual error. Note that these convergence properties are broken down in Lemmata~\ref{thm:local convergence-poweriteration}~and~\ref{thm:coordinate descent}, respectively.
%Looking at Theorem \ref{thm:local convergence}, the linear convergence term as $q^t \epsilon_0$ for some contraction factor $q<1$ stands for this one-step contraction argument after $t$ iterations.

Since we assume generic factor matrices $A,B$ and $C$, we utilize many useful properties such as incoherence, bounded spectral norm of the matrices $A,B$ and $C$, bounded tensor spectral norm and so on. We list the precise set of deterministic conditions required to establish the local convergence result in Appendix~\ref{sec:assumptions}. Under these conditions,  with a  good initialization (i.e., small enough $\max \{\dist(\ha,a_j), \dist(\hb,b_j)\} \leq \epsilon_0$), we show that the iterative update in \eqref{eqn:asymmetric power update} provides an estimate $\hc$ with 
$$\dist(\hc,c_j)< \tl{O} \left( \frac{\psi}{w_{\min}} \right) + \tl{O} \left( \gamma \frac{\sqrt{k}}{d} \right) + q \epsilon_0,$$
for some contraction factor $q<1/2$. The incoherence condition is crucial for establishing  this result. See Appendix \ref{sec:convergence proof} for the complete proof.

The initialization argument for SVD-based technique in Procedure~\ref{algo:SVD init} has two parts. The first part claims that by performing enough number of initializations (large enough $L$), a gap condition is satisfied, meaning that we obtain  a vector $\theta$ which is relatively close  to $c_j$ compared to any $c_i,i \neq j$. This is a standard result for Gaussian vectors, e.g., see Lemma B.1 of~\citet{AnandkumarEtal:tensor12}. In the second part of the argument, we analyze the  dominant singular vectors of $T(I,I,\theta)$, for a vector $\theta$ with a good relative gap, to obtain an error bound on the initialization   vectors. This is obtained   through standard matrix perturbation results (Weyl and Wedin's theorems). See Appendix \ref{sec:initialization} for the complete proof.

\section{Experiments}

%\subsection{Synthetic experiments}
In this section, we provide some synthetic experiments to evaluate the performance of Algorithm \ref{algo:Power method form}. Note that tensor power update in Algorithm~\ref{algo:Power method form} is the main step of our algorithm which is considered in this experiment.
A random true tensor $T$ is generated as follows. First, three components $A \in \Rbb^{d \times k}$, $B \in \Rbb^{d \times k}$, and $C \in \Rbb^{d \times k}$ are randomly generated with i.i.d standard Gaussian entries. Then, the columns of these matrices are normalized where the normalization factors are aggregated as coefficients $w_j, j \in [k]$. From decomposition form in \eqref{eqn:tensordecomp}, tensor $T$ is built through these random components. For each new initialization, $\ha^{(0)}$ and $\hb^{(0)}$ are randomly generated with i.i.d.\ standard Gaussian entries, and then normalized\,\footnote{Drawing i.i.d. standard Gaussian entries and normalizing them is equivalent to drawing  vectors uniformly from the $d$-dimensional unit sphere.}. Initialization vector $\hc^{(0)}$ is  generated through update formula in \eqref{eqn:asymmetric power update}.

For each initialization $\tau \in [L]$, an alternative option of running the algorithm with a fixed number of iterations $N$ is to stop the iterations based on some stopping criteria. In this experiment, we stop the iterations when the improvement in subsequent steps is small as
$$
\max \left( \left\| \ha_\tau^{(t)} - \ha_\tau^{(t-1)} \right\|^2, \left\| \hb_\tau^{(t)} - \hb_\tau^{(t-1)} \right\|^2, \left\| \hc_\tau^{(t)} - \hc_\tau^{(t-1)} \right\|^2  \right) \leq t_{\Stopping},
$$
where $t_{\Stopping}$ is the stopping threshold. According to the  bound in Theorem \ref{thm:local convergence}, we set
\begin{align} \label{eqn:stopping threshold}
t_{\Stopping} := t_1 (\log d)^2 \frac{\sqrt{k}}{d},
%t_{\Recovery} := t_1 \max \left( (\log d)^2 \frac{\sqrt{k}}{d}, (\log d)^3 \frac{k}{d^{1.5}} \right),
\end{align}
for some constant $t_1>0$.

\subsubsection*{Effect of size $d$ and $k$}
\begin{figure}\centering{
\bp
\psfrag{d eq 1000, k eq 10}[l]{\tiny $d\!=\!1000, k\!=\!10$}
\psfrag{d eq 1000, k eq 50}[l]{\tiny $d\!=\!1000, k\!=\!50$}
\psfrag{d eq 1000, k eq 100}[l]{\tiny $d\!=\!1000, k\!=\!100$}
\psfrag{d eq 1000, k eq 200}[l]{\tiny $d\!=\!1000, k=\!200$}
\psfrag{d eq 1000, k eq 500}[l]{\tiny $d\!=\!1000, k\!=\!500$}
\psfrag{d eq 1000, k eq 1000}[l]{\tiny $d\!=\!1000, k\!=\!1000$}
\psfrag{d eq 1000, k eq 2000}[l]{\tiny $d\!=\!1000, k\!=\!2000$}
\psfrag{d eq 1000, k eq 5000}[l]{\tiny $d\!=\!1000, k\!=\!5000$}
\psfrag{recovery rate of algorithm}[l]{\scriptsize recovery rate of algorithm}
\psfrag{number of initializations}[l]{\scriptsize number of initializations}
\psfrag{ratio of recovered columns}[l]{\scriptsize ratio of recovered columns}
\includegraphics[width=3.0in]{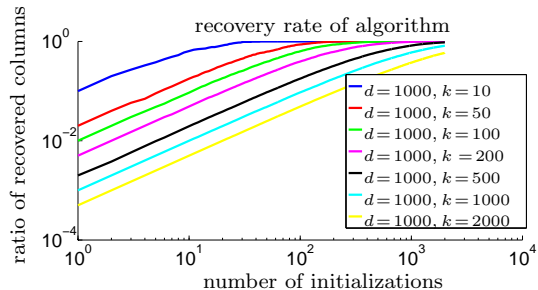}\ep}
%\subcaption{\small Both x and y axes are log scaled.} \label{fig:RecoveryRatio vs Init_LogXLogY}
\caption{\small Ratio of recovered columns versus the number of initializations for $d=1000$, and $k= \{ 10, 50, 100, 200, 500, 1000, 2000 \}$. The number of initializations is $L=2000$. The stopping parameter is set to $t_1 = 1e-08$. The figure is an average over 10 random runs.}
\label{fig:RecoveryRatio vs Init}
\end{figure}

Algorithm \ref{algo:Power method form} is applied to random tensors with $d=1000$ and $k= \{ 10, 50, 100, 200, 500, 1000, 2000 \}$. The number of initializations is $L=2000$. The parameter $t_1$ in \eqref{eqn:stopping threshold} is fixed as $t_1 = 1e-08$. Figure \ref{fig:RecoveryRatio vs Init} and Table \ref{table:RecoveryRatio vs Init} illustrate the outputs of running experiments which is the average of 10 random runs.

Figure \ref{fig:RecoveryRatio vs Init} depicts the ratio of recovered columns versus the number of initializations. Both horizontal and vertical axes are plotted in $\log$-scale. We observe that it is much easier to recover the columns in the undercomplete settings ($k \leq d$), while it becomes harder when $k$ increases. Linear start in Figure \ref{fig:RecoveryRatio vs Init} suggests that recovering the first bunch of columns only needs polynomial number of initializations. For highly undercomplete settings like $d=1000$ and $k=10$, almost all columns are recovered in this linear phase. After this start, the concave part means that it needs many more initializations for recovering the next bunch of columns. As we go ahead, it becomes harder to recover true columns, which is intuitive.

%This is because of two reasons. First, many recovered columns are already recovered, and it is hard to recover remaining ones, and we need very close initializations to them for successful recovery. Second, the overcomplete case makes it even harder and we need more initializations for that.
%Figure \ref{fig:RecoveryRatio vs Init_Regular} depicts the ratio of recovered columns versus the number of initializations. we observe that it is much easier to recover the columns in the undercomplete settings ($k \leq d$), while it becomes harder when $k$ increases. There are interesting observations in rescaled Figures \ref{fig:RecoveryRatio vs Init_LogX} and \ref{fig:RecoveryRatio vs Init_LogXLogY}. Linear start in Figure \ref{fig:RecoveryRatio vs Init_LogXLogY} suggests that recovering the first bunch of columns only needs polynomial number of initializations. For highly undercomplete settings like $d=100$ and $k=10$, almost all columns are recovered in this linear area. After this start, the linear part in the middle of Figure \ref{fig:RecoveryRatio vs Init_LogX} suggests that we need exponential number of initializations for recovering the next bunch of columns. And, at the end, it even becomes harder. This is because of two reasons. First, as we go ahead, many recovered columns are already recovered, and it is hard to recover remaining ones, and we need very close initializations to them for successful recovery. Second, the overcomplete case makes it even harder and we need more initializations for that.

%\begin{figure}
%\centering\begin{minipage}[b]{3.0in}
%\bp
%%\psfrag{X}[l]{$Y$}
%\centering \includegraphics[width=3.0in]{figures/RecoveryRatio_vs_Init.eps}\ep
%\subcaption{\small Non-scaled axes.} \label{fig:RecoveryRatio vs Init_Regular}
%\end{minipage}
%\hfil
%\begin{minipage}[b]{3.0in}
%\bp
%%\psfrag{X}[l]{$Y$}
%\centering \includegraphics[width=3.0in]{figures/RecoveryRatio_vs_Init_LogX.eps}\ep
%\subcaption{\small x-axis is log scaled.} \label{fig:RecoveryRatio vs Init_LogX}
%\end{minipage}
%\hfil
%\begin{minipage}[b]{3.0in}
%\bp
%%\psfrag{X}[l]{$Y$}
%\centering \includegraphics[width=3.0in]{figures/RecoveryRatio_vs_Init_LogXLogY.eps}\ep
%\subcaption{\small Both x and y axes are log scaled.} \label{fig:RecoveryRatio vs Init_LogXLogY}
%\end{minipage}
%\caption{\small Ratio of recovered columns versus the number of initializations for fixed $d=100$, and different $k= \{ 10, 20, 50, 100, 200, 500, 1000 \}$. The acceptance and stopping parameters are set to $t_1 = 0.01$ and $t_2 = 10^{-5}$. The figure is an average of 10 random runs.}
%\label{fig:RecoveryRatio vs Init}
%\end{figure}

Table \ref{table:RecoveryRatio vs Init} has the results from the experiments. Parameters $k$, %acceptance threshold $t_{\Recovery}$,
stopping threshold $t_{\Stopping}$, and the average square error of the output, the average weight error and the average number of iterations are stated.
%In order to have a fair comparison, the acceptance threshold is assumed to be the same for all different $k$. We chose the one from equation \eqref{eqn:acceptance threshold} for $d=1000$ and $k=5000$. But, note that the stopping thresholds are  different for distinct $k$ discussed earlier.
The output averages are over several initializations and random runs. The square error is given by
$$
\frac{1}{3} \left[ \left\| a_j - \ha \right\|^2 + \left\| b_j - \hb \right\|^2 + \left\| c_j - \hc \right\|^2 \right],
$$
%the left-hand side of \eqref{eqn:AvgSqError} which is compared with the acceptance threshold to see if the resulting estimates are fair estimates for some true columns.
for the corresponding recovered $j$. The error in estimating the weights is defined as $|\hw - w_j |^2/w_j^2$  which is the square relative error of weight estimate. The number of iterations performed before stopping the algorithm is mentioned in the last column.
We observe that by increasing $k$, all of these outputs are increased which means we get less accurate estimates with higher computation. This shows that recovering the overcomplete components is much harder. Note that by running the coordinate descent Algorithm~\ref{algo:coordinate-descent}, we can also remove this additional residual error left after the tensor power iteration step. 
Similar results and observations as above are seen when $k$ is fixed and $d$ is changed.

%It is worth mentioning that for a fixed $k$, decreasing stopping threshold (smaller $t_2$), results in smaller square error and larger number of iterations. In other words, we get more accurate estimates by doing more iterations when smaller stopping threshold is considered.
Running experiments with SVD initialization instead of random initialization  yields nearly  the same recovery rates, but  with slightly smaller  number of iterations. But, since the SVD computation is more expensive, in practice, it is desirable to initialize with random vectors. Our theoretical results for random initialization appear to be highly pessimistic compared to the efficient recovery results in our experiments. This suggests additional room for improving our theoretical guarantees under random initialization.

\begin{table}
\caption{\small Parameters and more outputs related to results of Figure \ref{fig:RecoveryRatio vs Init}. Note that $d=1000$.} \label{table:RecoveryRatio vs Init}
\bc{\small \begin{tabular}{c|c||c|c|c}
\hline
\multicolumn{2}{c||}{Parameters} & \multicolumn{3}{|c}{Outputs} \\
\hline
$k$ &
\begin{tabular}{c} $t_{\Stopping}$ \end{tabular} &
%\begin{tabular}{c} stopping \\ threshold $t_{\Stopping}$ \end{tabular} &
\begin{tabular}{c} avg. square \\ error \end{tabular} &
\begin{tabular}{c} avg. weight \\ error \end{tabular} &
\begin{tabular}{c} avg. \# of \\ iterations \end{tabular}  \\
\hline \hline
10 &  1.51e-08 & 1.03e-05 & 9.75e-09  & 7.71 \\
%\hline
%20 & 0.0195 & 0.0195 & 0.0023  & 10.5315 & \\
%\hline
50 &  3.37e-08 & 5.54e-05 & 6.69e-08 & 8.53 \\
%\hline
100&  4.77e-08 & 1.08e-04  & 1.51e-07 & 8.81 \\
%\hline
200 &  6.75e-08 & 2.07e-04 & 3.41e-07 & 9.09 \\
%\hline
500  &  1.07e-07 & 5.09e-04 & 1.14e-06 & 9.52 \\
%\hline
1000  & 1.51e-07 & 1.01e-03 & 3.40e-06 & 10.01 \\
%\hline
2000 & 2.13e-07  & 2.00e-03  & 1.12e-05  & 10.69 \\
\hline
%5000 &  &  &  & \\
%\hline
\end{tabular}}\ec
\end{table}

\subsubsection*{Acknowledgements}
We acknowledge detailed discussions with Sham Kakade and Boaz Barak. We thank  Praneeth Netrapalli for discussions on alternating minimization. We also thank Sham Kakade, Boaz Barak,  Jonathan Kelner, Gregory Valiant and Daniel Hsu for earlier discussions on the $2 \to p$ norm bound for random matrices, used in Lemma~\ref{lem:2 to p norm bound}. We also thank Niranjan U.N. for discussions on running experiments.
A. Anandkumar is supported in part by Microsoft Faculty Fellowship, NSF Career award CCF-$1254106$, NSF Award CCF-$1219234$, and ARO YIP Award W$911$NF-$13$-$1$-$0084$. M. Janzamin is supported by  NSF Award CCF-1219234, ARO Award W911NF-12-1-0404  and ARO YIP Award W911NF-13-1-0084.

%Funding

\renewcommand{\appendixpagename}{Appendix}

\appendixpage

\appendix

\section*{More Matrix Notations}

Given vector $w \in \R^d$, let $\diag(w) \in \R^{d \times d}$ denote the diagonal matrix with $w$ on its main diagonal.
Given matrix $A \in \Rbb^{d \times k}$, the following notations are defined to refer to its sub-matrices. $A_j$ denotes the $j$-th column and $A^j$ denotes the $j$-th row of $A$. In addition, $A_{\setminus j} \in \Rbb^{d \times (k-1)}$ is $A$ with its $j$-th column removed, and $A^{\setminus j} \in \Rbb^{(d-1) \times k}$ is $A$ with its $j$-th row removed.

%For $A \in \Rbb^{p \times q}$ and $B \in \Rbb^{m \times n}$, the {\em Kronecker} product\,\footnote{Throughout this paper, notation $\otimes$ is usually used for outer product introduced in \eqref{eqn:rank-1 tensor}. But, with slightly abuse of notation, we also use that for Kronecker product.} $A \otimes B \in \Rbb^{pm \times qn}$ is defined as~\citep{golub2012matrix}
%\begin{align*}
%A \otimes B = \left[
%\begin{array}{cccc}
%a_{11} B & a_{12}B & \dotsb & a_{1q} B \\
%a_{21} B & a_{22}B & \dotsb & a_{2q} B \\
%\vdots & \vdots & \ddots & \vdots \\
%a_{p1} B & a_{p2} B & \dotsb & a_{pq} B
%\end{array} \right].
%\end{align*}
%Thus, for vectors $a,b \in \R^d$, we have $a \otimes b\in \R^{d^2}$ such that
%\[ a \otimes b := \left[ \begin{array}{c} a_{1} b\\ a_{2} b \\ \vdots \\ a_{d}b \end{array}\right].\]

For two matrices $A \in \Rbb^{d_1 \times k}$ and $B\in \Rbb^{d_2 \times k}$,  the {\em Khatri-Rao} product is denoted by $A \odot B \in \Rbb^{d_1d_2 \times k}$, and its $(\bfi, j)^{\tha}$ entry is given by
\[ A\odot B(\bfi, j) := A_{i_1, j} B_{i_2, j}, \quad \bfi= (i_1, i_2)\in [d_1]\times[d_2], j \in [k].\]

For two matrices $A \in \Rbb^{d\times k}$ and $B\in \Rbb^{d\times k}$,  the {\em Hadamard} product is defined as the entry-wise multiplication of the matrices,
\[ A*B(i,j):= A(i,j) B(i,j), \quad i \in [d], j\in [k].\]

%For any $A \in \Rbb^{p \times k}, B \in \Rbb^{q \times k}, C \in \Rbb^{p \times n},$ and $D \in \Rbb^{q \times n}$, we have the following property
%\begin{align} \label{eqn:KR1}
%(A\odot B)^\top (C\odot D)&= (A^\top C) * (B^\top D).
%%(A\odot B)^\dagger&= ((A^\top A) * (B^\top B))^\dagger (A\odot B)^\top %\label{eqn:KR2}
%\end{align}
%We will also use the fact that the spectral norm is sub-multiplicative in the Hadamard product~\cite{horn2012matrix}:
%\beq \label{eqn:submultiplicative} \| A * B\| \leq \|A\| \cdot \|B\| \eeq

Let $\|u\|_p$ denote the $\ell_p$ norm of vector $u$. Let $\|A\|_\infty$ denote the $\ell_\infty$ element-wise norm of matrix $A$, and the induced $q \rightarrow p$ norm is defined as
$$
\|A\|_{q \to p} := \sup_{\|u\|_q = 1} \|A u\|_p.
$$
%Notice that this definition is a bit different from the usual definition where the transpose of $A$ is considered.

%\section{Comparison with Alternating Least Square (ALS)} \label{sec:ALS comparison}

\section{Deterministic Assumptions} \label{sec:assumptions}

In the main text, we assume matrices $A$, $B$, and $C$ are randomly generated. However, we are not using all the properties of randomness. In particular, we only need the following assumptions.

\begin{enumerate}[label={(A\arabic*)}]

\item \label{cond:rank-k} {\bf Rank-$k$ decomposition: }
The third order tensor $T$ has a CP rank of $k\geq 1$ with  decomposition \beq \label{eqn:true-T}T= \sum_{i\in [k]} \wstar_i (\astar_i \otimes \bstar_i \otimes \cstar_i),\quad w_i>0, \astar_i, \bstar_i, \cstar_i \in \Sc^{d-1}, \forall\, i \in [k], \eeq where $\Sc^{d-1}$ denotes the unit $d$-dimensional sphere,   i.e. all the vectors have unit\,\footnote{This normalization is for convenience and the results hold for general case.} $2$-norm as $\|\astar_i\|= \|\bstar_i\|= \|\cstar_i\|=1, i \in [k]$. Furthermore, define $w_{\min} := \min_{i \in [k]} w_i$ and $w_{\max} := \max_{i \in [k]} w_i$.

%\item {\bf Non-degeneracy: }
%Denote $\Astar:=[\astar_1\,\, \astar_2\,\, \ldots \astar_k]$, and similarly for $\Bstar$ and $\Cstar$. We assume that these matrices have full column rank
%\[ \rank(\Astar)= \rank(\Bstar)=\rank(\Cstar)=k.\]

%\item \label{cond:randomness} {\bf Random components: }
%The columns of $A$, $B$, and $C$ are uniformly drawn at random from unit $d$-dimensional sphere $\Sc^{d-1}$.
%The entries of components $A$, $B$, and $C$ are randomly drawn from a standard Gaussian distribution. And, then the columns are normalized such that $\|a_i\|= \|b_i\|= \|c_i\|=1, i \in [k]$.

\item \label{cond:incoherence} {\bf Incoherence: }
The components are incoherent, and let
\beq \label{eqn:incoherence} \rho:= \max_{i \neq j}\{| \inner{\astar_i,\astar_j}|,  |\inner{\bstar_i,\bstar_j}|, |\inner{\cstar_i,\cstar_j}|\} \leq \frac{\alpha}{\sqrt{d}},\eeq
for some $\alpha=\polylog(d)$. In other words, $A^\top A = I + J_A$, $B^\top B = I + J_B$, and $C^\top C = I + J_C$, where $J_A$, $J_B$, and $J_C$, are incoherence matrices with zero diagonal entries. We have $\max \left\{ \|J_A\|_\infty,\|J_B\|_\infty, \|J_C\|_\infty \right\} \leq \rho$ as in \eqref{eqn:incoherence}.

%\item {\bf Separation of weights: }We assume that the weights are sufficiently separated \[ |w_1|> |w_2|>\ldots |w_k|>0\]

%\item {\bf Initialization: }We  initialize ALS using the SVD of the mode-$k$ matricization of the tensor $T$, for each $k \in [3]$. So we have \[ \mat(T,1)= U \Lambda V^\top, \quad A^{(0)} = U,\] and similarly for $B^{(0)}$ and $C^{(0)}$.

\item \label{cond:spectral bound} {\bf Spectral norm conditions: }
The components satisfy spectral norm bound
$$
\max \left\{ \|A\|,\|B\|,\|C\| \right\} \leq 1 + \alpha_0  \sqrt{\frac{k}{d}},
$$
%and under size condition \ref{cond:size}, tensor $T$ satisfies spectral norm bound
%\begin{align*}
%\|T\| \leq \alpha,
%\end{align*}
for some constant $\alpha_0>0$.

\item \label{cond:tensor bounds} {\bf Bounds on tensor norms: }
Tensor $T$ satisfies the bound
\begin{align*}
& \| T \| \leq w_{\max} \alpha_0, \\
& \left\| T_{\setminus j}(a_j,b_j,I) \right\|
:= \biggl\|  \sum_{i \neq j} w_i \langle a_i,a_j \rangle \langle b_i,b_j \rangle c_j \biggr\|
\leq \alpha w_{\max} \frac{\sqrt{k}}{d},
\end{align*}
for some constant $\alpha_0$ and $\alpha = \polylog(d)$.

%\item \label{cond:spectral bound KR} {\bf Spectral Norm Condition on Khatri-Rao Products: }
%The following spectral bound on the Khatri-Rao products is satisfied
%\begin{align*}
%\max \left\{ \|A \odot B\|, \|B \odot C\|, \|A \odot C\| \right\} \leq 1 + \alpha_k \frac{\sqrt{k}}{d},
%%& \max \left\{\left\| A \odot J_B \right\|, \left\| B \odot J_A \right\| \right\} \leq \alpha_J \frac{\sqrt{k}}{d},
%\end{align*}
%for some constant $\alpha_k>0$.

\item \label{cond:size} {\bf Rank constraint: } The rank of the tensor is bounded by $k = o \left( d^{1.5}/\polylog d \right)$. %, i.e., $\frac{k}{d^{1.5}}$ converges to zero in the asymptotic regime.
%\item \label{cond:size} {\bf Size constraint: } Suppose $k = \beta d^\gamma$ for some constants $0 < \gamma < 1.5$, and $\beta>0$. Note that this condition also allows for overcomplete regime $k>d$. %$\beta < \frac{1}{w_{\max}^2 (\alpha+1)^2}$.

\item \label{cond:perturbation bound} {\bf Bounded perturbation: }
Let $\psi$ denote the spectral norm of perturbation tensor as
%the spectral norm (in mode-$r$ matricization) of perturbation tensor $\Psi$ as
\begin{align} \label{eqn:psi definition}
\psi := \| \Psi \|.
%\psi := \max \left\{ \left\| \mat(\Psi,1) \right\| , \left\| \mat(\Psi,2) \right\| , \left\| \mat(\Psi,3) \right\| \right\}.
\end{align}
Suppose $\psi$ is bounded as\,\footnote{Note that for the local convergence guarantee, only the first condition $\psi \leq \frac{w_{\min}}{6}$ is required.}
$$
\psi \leq \min \left\{ \frac{1}{6}, \frac{\sqrt{\log k}}{\alpha_0 \sqrt{d}} \right\} \cdot w_{\min},
$$
where $\alpha_0$ is a constant.

\item \label{cond:weight ratio} {\bf Weights ratio: } The maximum ratio of weights $\gamma := \frac{w_{\max}}{w_{\min}}$ satisfies the bound
$$
\gamma = O \left( \min \left\{ \sqrt{d}, \frac{d^{1.5}}{k} \right\} \right).
$$

\item \label{cond:contraction factor} {\bf Contraction factor: } The contraction factor $q$ in Theorem~\ref{thm:local convergence} is defined as
\begin{align} \label{eqn:contracting factor}
q := \frac{2 w_{\max}}{w_{\min}} \left[ \frac{2 \alpha}{\sqrt{d}} \left( 1 + \alpha_0 \sqrt{\frac{k}{d}} \right)^2 + \beta'
 \right],
\end{align}
for some constants $\alpha_0, \beta'>0$, and $\alpha= \polylog(d)$. 
In particular, we need $\alpha\alpha_0\sqrt{k}/d +\beta' < w_{\max}/10w_{\min}$ which ensures $q < 1/2$. This is satisfied when $\sqrt{k}/d < w_{\max}/w_{\min}\poly\log d$ and $\beta' < w_{\max}/20w_{\min}$. The parameter $\beta'$ is determined by the following assumption (initialization).
%Suppose parameters $\beta'$, $d$, and $k$ are such that $q<1$. %Specially, when $\gamma$ is larger (more overcomplete setting), $\beta$ should be small enough to ensure $q<1$.

\item \label{cond:initialization} {\bf Initialization: } Let
$$
\epsilon_0 := \max \left\{ \dist \left(\ha^{(0)}, a_j\right), \dist \left(\hb^{(0)}, b_j \right) \right\},
$$
denote the initialization error w.r.t. to some $j \in [k]$. Suppose it is bounded as
$$
\epsilon_0 \leq \min \left\{ \frac{\beta'}{\alpha_0}, \sqrt{\frac{w_{\min}}{6 w_{\max}}}, \frac{w_{\min}q}{4 w_{\max}}, \frac{2 w_{\max}}{w_{\min} q} \left( \frac{w_{\min}}{6 w_{\max}} -  \alpha \frac{\sqrt{k}}{d} \right) \right\},
$$
for some constants $\alpha_0, \beta' >0$, $\alpha= \polylog(d)$, and $0<q<1/2$ which is defined in \eqref{eqn:contracting factor}.

\item \label{cond:2-to-pnorm} {\bf $2\to p$ norm:} For some fixed constant $p < 3$, $\max\{\|A^\top\|_{2\to p},\|B^\top\|_{2\to p},\|C^\top\|_{2\to p}\} \le 1+o(1)$.
%for some $\beta' < \frac{w_{\min}}{2\sqrt{2} w_{\max} (\alpha+1) \sqrt{\beta}}$, $C_1>0$.

\end{enumerate}

\begin{remark}Many of the assumptions are actually parameter choices. The only properties of random matrices required are \ref{cond:incoherence}, \ref{cond:spectral bound}, \ref{cond:tensor bounds} and \ref{cond:2-to-pnorm},. See Appendix~\ref{appendix:random arguments} for detailed discussion.
\end{remark}

Let us provide a brief discussion about the above assumptions. Condition \ref{cond:rank-k} requires the presence of a rank-$k$ decomposition for tensor $T$. We normalize the component vectors for convenience, and this removes the scaling indeterminacy issues which can lead to problems in convergence. % for the ALS method~\cite{uschmajew2012local}.
%Assumption (A2) supposes randomness. This provides useful bounds on spectral norm and incoherence properties of components which are exploited in local convergence analysis.
%The condition $(A2)$ on having full column rank for the components implies that we have a unique decomposition in \eqref{eqn:true-T}. In fact, uniqueness can occur under far weaker conditions. It has been shown that a  necessary condition for uniqueness is given by~\cite{jiang2004kruskal}
%\beq \rank(\Astar\odot \Bstar)= \rank(\Bstar\odot \Cstar) = \rank( \Astar\odot \Cstar) = k,\eeq which is weaker than $(A2)$. Since we assume non-degeneracy, we are in   the {\em undercomplete} or {\em overdetermined} setting where  $d \geq k$. In general, it is NP-hard to find the decomposition in the overcomplete regime~\cite{kolda_survey}, and our results are limited to the undercomplete setting.
Additionally, we impose incoherence constraint in \ref{cond:incoherence}, which allows us to provide convergence guarantee in the overcomplete setting. %Note that when the components form an orthonormal system (maximum incoherence), performing SVD in the undercomplete setting provides the decomposition.
Assumptions \ref{cond:spectral bound} and \ref{cond:tensor bounds} impose bounds on the spectral norm of tensor $T$ and its decomposition components.
%Note that if the vectors are generically drawn from $\Sc^{d-1}$, we have that $\rho = O(d^{-1/2})$ as assumed.
Note that assumptions \ref{cond:incoherence}-\ref{cond:tensor bounds} and \ref{cond:2-to-pnorm} are satisfied w.h.p. when the columns of $A$, $B$, and $C$ are generically drawn from unit sphere $\Sc^{d-1}$ (see Lemma \ref{lem:random incoherence & spectral bounds} and~\citet{guedon2007lp}), all others are parameter choices.
%Additionally, we assume in $(A4)$ that there is separation among the weights. This is needed to obtain good initialization through the matricized versions.
Assumption \ref{cond:size} limits the overcompleteness of problem which is required for providing convergence guarantees.
The first bound on perturbation in \ref{cond:perturbation bound} as $\psi \leq \frac{w_{\min}}{6}$ is required for local convergence guarantee and the second bound $\psi \leq \frac{w_{\min} \sqrt{\log k}}{\alpha_0 \sqrt{d}}$ is needed for arguing initialization provided by  Procedure~\ref{algo:SVD init}.  %but we later see that we need stricter bounds to ensure contraction depending on the other parameters such as $k$ and $d$.
Assumption \ref{cond:weight ratio} is required to ensure contraction happens in each iteration. Assumption \ref{cond:contraction factor} defines contraction ratio $q$ in each iteration, and Assumption \ref{cond:initialization} is the initialization condition required for local convergence guarantee.
%In the initialization Assumption \ref{cond:initialization}, when $d$ is large enough, the first bound dominates, and the other ones are only constants.  Therefore, we need $\epsilon_0 = O(1/\alpha) = O(1/\polylog(d))$.
%Assumption \ref{cond:contraction factor} is needed to ensure contraction in each iteration.

The tensor-spectral norm and $2\to p$ norm assumptions \ref{cond:tensor bounds} and \ref{cond:2-to-pnorm} may seem strong as we cannot even verify them given the matrix. However, when $k < d^{1.25-\epsilon}$ for arbitrary constant $\epsilon > 0$, both conditions are implied by incoherence. See Lemma~\ref{lem:incoherencetonorm}. We only need these assumptions to go to the very overcomplete setting.

\subsection{Random matrices satisfy the deterministic assumptions} \label{appendix:random arguments}
Here, we provide arguments that random matrices satisfy conditions \ref{cond:incoherence}, \ref{cond:spectral bound}, \ref{cond:tensor bounds}, and \ref{cond:2-to-pnorm}.
It is well known that random matrices are incoherent, and have small spectral norm (bound on spectral norm dates back to \citet{wigner1955characteristic}). See the following lemma.

\begin{lemma} \label{lem:random incoherence & spectral bounds}
Consider random matrix $X \in \Rbb^{d \times k}$ where its columns  are uniformly drawn at random from unit $d$-dimensional sphere $\Sc^{d-1}$.
%Consider random matrix $X \in \Rbb^{d \times k}$ where its entries are randomly drawn from a mean-zero subgaussian distribution, and then its columns are normalized such that $\| X_{(i)} \| = 1$.
Then, it satisfies the following incoherence and spectral bounds with high probability as
\begin{align*}
\max_{i,j \in [k], i \neq j} | \langle X_{i},X_{j} \rangle | & \leq \frac{\alpha}{\sqrt{d}}, \\
\| X \| & \leq 1 + \alpha_0 \sqrt{\frac{k}{d}},
\end{align*}
for some $\alpha = O(\sqrt{\log k})$ and $\alpha_0 = O(1)$.
\end{lemma}

 The spectral norm of the tensor is less well-understood. However, it can be bounded by the $2 \to 3$ norm of matrices. Using tools from ~\citet{guedon2007lp, adamczak2011chevet}, we have the following result.

\begin{lemma} \label{lem:2 to p norm bound}
Consider a random matrix $A \in \R^{d\times k}$ whose columns are drawn uniformly at random from unit sphere. If $k < d^{p/2}/\polylog(d)$, then
$$\bigl\| A^\top \bigr\|_{2\to p} \leq 1+o(1).$$
\end{lemma}

This directly implies Assumption \ref{cond:2-to-pnorm}. In particular, since we only apply Assumption \ref{cond:2-to-pnorm} to unsupervised setting ($k \le O(d)$) in Appendix~\ref{sec:clustering}, for randomly generated tensor, Assumption \ref{cond:2-to-pnorm} holds for all $p > 2$ (notice that we only need it to hold for some $p < 3$).

We also give an alternative proof of $2\to p$ norm which does not assume randomness and only relies on incoherence.

\begin{lemma} \label{lem:incoherencetonorm}
Suppose columns of matrix $A\in \R^{d\times k}$ have unit norm and satisfy the incoherence condition~\ref{cond:incoherence}. If $k \le d^{1.25-\epsilon}$ for arbitrary constant $\epsilon > 0$, then for any $p > 3-2\epsilon$, we have
$$\bigl\| A^\top \bigr\|_{2\to p} \leq 1+o(1).$$
\end{lemma}

\bprf
Let $L = \sqrt{d}/\poly\log d$. By incoherence assumption we know every subset of $L$ columns in $A$ has singular values within $1\pm o(1)$ (by Gershgorin Disk Theorem). 

For any unit vector $u$, let $S$ be the set of $L$ indices that are largest in $A^\top u$. By the argument above we know $\|(A_S)^\top u\| \le \|A_S\|\|u\| \le 1+o(1)$. In particular, the smallest entry in $A_S^\top u$ is at most $2/\sqrt{L}$. By construction of $S$ this implies for all $i$ not in $S$, $|A_i^\top u|$ is at most $2/\sqrt{L}$. Now we can write the $\ell_p$ ($p>2$) norm of $A^\top u$ as

\begin{align*}
\|A^\top u\|_p^p & = \sum_{i\in S} |A_i^\top u|^p+\sum_{i\not\in S}|A_i^\top u|^p\\
& \le \sum_{i\in S} |A_i^\top u|^2 + (2/\sqrt{L})^{p-2} \sum_{i\not\in S}|A_i^\top u|^2\\
& \le 1+o(1).
\end{align*}
Here the second inequality uses that every entry outside $S$ is small, and last inequality uses the fact that $p > 3-2\epsilon$.
\eprf

The $2\to 3$ norm implies a bound on the tensor spectral norm by H\"older's inequality.

\begin{fact}[H\"older's Inequality]
When $1/p+1/q = 1$, for two sequence of numbers $\{a_i\}, \{b_i\}$, we have
$$
\sum_i a_ib_i \le \left(\sum_i |a_i|^p\right)^{1/p}\left(\sum_i |b_i|^q\right)^{1/q}.
$$
\end{fact}

Consequently, we have the following corollary.

\begin{corollary}\label{cor:holder}
For vectors $f,g,h$, and weights $w_i \ge 0$, we have
\[\sum_i w_i f_ig_ih_i \le w_{\max}\|f\|_3\|g\|_3\|h\|_3.\]
\end{corollary}

\bprf
The proof applies H\"older's inequality twice as
$$\sum_i {w_i f_i g_i h_i} \le w_{\max} \sum_i|f_ig_ih_i| \le w_{\max}(\sum |f_i|^3)^{1/3} (\sum |g_ih_i|^{3/2})^{2/3} \le w_{\max}\|f\|_3\|g\|_3\|h\|_3,$$
where in the first application, $p = 3$ and $q = 3/2$, and in the second application, $p = q = 2$ (which is the special case known as Cauchy-Schwartz).
\eprf

In the following lemma, it is shown that the first bound in Assumption \ref{cond:tensor bounds} holds for random matrices w.h.p.

\begin{lemma} \label{lem:tensorspectral}
Let $A$, $B$, and $C$ be random matrices in $\R^{d\times k}$ whose columns are drawn uniformly at random from unit sphere. If $k < d^{3/2}/\polylog(d)$, and
\[T = \sum_{i\in [k]} w_i a_i\otimes b_i\otimes c_i,\]
then $$\|T\| \le O(w_{\max}).$$
\end{lemma}

\bprf
For any unit vectors $\ha, \hb, \hc$, we have
\begin{align*}
T(\ha, \hb,\hc) & = \sum_{i\in [k]} w_i (A^\top \ha)_i(B^\top \hb)_i(C^\top \hc)_i \\
& \leq w_{\max}\|A^\top \ha\|_3\|B^\top \hb\|_3\|C^\top \hc\|_3 \\
& \leq w_{\max} \|A^\top\|_{2\to 3}\|\ha\| \cdot \|B^\top\|_{2\to 3}\|\hb\| \cdot \|C^\top\|_{2\to 3}\|\hc\| \\
& = O(w_{\max}),
\end{align*}
where Corollary~\ref{cor:holder} is exploited in the first inequality, and Lemma~\ref{lem:2 to p norm bound} is used in the last inequality.
\eprf

For the case with two undercomplete and one overcomplete dimensions (see Corollary~\ref{corollary:TwoUnder OneOver}), we can prove the tensor spectral norm using basic properties of the matrices $A,B,C$.

\begin{lemma} \label{lem:tensorspectral2}
Let $A,B\in \R^{d_u\times k}$ be matrices with spectral norm bounded by $O(1)$, and $C \in \R^{d_o\times k}$ be a matrix whose columns have unit norm. Let $$T = \sum_{i=1}^k w_i a_i\otimes b_i\otimes c_i,$$ then we have $$\|T\| \le O(w_{\max}).$$
\end{lemma}

\bprf
For any unit vectors $u,v \in \R^{d_u}$ and $w\in \R^{d_o}$, by assumptions we know $\|A^\top  u\| \le O(1), \|B^\top v\|\le O(1)$ and $\|C^\top w\|_\infty \le 1$. Now we have
\begin{align*}
T(u,v,w) & = \sum_{i=1}^k w_i \inner{a_i,u}\inner{b_i,v}\inner{c_i,w} \\
& \le w_{\max} \sum_{i=1}^k |\inner{a_i,u}\inner{b_i,v}| \\
& \le w_{\max} \|A^\top u\|\|B^\top v\| \\
& = O(w_{\max}).
\end{align*}
The first inequality uses triangle inequality and the fact that $|\inner{c_i,w}|\le 1$. The Cauchy-Schwartz inequality is exploited in the second inequality. Therefore, the spectral norm of the tensor is bounded by $O(w_{\max})$.
\eprf

Finally, we show in the following lemma that the second bound in Assumption \ref{cond:tensor bounds} is satisfied for random matrices.

\begin{lemma} \label{lem:aux local convergence}
Let $A, B, C \in \R^{d\times k}$ be independent, normalized (column) Gaussian matrices. Then for all $i \in [k]$, we have with high probability
\[ \left\|C_{\backslash i}\diag(w^{\setminus i})(J_A*J_B)^{\backslash i}_{i} \right\| = \tilde{O} \left( w_{\max} \frac{\sqrt{k}}{d} \right).\]
%If $A$, $B$, and $C \in \R^{d\times k}$ are independent, normalized Gaussian matrices, then for all $i$, we have with high probability, $\|C_{\backslash i}\diag(w^{\setminus i})(J_A*J_B)^{\backslash i}_{i}\| = \tilde{O}(w_{\max}\sqrt{k}/d)$.
\end{lemma}

\bprf
We have
\[C_{\backslash i}\diag(w^{\setminus i})(J_A*J_B)^{\backslash i}_{i} = \sum_{j\ne i} C_j w_j\langle A_i, A_j\rangle\langle B_i, B_j\rangle = \sum_{j\ne i} C_j \delta_j,\]
where $\delta_j := w_j\langle A_i, A_j\rangle\langle B_i, B_j\rangle$ is independent of $C_j$. From Lemma~\ref{lem:random incoherence & spectral bounds}, columns of $A$ and $B$ are incoherent, and therefore, for $j\ne i$, we have
\[|\delta_j| = \tilde{O}(w_{\max}/d).\]
Now since $C_j$'s are independent, zero mean vectors, the sum $\sum_{j \neq i} \delta_j C_j$ is zero mean and its variance is bounded by $\tilde{O}(w_{\max}^2k/d^2)$. Then, from vector Bernstein's bound we have with high probability
\[\left\|C_{\backslash i}\diag(w^{\setminus i})(J_A*J_B)^{\backslash i}_{i}\right\| = \tilde{O} \left( w_{\max} \frac{\sqrt{k}}{d} \right).\]
The proof is completed by applying union bound.
\eprf

\subsubsection*{Spectral norm of Khatri-Rao product}

For the convergence guarantees of the second step of algorithm on removing residual error, we need the following additional bound on the spectral norm of Khatri-Rao product of random matrices.

\begin{enumerate}[label={(A\arabic*)}]
\setcounter{enumi}{10}

\item \label{cond:spectral bound KR} {\bf Spectral Norm Condition on Khatri-Rao Products: }
The components satisfy the following spectral norm bound on the Khatri-Rao products as
\begin{align*}
\max \left\{ \|A \odot B\|, \|B \odot C\|, \|A \odot C\| \right\} \leq 1 + \alpha_0 \frac{\sqrt{k}}{d},
%& \max \left\{\left\| A \odot J_B \right\|, \left\| B \odot J_A \right\| \right\} \leq \alpha_J \frac{\sqrt{k}}{d},
\end{align*}
for $\alpha_0 \le \poly\log d$.

\end{enumerate}

We now prove that Assumption~\ref{cond:spectral bound KR} is satisfied with high probability, if the columns of $A$, $B$ and $C$ are uniformly i.i.d. drawn from unit $d$-dimensional sphere.

The key idea is to view $(A\odot B)^\top (A\odot B)$ as the sum of random matrices, and use the following Matrix Bernstein's inequality %\cite{} 
to prove concentration results.

\begin{lemma}
Let $M = \sum_{i=1}^n M_i$ be sum of independent symmetric $d\times d$ matrices with $\E[M_i] = 0$, assume all matrices $M_i$'s have spectral norm at most $R$ almost surely, let $\sigma^2 = \|\E[M_i^2]\|$, then for any $\tau$
\[
\Pr[\|M\| \ge \tau] \le 2d \exp\left(\frac{-\tau^2/2}{\sigma^2 + R\tau/3}\right).
\]
\end{lemma}

\paragraph{Remark:} Although the lemma requires all $M_i$'s to have spectral norm at most $R$ almost surely, it suffices to have spectral norm bounded by $R$ with high probability and bounded by $R^\infty = \mbox{poly}(d,k)$ almost surely. This is because we can always condition on the fact that $\|M_i\| \le R$ for all $i$. Such conditioning can only change the expectations by a negligible amount, and does not affect independence between $M_i$'s.

Random unit vectors are not easy to work with, as entries in the same column are not independent. Thus, we first prove the result for matrices $A$ and $B$ whose entries are independent Gaussian variables.

\begin{lemma}
Suppose $A$, $B \in \R^{d\times k} (k > \polylog d)$ are independent random matrices with independent Gaussian entries, let $M = (A \odot B)^\top (A\odot B) = (A^\top A) * (B^\top B)$, then with high probability
\[
\| M- \diag(M)\| \le O(d\sqrt{k\log d})
\]
\end{lemma}

\bprf
Let $a_1, a_2, ..., a_d \in \R^k$ be the columns of $A^\top$ (the rows of $A$, but treated as column vectors). We can rewrite $M - \diag M$ as
\[
M - \diag M = (\sum_{i\in[d]} a_ia_i^\top)*(B^\top B - \diag(B^\top B)) = \sum_{i\in [d]} (a_ia_i^\top)*(B^\top B - \diag(B^\top B)).
\]

Now let $Q = B^\top B - \diag(B^\top B)$, and $M_i = (a_ia_i^\top)*Q$, we would like to bound the spectral norm of the sum $M = \sum_{i\in [d]} M_i$. Clearly these entries are independent, $\E[M_i] = \E[a_ia_i^\top]*Q = I * Q = 0$, so we can apply Matrix Bernstein bound.

Note that when $d < k$, by standard random matrix theory we know $\|Q\| \le O(k)$. Also,  every row of $Q$ has norm smaller than the corresponding row of $B^\top B$, which is bounded by $\|B\|\|b_{(i)}\| \le O(\sqrt{kd})$. When $d \ge k$, again by matrix concentration we know $\|Q\| \le O(\sqrt{dk \log d})$. Every row of $Q$ has norm bounded by $O(\sqrt{kd})$ (because entries in a row are independently random, with variance equal to $d$).

First let us bound the spectral norm for each of the $M_i$'s. Notice that for any vector $v$, $v^\top[(a_ia_i^\top) *Q]v = (v*a_i)^\top Q (v*a_i)$ by definition of Hadamard product. On the other hand, $\|v*a_i\| \le\|v\| \|a_i\|_\infty$. With high probability $\|a_i\|_\infty \le O(\sqrt{\log k})$, hence $\|M_i\| \le \|a_i\|_\infty^2 \|Q\|$. This is bounded by $O(k\log d)$ when $d < k$ and $O(\sqrt{kd}\log^2 d)$ when $k \le d$.

Next we bound the variance $\|\E[\sum_{i\in[d]}M_i^2]\|$. Since all the $M_i$'s are i.i.d., it suffices to analyze $\E[M_1^2]$. Let $T = \E[M_1^2] = \E[((a_1a_1^\top)*Q)^2]$, by definition of Hadamard product, 
we know
\[
T_{p,q} = \E[\sum_{r\in [k]} Q_{p,r}Q_{r,q} a_1(p)a_1(q)a_1(r)^2].
\]

This number is $0$ when $p \ne q$ by independence of entries of $a_1$. When $p = q$, this is bounded by $3\sum_{r\in[k]}Q_{p,r}^2$ because $\E[a_1(p)^2a_1(r)^2]$ is $1$ when $p\ne r$ and $3$ when $p = r$. Therefore $T_{p,p} \le 3\sum_{r\in[k]}Q_{p,r}^2 = 3\|Q^{(p)}\|^2 \le O(dk)$. Since $T$ is a diagonal matrix, we know $\|T\| \le O(dk)$, and $\sigma^2 = \|dT\| = O(d^2k)$.

By Matrix Bernstein we know with high probability $\|M\| \le O(d\sqrt{k \log d})$. 
\eprf

Using this lemma, it is easy to get a bound when columns of $A$, $B$ are unit vectors. In this case, we just need to normalize the columns, the normalization factor is bounded between $d^2/2$ and $2d^2$ with high probability, and therefore, $\|(A^\top A)(B^\top B)-I\| \le O(\sqrt{k\log d}/d)$.

\section{Proof of Convergence Results in Theorems \ref{thm:local convergence} and \ref{thm:global convergence}} \label{sec:convergence proof}

The main part of the proof is to show that error contraction happens in each iteration of Algorithms~\ref{algo:Power method form}~and~\ref{algo:coordinate-descent} as the two main parts of the algorithm. Then, the contraction result after $t$ iterations is directly argued.

In the following, we first provide a local contraction result for the tensor power iteration~\eqref{eqn:asymmetric power update} in Algorithm~\ref{algo:Power method form} given noisy tensor $\hT$. This leads to Lemma~\ref{thm:local convergence-poweriteration} which is the local convergence guarantee of the tensor power updates.
Then, we provide a local contraction argument for the coordinate descent step~\eqref{eqn:BiasRemoval} in Algorithm~\ref{algo:coordinate-descent}.

Combining the above convergence arguments for both updates conclude the overall local convergence guarantee in Theorem.~\ref{thm:local convergence}. Then, combing this local convergence guarantee and the initialization result in Theorem \ref{thm:initialization} leads to the global convergence guarantee in Theorem~\ref{thm:global convergence}. In addition, the result in Corollary~\ref{corollary:TwoUnder OneOver} is similarly argued where the bound on the spectral norm of the tensor is argued in Lemma~\ref{lem:tensorspectral2}.

\subsection{Convergence of tensor power iteration: Algorithm~\ref{algo:Power method form}}

In this section, we prove Lemma~\ref{thm:local convergence-poweriteration} which is the local convergence guarantee of the tensor power updates in  Algorithm~\ref{algo:Power method form}.

Define function $f(\epsilon;k,d)$ as
\begin{align} \label{eqn:function f}
f(\epsilon;k,d) := \alpha  \frac{\sqrt{k}}{d} + \frac{2 \alpha}{\sqrt{d}} \left( 1 + \alpha_0 \sqrt{\frac{k}{d}} \right)^2 \epsilon + \alpha_0 \epsilon^2,
\end{align}
where $\alpha = \polylog(d)$ and $\alpha_0 = O(1)$. Notice that this function is a small constant when $k < d^{1.5}/\poly\log d$. 

\begin{lemma}[Contraction result  of Algorithm \ref{algo:Power method form} in one update] \label{lem:local convergence}
Consider $\hT = T + \Psi$ as the input to Algorithm \ref{algo:Power method form}, where $T$ is a rank-$k$ tensor, and $\Psi$ is a perturbation tensor. Suppose Assumptions \ref{cond:rank-k}-\ref{cond:size} hold, and estimates $\ha$ and $\hb$ satisfy distance bounds
\begin{align*}
\dist (\ha, a_j) & \leq \epsilon_a, \\
\dist (\hb, b_j) & \leq \epsilon_b,
\end{align*}
for some $j \in [k]$, and $\epsilon_a, \epsilon_b>0$. Let $\epsilon := \max \{ \epsilon_a,\epsilon_b \}$, and suppose $\psi$ defined in \eqref{eqn:psi definition} be small enough such that\,\footnote{This is the denominator of bound provided in \eqref{eqn:general contraction}.}
$$
w_j - w_j \epsilon^2 - w_{\max} f(\epsilon;k,d) - \psi > 0,
$$
where $f(\epsilon;k,d)$ is defined in \eqref{eqn:function f}. Then, update $\hc$ in \eqref{eqn:asymmetric power update} satisfies the following distance bound with high probability (w.h.p.)
\begin{align} \label{eqn:general contraction}
\dist(\hc,c_j)
\leq \frac{ w_{\max} f(\epsilon;k,d) + \psi}{w_j - w_j \epsilon^2 - w_{\max} f(\epsilon;k,d) - \psi}.
\end{align}
Furthermore, if the bound in \eqref{eqn:general contraction} is such that $\dist(\hc,c_j) \leq \epsilon$, then the update $\hw := \hT(\ha, \hb, \hc)$ in \eqref{eqn:weight update} also satisfies w.h.p.
$$
| \hw - w_j |
\leq  2 w_j \epsilon^2 + w_{\max} f(\epsilon;k,d) + \psi.
$$
\end{lemma}

%Note that without using $\dist$ notion in the proof of above proposition, we can argue the following error bound
%\begin{align*}
%\left\| \tl{C}_{(j)} - w_j C_{(j)} \right\|
%\leq w_j (2 \epsilon+\epsilon^2) + w_{\max} f(\epsilon;k,d) + \psi,
%\end{align*}
%which may be easier to handle.

\begin{remark}
In the asymptotic regime, $f(\epsilon;k,d)$ is
$$
f(\epsilon;k,d)
= \tl{O} \left( \frac{\sqrt{k}}{d} \right)
+ \tl{O} \left( \max \left\{ \frac{1}{\sqrt{d}}, \frac{k}{d^{3/2}} \right\} \right) \epsilon
+ O (1) \epsilon^2.
$$
%\begin{align*}
%f(\epsilon;k,d)
%= \left\{ \begin{array}{ll}
%\tl{O} \left( \frac{\sqrt{k}}{d} \right) (1+\epsilon^2) + \tl{O} \left( \frac{1}{\sqrt{d}} \right) \epsilon + c \epsilon^2 & ,k<d, \\
%\tl{O} \left( \frac{k}{d^{3/2}} \right) (1+\epsilon+\epsilon^2) + \tl{O} \left( \sqrt{\frac{k}{d}} \right) \epsilon^2 & ,k>d,
%\end{array}\right.
%\end{align*}
%for some constant $c>0$.
Note that the last term is the only effective contracting term. The other terms include a constant term, and the term involving $\epsilon$ disappears in only one iteration as long as $k,d \rightarrow \infty$, and $\tl{O} \left( \frac{k}{d^{3/2}} \right) \rightarrow 0$.
\end{remark}

\begin{remark}[Rate of convergence]
The local convergence result provided in Theorem \ref{thm:local convergence} has a linear convergence rate. But, Algorithm \ref{algo:Power method form} actually provides an almost-quadratic convergence rate in the beginning, and linear convergence rate later on. It can be seen by referring to one-step contraction argument provided in Lemma \ref{lem:local convergence} where the quadratic term $\alpha_0 \epsilon^2$ exists. In the beginning, this term is dominant over linear term involving $\epsilon$, and we have almost-quadratic convergence. Writing $\alpha_0 \epsilon^2 = \alpha_0 \epsilon^\zeta \epsilon^{2-\zeta}$, we observe that we get rate of convergence equal to $2-\zeta$ as long as we have initialization error bounded as $\epsilon_0^\zeta = O(1)$.
Therefore, we can get arbitrarily close to quadratic convergence with appropriate initialization error. Note that when the model is more overcomplete, the algorithm more rapidly reaches to the linear convergence phase. For the sake of clarity, in proposing Theorem \ref{thm:local convergence}, we approximated the almost-quadratic convergence rate in the beginning with linear convergence.
\end{remark}

Lemma \ref{lem:local convergence} is proposed in the general form. In Lemma \ref{lem:local convergence 2}, we provide explicit contraction result by imposing additional perturbation, contraction and initialization Assumptions \ref{cond:perturbation bound}, \ref{cond:contraction factor} and \ref{cond:initialization}. We observe that under reasonable rank, perturbation and initialization conditions, the denominator in \eqref{eqn:general contraction} can be lower bounded by a constant, and the numerator is explicitly bounded by a term involving $\epsilon$, and a constant non-contracting term.

\begin{lemma}[Contraction result of Algorithm \ref{algo:Power method form} in one update] \label{lem:local convergence 2}
Consider $\hT = T + \Psi$ as the input to Algorithm \ref{algo:Power method form}, where $T$ is a rank-$k$ tensor, and $\Psi$ is a perturbation tensor. Let Assumptions\,\footnote{As mentioned in the assumptions, from perturbation bound in \ref{cond:perturbation bound}, only the bound $\psi \leq \frac{w_{\min}}{6}$ is required here.} \ref{cond:rank-k}-\ref{cond:initialization} hold. Note that initialization bound in \ref{cond:initialization} is satisfied for some $j \in [k]$. Then, update $\hc$ in \eqref{eqn:asymmetric power update} satisfies the following distance bound with high probability (w.h.p.)
$$
\dist (\hc, c_j)
\leq \underbrace{\Const}_{\textnormal{non-contracting term}} + \underbrace{q \epsilon_0}_{\textnormal{contracting  term}},
$$
where
\begin{align} \label{eqn:Constant bound}
\Const := \frac{2}{w_{\min}} \left( \psi + w_{\max} \alpha \frac{\sqrt{k}}{d} \right),
\end{align}
and contraction ratio $q<1/2$ is defined in \eqref{eqn:contracting factor}. Note that $\alpha = \polylog(d)$. 
In addition, if the above bound be such that $\dist(\hc,c_j) \leq \epsilon_0$, then the update $\hw := \hT(\ha, \hb, \hc)$ in \eqref{eqn:weight update} also satisfies w.h.p.
$$
| \hw - w_j |
\leq \frac{w_{\min}}{2} \Const + w_{\min} q \epsilon_0.
$$
%The final estimation of weights satisfy
%\begin{align*}
%| \hw - w_j |
%\leq O(w_{max}\alpha\sqrt{k}/d + \psi).
%\end{align*}
\end{lemma}

\bprfof{Lemma~\ref{thm:local convergence-poweriteration}}
We incorporate condition \ref{cond:weight ratio} to show that $q<1/2$ in assumption \ref{cond:contraction factor} is satisfied. In addition,  \ref{cond:weight ratio} implies that the bound on $\epsilon_0$ in assumption \ref{cond:initialization} holds where it can be shown that the bound in \ref{cond:initialization} is bounded as $O(1/\gamma)$.
Then, the result is directly proved by iteratively applying the result of Lemma \ref{lem:local convergence 2}.
\eprfof

\subsection*{Proof of auxiliary lemmata: tensor power iteration in Algorithm~\ref{algo:Power method form}}

Before providing the proofs, we remind a few definitions and notations.

In Assumption \ref{cond:incoherence}, matrices $J_A$, $J_B$, and $J_C$, are defined as incoherence matrices with zero diagonal entries such that $A^\top A = I + J_A$, $B^\top B = I + J_B$, and $C^\top C = I + J_C$. We have $\max \left\{ \|J_A\|_\infty,\|J_B\|_\infty, \|J_C\|_\infty \right\} \leq \rho$ as in \eqref{eqn:incoherence}.

Given matrix $A \in \Rbb^{d \times k}$, the following notations are defined to refer to its sub-matrices. $A_j$ denotes the $j$-th column and $A^j$ denotes the $j$-th row of $A$. Hence, we have $A_j = a_j, j \in [k]$. In addition, $A_{\setminus j} \in \Rbb^{d \times (k-1)}$ is $A$ with its $j$-th column removed, and $A^{\setminus j} \in \Rbb^{(d-1) \times k}$ is $A$ with its $j$-th row removed.

\bprfof{Lemma \ref{lem:local convergence}}
Let $z_a^* \perp a_j$ and $z_b^* \perp b_j$ denote the vectors that achieve supremum value in \eqref{eqn:dist function definition} corresponding to $\dist (\ha, a_j)$ and $\dist (\hb, b_j)$, respectively. Furthermore, without loss of generality, assume $\| z_a^* \| = \| z_b^* \| = 1$. Then, $\ha$ and $\hb$ are decomposed as
\begin{align}
\sublabon{equation}
\ha & = \langle a_j, \ha \rangle a_j + \dist(\ha,a_j) z_a^*, \label{eqn:ha expansion on aj} \\
\hb & = \langle b_j, \hb \rangle b_j + \dist(\hb,b_j) z_b^*.\label{eqn:hb expansion on bj}
\end{align}
\sublaboff{equation}Let $\overline{C} := C \Diag(w)$ denote the unnormalized matrix $C$, and $\tl{c} := \hT(\ha,\hb,I)$ denote the unnormalized update in \eqref{eqn:asymmetric power update}. The goal is to bound $\dist \left( \tl{c}, \overline{C}_{j} \right)$. Consider any $z_c \perp \overline{C}_{j}$ such that $\|z_c\|=1$. Then, we have
$$
\langle z_c, \tl{c} \rangle = \hT(\ha,\hb,z_c) = T(\ha, \hb, z_c) + \Psi(\ha, \hb, z_c).
$$
Substituting $\ha$ and $\hb$ from \eqref{eqn:ha expansion on aj} and \eqref{eqn:hb expansion on bj}, we have
\begin{align*}
T(\ha, \hb, z_c)
= & \ \underbrace{\langle a_j, \ha \rangle \langle b_j, \hb \rangle T(a_j, b_j, z_c)}_{S_1}
 + \underbrace{\langle a_j, \ha \rangle \dist(\hb,b_j) T(a_j, z_b^*, z_c)}_{S_2} \\
& + \underbrace{\dist(\ha,a_j) \langle b_j, \hb \rangle T(z_a^*, b_j, z_c)}_{S_3}
 + \underbrace{\dist(\ha,a_j) \dist(\hb,b_j) T(z_a^*, z_b^*, z_c)}_{S_4}.
\end{align*}
In the following derivations, we repeatedly use the equality that for any $u,v \in \Rbb^d$, we have $T(u, v, I) = \overline{C} ( A^\top u * B^\top v)$. For $S_1$, we have
\begin{align*}
S_1
& \leq | T(a_j, b_j, z_c) |
= |z_c^\top \overline{C} ( A^\top a_j * B^\top b_j)| \\
& = \left| z_c^\top \overline{C} \left[ e_j + \left(J_A * J_B\right)_{j} \right] \right| \\
& = \left| z_c^\top \overline{C}_{\setminus j} \left( J_A * J_B \right)_{j}^{\setminus j} \right| \\
& \leq w_{\max} \alpha \frac{\sqrt{k}}{d},
\end{align*}
where equalities $A^\top A = I + J_A$ and $B^\top B = I + J_B$ are exploited in the second equality, and the assumption that $z_c \perp \overline{C}_{j}$ is used in the last equality. The last inequality is from Assumption \ref{cond:tensor bounds}. %, where it is shown in Lemma \ref{lem:aux local convergence} that this condition holds for random matrices. 
For $S_2$, we have
\begin{align*}
S_2
& \leq \epsilon_b | T(a_j, z_b^*, z_c) |
= \epsilon_b |z_c^\top \overline{C} ( A^\top a_j * B^\top z_b^*)| \\
& = \epsilon_b \left| z_c^\top \overline{C}_{\setminus j} \left[ (J_A)_{j}^{\setminus j} * \left( B_{\setminus j} \right)^\top z_b^* \right] \right| \\
& \leq \epsilon_b \left\| \overline{C}_{\setminus j} \right\| \cdot \left\| (J_A)_{j}^{\setminus j} \right\|_{\infty} \cdot \left\| \left( B_{\setminus j} \right)^\top z_b^* \right\| \\
& \leq w_{\max} \frac{\alpha}{\sqrt{d}} \left( 1 + \alpha_0 \sqrt{\frac{k}{d}} \right)^2 \epsilon_b,
\end{align*}
for some $\alpha = \polylog(d)$ and $\alpha_0 = O(1)$. Second inequality is concluded from $\|u * v\| \leq \|u\|_\infty \cdot \|v\|,$ and Assumptions \ref{cond:incoherence} and \ref{cond:spectral bound} are exploited in the last inequality. Similarly, for $S_3$, we have
\begin{align*}
S_3
& \leq \epsilon_a \left| z_c^\top \overline{C}_{\setminus j} \left[ (J_B)_{j}^{\setminus j} * \left( A_{\setminus j} \right)^\top z_a^* \right] \right| \\
& \leq w_{\max} \frac{\alpha}{\sqrt{d}} \left( 1 + \alpha_0 \sqrt{\frac{k}{d}} \right)^2 \epsilon_a.
\end{align*}
Finally, for $S_4$, we have
$$
S_4
\leq \epsilon_a \epsilon_b |T(z_a^*, z_b^*, z_c) |
\leq \epsilon_a \epsilon_b \| T \|
\leq  w_{\max} \alpha_0 \epsilon_a \epsilon_b,
$$
for some $\alpha_0 = O(1)$. The bound on $\|T\|$ is from Assumption \ref{cond:tensor bounds}. Note that for random components, we showed in Lemma~\ref{lem:tensorspectral} that this bound holds w.h.p. exploiting Assumption \ref{cond:size} and results of \citet{guedon2007lp}. For the error term $\Psi(\ha, \hb, z_c)$, we have
$$
\Psi(\ha, \hb, z_c) \leq \psi,
$$
which is concluded from the definition of spectral norm of a tensor. Note that all vectors $\ha$, $\hb$, $z_c$ have unit norm.
%\begin{align*}
%\Psi(\ha, \hb, z_c)
%= z_c^\top \mat(\Psi,3)(\hb \odot \ha)
%\leq \| z_c^\top\| \cdot \| \mat(\Psi,3) \| \cdot \left\| \hb \odot \ha \right\|
%= \left\| \mat(\Psi,3) \right\|
%\leq \psi,
%\end{align*}
%where equality $\|\hb \odot \ha\| = \|\hb\| \cdot \|\ha\| = 1$ is exploited in the last equality, and definition of $\psi$ in \eqref{eqn:psi definition} is used in the last inequality.

Let $\epsilon := \max\{\epsilon_a, \epsilon_b\}$. Then, combining all the above bounds, we have w.h.p.
$$
\langle z_c, \tl{c} \rangle \leq w_{\max} f(\epsilon;k,d) + \psi,
$$
where $f(\epsilon;k,d)$ is
$$
f(\epsilon;k,d) := \alpha  \frac{\sqrt{k}}{d} + \frac{2 \alpha}{\sqrt{d}} \left( 1 + \alpha_0 \sqrt{\frac{k}{d}} \right)^2 \epsilon + \alpha_0 \epsilon^2.
$$
For $\tl{c}$, we have
\begin{align*}
\tl{c} & = T(\ha,\hb,I) + \Psi(\ha,\hb,I) \\
& = \sum_i w_i \langle a_i,\ha \rangle \langle b_i,\hb \rangle c_i + \Psi(\ha,\hb,I) \\ %\mat(\Psi,3)(\hb \odot \ha) \\
& = w_j \langle a_j,\ha \rangle \langle b_j,\hb \rangle c_j + \sum_{i \neq j} w_i \langle a_i,\ha \rangle \langle b_i,\hb \rangle c_i + \Psi(\ha,\hb,I), %\mat(\Psi,3)(\hb \odot \ha),
\end{align*}
and therefore,
\begin{align*}
\| \tl{c} \|
& \geq \left\| w_j \langle a_j,\ha \rangle \langle b_j,\hb \rangle c_j \right\| - \biggl\| \sum_{i \neq j} w_i \langle a_i,\ha \rangle \langle b_i,\hb \rangle c_i \biggr\| - \| \Psi(\ha,\hb,I) \| \\ %\| \mat(\Psi,3)(\hb \odot \ha) \| \\
& \geq w_j - w_j \epsilon^2 - w_{\max} f(\epsilon;k,d) - \psi,
\end{align*}
where inequality $\langle a_j,\ha \rangle \langle b_j,\hb \rangle \geq 1-\epsilon^2$ is exploited in the last inequality.
Hence, as long as this lower bound on $\| \tl{c} \|$ is positive (small enough $\epsilon$ and $\psi$), we have
\begin{align}
\dist(\tl{c},\overline{C}_{j})
\leq \frac{ w_{\max} f(\epsilon;k,d) + \psi}{w_j - w_j \epsilon^2 - w_{\max} f(\epsilon;k,d) - \psi}.
\end{align}
Since $\dist(\cdot,\cdot)$ function is invariant with respect to norm, we have $\dist \left( \hc, c_j \right) = \dist \left( \tl{c}, \overline{C}_{j} \right)$ which finishes the proof for bounding $\dist \left( \hc, c_j \right)$. Note that $\tl{c} = \| \tl{c} \| \hc$, and $\overline{C}_{j} = w_j c_j$ where $w_j>0$.

Now, we provide the bound on $|w_j - \hw|$. As assumed in the lemma, we have distance bounds
$$
\max \left\{ \dist \left( \ha, a_j \right), \dist \left( \hb, b_j \right), \dist \left( \hc, c_j \right) \right\} \leq \epsilon.
$$
The estimate $\hw = \hT (\ha,\hb,\hc)$ proposed in \eqref{eqn:weight update} can be expanded as
\begin{align*}
\hw & = T(\ha,\hb, \hc) + \Psi(\ha,\hb, \hc) \\
& = \sum_i w_i \langle a_i,\ha \rangle \langle b_i,\hb \rangle \langle c_i,\hc \rangle + \Psi(\ha,\hb, \hc) \\
& = w_j \langle a_j,\ha \rangle \langle b_j,\hb \rangle \langle c_j,\hc \rangle + \sum_{i \neq j} w_i \langle a_i,\ha \rangle \langle b_i,\hb \rangle \langle c_i,\hc \rangle + \Psi(\ha,\hb, \hc),
\end{align*}
and therefore,
\begin{align*}
| w_j - \hw |
& \leq \left| w_j \left( 1 - \langle a_j,\ha \rangle \langle b_j,\hb \rangle \langle c_j, \hc \rangle \right) \right| + \biggl| \sum_{i \neq j} w_i \langle a_i,\ha \rangle \langle b_i,\hb \rangle \langle c_i, \hc \rangle \biggr| + \left| \Psi(\ha,\hb, \hc) \right| \\
& \leq  w_j \left(1 - \left( 1-\epsilon^2 \right)^{1.5} \right) + w_{\max} f(\epsilon;k,d) + \psi \\
& \leq  2 w_j \epsilon^2 + w_{\max} f(\epsilon;k,d) + \psi,
\end{align*}
where $\langle a_j,\ha \rangle \langle b_j,\hb \rangle \langle c_j, \hc \rangle \geq \left( 1-\epsilon^2 \right)^{1.5}$ is exploited in the second inequality. Notice that this argument is similar to the argument provided earlier for lower bounding $\| \tilde{c} \|$.

\eprfof

\bprfof{Lemma \ref{lem:local convergence 2}}
The result is proved by applying Lemma \ref{lem:local convergence}, and incorporating additional conditions \ref{cond:perturbation bound}, \ref{cond:contraction factor}, and \ref{cond:initialization}. $f(\epsilon_0;k,d)$ in \eqref{eqn:function f} can be bounded as
\begin{align*}
f(\epsilon_0;k,d)
& = \alpha  \frac{\sqrt{k}}{d} + \frac{2 \alpha}{\sqrt{d}} \left( 1 + \alpha_0 \sqrt{\frac{k}{d}} \right)^2 \epsilon_0 + \alpha_0 \epsilon_0^2 \\
& \leq \alpha  \frac{\sqrt{k}}{d} + \left[ \frac{2 \alpha}{\sqrt{d}} \left( 1 + \alpha_0 \sqrt{\frac{k}{d}} \right)^2 + \beta' \right] \epsilon_0 \\
& = \alpha  \frac{\sqrt{k}}{d} + \frac{w_{\min}}{2 w_{\max}} q \epsilon_0,
\end{align*}
where %first inequality is concluded from size constraint \ref{cond:size}, and
 $\epsilon_0 \leq \frac{\beta'}{\alpha_0}$ from Assumption \ref{cond:initialization} is exploited in the inequality. The last equality is concluded from definition of contracting factor $q$ in \eqref{eqn:contracting factor}.
On the other hand, the denominator in \eqref{eqn:general contraction} can be lower bounded as
$$
%w_{\min} - w_{\max} (2 \epsilon_0+\epsilon_0^2) - w_{\max} f(\epsilon_0;k,d) -\psi =
w_{\min} \left[ 1 - \frac{w_{\max}}{w_{\min}} \epsilon_0^2 - \frac{w_{\max}}{w_{\min}} f(\epsilon_0;k,d) - \frac{\psi}{w_{\min}} \right]
\geq w_{\min} \left[ 1 - \frac{1}{6} - \frac{1}{6} - \frac{1}{6} \right]
= \frac{w_{\min}}{2},
$$
where Assumptions \ref{cond:initialization} and \ref{cond:perturbation bound} are used in the inequality. Applying Lemma \ref{lem:local convergence}, the result on $\dist(\hc, c_j)$ is proved.

From Lemma \ref{lem:local convergence}, we also have
\begin{align*}
| \hw - w_j |
& \leq  2 w_j \epsilon_0^2 + w_{\max} f(\epsilon_0;k,d) + \psi \\
& \leq \frac{w_{\min}}{2} \Const + 2 w_j \epsilon_0^2 + \frac{w_{\min}}{2} q \epsilon_0 \\
& \leq \frac{w_{\min}}{2} \Const + w_{\min} q \epsilon_0.
\end{align*}
where $\epsilon_0 \leq \frac{w_{\min}q}{4 w_{\max}}$ from Assumption \ref{cond:initialization} is used in the last inequality.
%From Lemma \ref{lem:local convergence}, we also have
%$$
%| \hw - w_j |
% \leq  2 w_j \epsilon^2 + w_{\max} f(\epsilon;k,d) + \psi ,
%$$
%where $\epsilon$ can be replaced by the final error $Const = 2(\psi+w_{max}\alpha\sqrt{k}/d)/w_{min}$ (see Equation~\ref{eqn:Constant bound}). Substituting this, and notice the bounds on $w_{max}/w_{min}$ and $\psi$, we know $| \hw - w_j | \le O(w_{max}\alpha\sqrt{k}/d + \psi)$.
\eprfof

\subsection{Convergence of removing residual error: Algorithm~\ref{algo:coordinate-descent}} \label{sec:convergence proof-coordinate descent}

In this section, we provide convergence of the coordinate descent of Algorithm~\ref{algo:coordinate-descent} for removing the residual error.
We first provide the following definition.

%\mjcomment{I changed $(C_0,C_1)$ to $(\eta_0,\eta_1)$ since it is also one of the Greek letters which is not used in the previous draft!}
%
%\rgcomment{OK. Maybe it's just me but it does feel a bit strange, because here the values $\eta_0,\eta_1$ are large constants or polylogs, but when I see an $\eta$ I usually feel it is a small number. It would be nice if there is some capitalized letter that we can still use.}

\begin{definition}[$(\eta_0,\eta_1)$-nice] \label{def:nice-property}
Suppose $$ \max\{\|A\|,\|B\|,\|C\|\} \leq \eta_1 \sqrt{\frac{k}{d}}.$$
Given an approximate solution $\{\h{A}, \h{B}, \h{C}, \h{w}\}$, we call it $(\eta_0,\eta_1)$-nice if matrix $\h{A}$ (similarly $\h{B}$ and $\h{C}$) satisfies 
\begin{align*}
\|\dt{A}_i\| := \|\h{a}_i - a_i\| & \le \eta_0 \frac{\sqrt{k}}{d}, \quad \forall i \in [k], \\
\|\h{A}\| &\le \eta_1\sqrt{\frac{k}{d}},
\end{align*}
and the weights satisfy $$|\h{w}_i-w_i|\le \eta_0 w_{\max} \frac{\sqrt{k}}{d}.$$
\end{definition}
%(the true weights $w_i$ satisfy $w_i \ge 1$ and $w_i \le R = O(1)$) \mjcomment{Is this also part of the definition? (As the bound on the spectral norm of true matrices!) It might be better to put these as assumption before the definition, not as part of the definition.}

%\rgcomment{This is not a part of definition. In fact, I would suggest when we introduce the tensor problem say without loss of generality assume $w_i\ge 1$ and $w_i \le R$. An alternative way of doing things would be to add a $w_{\max}$ factor to all the equations that appear in the definition (which is roughly what we are doing now, but it makes the math looks more complicated).}

%Note that the optimization program in~\eqref{eqn:coordinate descent-opt} is proposed to ensure the above bound on $\|\hA\|$ is satisfied on the input of this part of algorithm. 
Given above conditions are satisfied, we prove the following guarantees for removing residual error, Algorithm~\ref{algo:coordinate-descent}.

\begin{lemma}[Local convergence guarantee of the iterations for removing residual error, Algorithm~\ref{algo:coordinate-descent}] \label{thm:coordinate descent}
Consider $T$ as the input to Algorithm~\ref{algo:coordinate-descent}, where $T$ is a rank-$k$ tensor. Suppose Assumptions \ref{cond:rank-k}-\ref{cond:size} and \ref{cond:spectral bound KR} hold (which are satisfied whp when the components are uniformly i.i.d.\ drawn from unit $d$-dimensional sphere).
Given initial solution $\left\{\h{A}^{(0)}, \h{B}^{(0)}, \h{C}^{(0)}, \h{w}^{(0)} \right\}$ which is $(\eta_0,\eta_1)$-nice, all the following iterations of Algorithm~\ref{algo:coordinate-descent} are $(2\eta_0, 3\eta_1)$-nice. Furthermore, given the exact tensor $T$, the Frobenius norm error $\max\{\|\dt{A}\|_F, \|\dt{B}\|_F, \|\dt{C}\|_F, \|\dt{w}\|/w_{\min}\}$ shrinks by at least a factor of 2 in every iteration. In addition, if we have a noisy tensor $\hT = T + \Psi$ such that $\|\Psi\| \le \psi$, then
$$
\max\{\|\dt{A}^{(t)}\|_F, \|\dt{B}^{(t)}\|_F, \|\dt{C}^{(t)}\|_F, \|\dt{w}^{(t)}\|/w_{\min}\}
\le 2^{-t} \eta_0 \frac{ k}{d} + O\left(\frac{\psi \sqrt{k}}{w_{\min}}\right).
$$
\end{lemma}

\subsection*{Proof: iteration for removing residual error in Algorithm~\ref{algo:coordinate-descent}}

We now prove Lemma~\ref{thm:coordinate descent} as the local convergence guarantee of the iterations for removing residual error, Algorithm~\ref{algo:coordinate-descent}.

To prove this lemma, we first observe that the algorithm update formula in~\eqref{eqn:BiasRemoval} is (before normalization)
$w_i\inner{a_i,\h{a}_i}\inner{b_i,\h{b}_i} c_i + \epsilon_i$ 
where 
$$\epsilon_i = \sum_{j\ne i}(w_i\inner{a_j,\h{a}_i}\inner{b_j,\h{b}_i} c_j- \h{w}_i \inner{\h{a}_i,\h{a}_j}\inner{\h{b}_i,\h{b}_j}\h{c}_j).$$

In the following lemma, we show that the error terms $\epsilon_i$'s are small.

\begin{lemma} \label{lem:aux coordinate descent}
Before normalization $\tl{w}_i\tl{c}_i = w_i \inner{a_i,\h{a}_i}\inner{b_i,\h{b}_i} c_i + \epsilon_i$ where 
$$\sum_{i=1}^k \|\epsilon_i\|^2 \le o(1)(w_{\max}(\|\dt(A)\|_F^2+\|\dt(B)\|_F^2+\|\dt(C)\|_F^2)+\|\dt{w}\|^2).$$
%\mjcomment{Many equations like this would be nicer not to be inline!}
%\rgcomment{Sure, change whichever you like}
\end{lemma}

\bprf
By the update formula in~\eqref{eqn:BiasRemoval}, we know $$\epsilon_i = \sum_{j\ne i}(w_i\inner{a_j,\h{a}_i}\inner{b_j,\h{b}_i} c_j- \h{w}_i\inner{\h{a}_i,\h{a}_j}\inner{\h{b}_i,\h{b}_j}\h{c}_j).$$ 
We expand it into several terms as follows.
\begin{align*}
\epsilon_i & = \sum_{j\ne i}(w_i\inner{a_j,\h{a}_i}\inner{b_j,\h{b}_i} c_j- \h{w}_i\inner{\h{a}_i,\h{a}_j}\inner{\h{b}_i,\h{b}_j}\h{c}_j)\\
& = \sum_{j\ne i} \inner{a_i, a_j}\inner{b_i, b_j} (w_jc_j - \h{w}_j\h{c}_j)  \quad \mbox{(type 1)}\\
&\quad + \sum_{j\ne i}w_j\inner{a_j,\dt{A}_i}\inner{b_j,b_i}c_j 
+ \sum_{j\ne i}w_j\inner{a_j,a_i}\inner{b_j,\dt{B}_i}c_j\quad \mbox{(type 2)}\\
&\quad - \sum_{j\ne i}\h{w}_j\inner{a_j,a_i}\inner{b_j,\dt{B}_i}\h{c}_j- \sum_{j\ne i}\h{w}_j\inner{a_j,a_i}\inner{\dt{B}_j,\h{b}_i}\h{c}_j\\
&\quad - \sum_{j\ne i}\h{w}_j\inner{a_j,\dt{A}_i}\inner{b_j,b_i}\h{c}_j - \sum_{j\ne i}\h{w}_j\inner{\dt{A}_j,\h{a}_i}\inner{b_j,b_i}\h{c}_j\\
&\quad + \sum_{j\ne i}\inner{a_j,\dt{A}_i}\inner{b_j,\dt{B}_i}c_j\quad \mbox{(type 3)} \\
& \quad - \sum_{j\ne i}\h{w}_j\inner{a_j,\dt{A}_i}\inner{b_j,\dt{B}_i}\h{c}_j- \sum_{j\ne i}\h{w}_j\inner{\dt{A}_j,\h{a}_i}\inner{b_j,\dt{B}_i}\h{c}_j\\
& \quad - \sum_{j\ne i}\h{w}_j\inner{a_j,\dt{A}_i}\inner{\dt{B}_j,\h{b}_i}\h{c}_j- \sum_{j\ne i}\h{w}_j\inner{\dt{A}_j,\h{a}_i}\inner{\dt{B}_j,\h{b}_i}\h{c}_j.
\end{align*}

The norm of three different types of terms mentioned above are bounded in Section~\ref{sec:aux claims-coordinate descent}, which conclude the desired bound in the lemma.
%There are one term of type 1, 6 terms of type 2 and 5 terms of type 3. Since all the terms can be bounded by the three claims above, we have the desired bound in the lemma.
\eprf

We are now ready to prove main Lemma~\ref{thm:coordinate descent}.

\bprfof{Lemma~\ref{thm:coordinate descent}}
Since $\tl{w}_i$ is the norm of $ w_i \inner{a_i,\h{a}_i}\inner{b_i,\h{b}_i} c_i + \epsilon_i$, we know $$|\tl{w}_i - w_i| \le  \|\epsilon_i\| + w_i (\Theta(\|\dt{A}_i\|^2+\|\dt{B}_i\|^2)),$$ and therefore 
$$\|\tl{w} - w\| \le o(1)(w_{\max}(\|\dt(A)\|_F+\|\dt(B)\|_F+\|\dt(C)\|_F)+\|\dt{w}\|).$$

On the other hand, since the coefficient $w_i \inner{a_i,\h{a}_i}\inner{b_i,\h{b}_i}$ is at least $1-o(1)$, we know $\|\tl{c}_i - c_i\| \le 4\|\epsilon_i\|/w_{\min}$. %\rgcomment{Here the factor 4 is something arbitrary, the way to prove is as follows: if $\|\epsilon\| < 1/2$ then the normalization factor is at most 2; if $\|\epsilon\|> 1/2$ then two unit vectors can differ by at most 2 anyways}. 
This implies $$\|\tl{C}- C\|_F \le o(1)((\|\dt(A)\|_F+\|\dt(B)\|_F+\|\dt(C)\|_F)+\|\dt{w}\|/w_{\min}).$$

By Lemma~\ref{thm:fix}, we know after the projection procedure, we get $\|\widehat{C} - C\|_F \le 2\|\tilde{C}-C\|_F$. Therefore combining the two steps we know
$$
\|\widehat{C} - C\|_F \le 2\|\tilde{C}-C\|_F \le o(1)(\|\dt(A)\|_F+\|\dt(B)\|_F+\|\dt(C)\|_F+\|\dt{w}\|/w_{\min}).
$$

When we have noise, all the $\epsilon_i$'s have an additional term $\Psi(\h{a}_i,\h{b}_i,I)$ which is bounded by $\psi$, and thus, the second part of the lemma follows directly.

%After the fixing step, by the constraints we always have the new solution is $(2\eta_0,\eta_1)$-nice. On the other hand, observe that the correct solution $\{A,B,C\}$ is always a feasible solution, so the Frobenius norm increases by at most a factor of 2. 

%\mjcomment{It would useful to refer to the lemma where this is proved, and also emphasize and clarify this is only for the fix step, and combining this with the shrinkage result in Lemma 1, we get the overall shrinkage result in Theorem 1. This was the place where I was also confused in the first read!}
%\rgcomment{In fact what we are doing is a projection to a convex set so the Frobenius norm never increases. This however will not be true once we have the more efficient fixing algorithm}
\eprfof

\paragraph{Handling Symmetric Tensors:}
For symmetric tensors we should change the algorithm as computing the following:
\begin{equation} \label{eqn:BiasRemoval-symmetric}
T(\h{a}_i,\h{b}_i,I) - \frac{1}{d}\sum_{i=1}^d T(e_i,e_i,I) - \sum_{j\ne i} \h{w}_j(\inner{\h{a}_i,\h{a}_j}\inner{\h{b}_i,\h{b}_j}-\frac{1}{d})\h{c}_j.
\end{equation}

The result of this will be a change in the term of type 1. Now the Q matrix will be $(A\odot A)^T(A\odot A) - (1-\frac{1}{d})I - \frac{1}{d} J$ which has desired spectral norm for random matrices.

%\rgcomment{To do: noisy case (very easy)}

\subsubsection*{Claims for proving Lemma~\ref{lem:aux coordinate descent}} \label{sec:aux claims-coordinate descent}

The first term deals with the difference between $C$ and $\h{C}$.

\begin{claim}  We have
$$\sqrt{\sum_{i=1}^k \| \sum_{j\ne i} \inner{a_i, a_j}\inner{b_i, b_j} (w_ic_i - \h{w}_i\h{c}_i)\|^2} \le o(1) (w_{\max}\|\dt{C}\|_F+\|\h{w}-w\|).$$
\end{claim}

\bprf
This sum is equal to the Frobenius norm of a matrix $M = Q Z$. Here the matrix $Q$ is a matrix such that is equal to $Q = (A\odot B)^\top (A\odot B) - I$: %has off-diagonal entries $Q_{i,j} = \inner{a_i, a_j}\inner{b_i, b_j}$ when %$i\ne j$ and diagonal entries $Q_{i,i} = 0$ (therefore $Q = (A\odot B)^\top (A\odot B) - I$). \mjcomment{For instance, I present $Q$ as follows. It makes easier for the reader to follow: Here $Q$ is a diagonal matrix as
\begin{align*}
Q_{i,j} = 
\left\{\begin{array}{ll}
\inner{a_i, a_j}\inner{b_i, b_j}, & i \neq j, \\
0, & i=j,
\end{array}\right.
\end{align*}
%and therefore, $Q = (A\odot B)^\top (A\odot B) - I$.}
%\rgcomment{That's good, please do so for other cases as well.}
The matrix $Z$ has columns $Z_i = w_ic_i - \h{w}_i\h{c}_i$. By assumption we know $\|Q\| \le o(1)$, and  $\|Z\|_F \le w_{\max}\|\dt{C}\|_F + \|\h{w} - w\|$. Therefore we have
$$
\|M\|_F = \|QZ\|_F \le \|Q\|\|Z\|_F \le o(1) ( w_{\max}\|\dt{C}\|_F + \|\h{w} - w\|).
$$
% \mjcomment{It is useful to mention $\|M\|_F$ is bounded by $\|Q\| \cdot \|Z\|_F$.}
% \rgcomment{Right, see the new equation above}
\eprf

Of course, in the error $\epsilon_i$, we don't have $\sum_{j\ne i} \inner{a_i, a_j}\inner{b_i, b_j} w_ic_i$, instead we have terms like $\sum_{j\ne i} \inner{\h{a}_i, a_j}\inner{\h{b}_i, b_j} w_ic_i$. The next two lemmas show that these two terms are actually very close.

\begin{claim} We have
$$\sqrt{\sum_{i=1}^k \| \sum_{j\ne i} \inner{\dt{A}_i, \h{a}_j}\inner{b_i, b_j} \h{w}_i\h{c}_i\|^2}\le o(w_{\max}) \|\dt{A}\|_F.$$
$$\sqrt{\sum_{i=1}^k \| \sum_{j\ne i} \inner{\dt{A}_j, \h{a}_i}\inner{b_i, b_j} \h{w}_i\h{c}_i\|^2}\le o(w_{\max}) \|\dt{A}\|_F.$$
Same is true if any $\h{\cdot}$ is replaced by the true value.
\end{claim}

\bprf
Similar as before, we treat the left hand side as the Frobenius norm of some matrix $M = QZ$. Here $Z_i = \h{w}_i\h{c}_i$, and $Q$ is the following matrix:
\begin{align*}
Q_{i,j} = 
\left\{\begin{array}{ll}
\inner{\dt{A}_i, \h{a}_j}\inner{b_i, b_j}, & i \neq j, \\
0, & i=j,
\end{array}\right.
\end{align*}
We shall bound $\|M\|_F$ by $\|Z\|\|Q\|_F$. By assumption we know $\|Z\| \le w_{\max}\cdot 2\eta_1 \sqrt{k/d} = O(w_{\max}\sqrt{k/d})$. On the other hand, we know $\inner{b_i,b_j} \le \tl{O}(1/\sqrt{d})$ hence $\|Q\|_F\le \tl{O}(1/\sqrt{d}) \|\h{A}^T\dt{A}\|_F \le \tl{O}(1/\sqrt{d}) \|\h{A}\| \|\dt{A}\|_F = \tl{O}(\sqrt{k}/d) \|\dt{A}\|_F$. Therefore we have 
$$
\|M\|_F \le \|Z\|\|Q\|_F \le O(w_{\max}\sqrt{k/d})\cdot \tl{O}(w_{\max}\sqrt{k}/d)\|\dt{A}\|_F = \tilde{O}(k/d\sqrt{d}) \|\dt{A}\|_F = o(w_{\max})\|\dt{A}\|_F.
$$
Notice that the proof works for both terms. 
\eprf

\begin{claim} We have
$$\sqrt{\sum_{i=1}^k \| \sum_{j\ne i} \inner{\dt{A}_i, \h{a}_j}\inner{\dt{B}_i, \h{b}_j} \h{w}_i\h{c}_i\|^2}\le o(w_{\max}) (\|\dt{A}\|_F+\|\dt{B}\|_F).$$
The same is true if the inner-products are between $\inner{\dt{A}_j, \h{a}_i}$ or $\inner{\dt{B}_j, \h{b}_i}$, or if any $\h{\cdot}$ is replaced by the true value.
\end{claim}

\bprf
Similar as before, we treat the left hand side as the Frobenius norm of some matrix $M = QZ$. Here $Z_i = \h{w}_i\h{c}_i$, and $Q$ is the following matrix
\begin{align*}
Q_{i,j} = 
\left\{\begin{array}{ll}
\inner{\dt{A}_i, \h{a}_j}\inner{\dt{B}_i, b_j}, & i \neq j, \\
0, & i=j,
\end{array}\right.
\end{align*}
Now using definition of $2\to 4$ norm and $2ab \le a^2 + b^2$ we first bound the Frobenius norm of the matrix $Q$: %\mjcomment{Good to state what inequalities are used here!}
$$
\sum_{i\ne j} (\inner{\dt{A}_i, \h{a}_j}\inner{\dt{B}_i, \h{b}_j})^2 \le 
\sum_{i\ne j} (\inner{\dt{A}_i, \h{a}_j})^4 + (\inner{\dt{B}_i, \h{b}_j})^4
\le \sum_{i=1}^k \|\h{A}^\top\|_{2\to 4} \|\dt{A}_i\|^4+\|\h{B}^\top\|_{2\to 4} \|\dt{B}_i\|^4
$$

Now we first bound the $2\to 4$ norm of the matrix $\h{A}^\top = A^\top + \dt{A}^\top$. By assumption we already know $\|A^\top\|_{2\to 4} \le O(1)$. On the other hand, for any unit vector $u$
$$
\sum_{i=1}^k \inner{\dt{A}_i,u}^4 \le \max_{i=1}^k \inner{\dt{A}_i,u}^2 \sum_{i=1}^k \inner{\dt{A}_i,u}^2 \le \tilde{O}(k^2/d^3) = o(1).
$$
Here we used the assumption that $\|\dt{A}_i\| \le \tilde{O}(\sqrt{k}/d)$ and $\|\dt{A}\| \le O(\sqrt{k/d})$. Therefore $\|\h{A}^\top \|_{2\to 4} \le \|A^\top \|_{2\to 4}+\|\dt{A}^\top \|_{2\to 4} \le O(1)$ (and similarly for $\h{B}^\top$).

Therefore
\begin{align*}
\|Q\|_F & \le \sqrt{\sum_{i=1}^k \|\h{A}^\top\|_{2\to 4} \|\dt{A}_i\|^4+\|\h{B}^\top\|_{2\to 4} \|\dt{B}_i\|^4} \\
& \le O(1)\sqrt{\sum_{i=1}^k  \|\dt{A}_i\|^4+\|\dt{B}_i\|^4} \\
& \le O(1)\cdot \max_{i=1}^k(\|\dt{A}\|_i+\|\dt{B}\|_i)\sqrt{\sum_{i=1}^k  \|\dt{A}_i\|^2+\|\dt{B}_i\|^2} \\
& \le \tilde{O}(\sqrt{k}/d) (\|\dt{A}\|_F+\|\dt{B}\|_F).
\end{align*}

On the other hand we know $\|Z\| \le O(w_{\max}\sqrt{k/d})$, hence $\|M\|_F \le \|Z\|\|Q\|_F \le o(w_{\max}) (\|\dt{A}\|_F+\|\dt{B}\|_F)$.

\eprf

\subsubsection*{Projection Procedure~\ref{algo:fix-procedure}}

In this section, we describe the functionality of projection Procedure~\ref{algo:fix-procedure}.
Suppose the initial solution $\{\h{A}^0, \h{B}^0, \h{C}^0, \h{w}^0\}$ is $(\eta_0,\eta_1)$-nice. Then, given an arbitrary solution $\{\tl{A},\tl{B},\tl{C},\tl{w}\}$, we run projection Procedure~\ref{algo:fix-procedure} to get a $(2\eta_0, 4\eta_1)$-nice solution without losing too much in Frobenius norm error. This is shown in the following Lemma.

%\begin{enumerate}
%\item Compute the SVD of $\tl{A} = UDV^\top$.
%\item Let $\h{D}$ be the truncated version of $D$: $\h{D}_{i,i} = \min\{D_{i,i},\eta_1\sqrt{k/d}\}$.
%\item Compute $Q = U\h{D}V^\top$, for each column if $\|Q_i-\h{A}^0_i\| \le \eta_0\sqrt{k}/d$ then let $\h{a}_i = Q_i$, otherwise let $\h{a}_i = \h{A}^0_i +\frac{\eta_0\sqrt{k}}{d} \cdot \frac{Q_i-\h{A}^0_i}{\|Q_i-\h{A}^0_i\|}$.
%\item Repeat to get $\h{B}$, $\h{C}$.
%\item Let $\h{w}_i'$ be $\tl{w}_i$ unless $|\tl{w}_i - \h{w}^0_i| > \eta_0\sqrt{k}/d$, in that case make sure $|\h{w}_i' - \h{w}^0_i| = \eta_0\sqrt{k}/d$, $h{w}_i'$ and $\tl{w}_i$ are on the same side of $\h{w}^0_i$.
%\end{enumerate}

\begin{lemma}
\label{thm:fix} Suppose the initial solution $\{\h{A}^0, \h{B}^0, \h{C}^0, \h{w}^0\}$ is $(\eta_0,\eta_1)$-nice. For any solution $\{\tl{A},\tl{B},\tl{C},\tl{w}\}$, let error $E = \max\{\|\tl{A}-A\|_F,\|\tl{B}-B\|_F,\|\tl{C}-C\|_F, \|\tl{w}-w\|/w_{\min}\}$. Then after the projection Procedure~\ref{algo:fix-procedure}, the new solution is $(2\eta_0, 3\eta_1)$-nice and has error at most $2E$.
\end{lemma}

\bprf
Intuitively, by truncating $D$ the matrix we get is closest to $\tl{A}$ among matrices with spectral norm $\eta_1\sqrt{k/d}$. We first prove this fact:
\begin{claim}
$$\|Q - \tl{A}\|_F = \min_{\|M\|\le \eta_1\sqrt{k/d}} \|M-\tl{A}\|_F.$$
\end{claim}
\bprf
By symmetric properties of Frobenius and spectral norm (both are invariant under rotation), we can rotate the matrices $Q,M,\tl{A}$ simultaneously, so that $\tl{A}$ becomes a diagonal matrix $D$. Since $M$ has spectral norm bounded by $\eta_1\sqrt{k/d}$, in particular all its entries must be bounded by $\eta_1\sqrt{k/d}$. Also, we know $\|D-\widehat{D}\|_F = \min_{\forall (i,j) M_{i,j} \le \eta_1\sqrt{k/d}} \|D-M\|_F$, therefore $\|D-\widehat{D}\|_F = \min_{\|M\|\le \eta_1\sqrt{k/d}} \|D-M\|_F$. By the rotation invariant property this implies the claim.
\eprf

Since the optimal solution $A$ has spectral norm bounded by $\eta_1\sqrt{k/d}$, in particular from above claim we know $\|Q-\tl{A}\|_F\le \|\tl{A} - A\|_F$. By triangle inequality we get $\|Q-A\|_F \le 2E$. In the next step we are essentially projecting the solution $Q$ to a convex set that contains $A$ (the set of matrices that are column-wise $\eta_1\sqrt{k}/d$ close to $\h{A}^0$), so the distance can only decrease. Similar arguments work for $\h{B},\h{C},\h{w}$, therefore the error of the new solution is bounded by $2E$.

By construction it is clear that the columns of the new solution is within $\eta_0\sqrt{k}/d$ to the columns of the initial solution, so they must be within $2\eta_0\sqrt{k}/d$ to the columns of the true solution. The only thing left to prove is that $\|\h{A}\| \le 3\eta_1\sqrt{k/d}$.

First we observe that $\h{A} = \h{A}^0 + Z$ where $Z$ is a matrix whose columns are multiples of $Q-\h{A}^0$, and the multiplier is never larger than 1. Therefore $\|\h{A}\| \le \|h{A}^0\|+\|Z\|\le \|\h{A}^0\|+\|Q-\h{A}^0\|\le 2\|\h{A}^0\|+\|Q\| \le 3\eta_1\sqrt{k/d}$.
\eprf

\section{SVD Initialization Result} \label{sec:initialization}

In this section, we analyze the SVD-based initialization technique proposed in Procedure~\ref{algo:SVD init}. The goal is to provide good initialization vectors close to the columns of true components $A$ and $B$ in the regime of $k=O(d)$.
%Tensor $T \in \Rbb^{d \times d \times d}$ acts as a linear operator $T(I,I,\theta):\Rbb^d \rightarrow \Rbb^{d \times d}$ as
%\begin{align*}
%T(I,I,\theta)_{i,j} = \sum_{l \in [d]} \theta_l T_{i,j,l}.
%\end{align*}

Given a vector $\theta \in \Rbb^d$, matrix $T(I,I,\theta)$ results a linear combination of slices of tensor $T$. For tensor $T$ in \eqref{eqn:true-T}, we have
\begin{align} \label{eqn:T(theta) expansion}
T(I,I,\theta) = \sum_{i\in [k]} \wstar_i \langle \theta,\cstar_i \rangle \astar_i \bstar_i^\top = \sum_{i\in [k]} \lambdastar_i \astar_i \bstar_i^\top = \Astar \Diag(\lambdastar) \Bstar^\top,
\end{align}
where $\lambdastar_i := \wstar_i \langle \theta,\cstar_i \rangle, i \in [k]$, and $\lambdastar :=[\lambdastar_1,\lambdastar_2,\dotsc,\lambdastar_k]^\top \in \Rbb^k$ is expressed as
$$
\lambdastar = \Diag(\wstar) \Cstar^\top \theta.
$$
Since $\Astar$ and $\Bstar$ are not orthogonal matrices, the expansion in \eqref{eqn:T(theta) expansion} is not the SVD\,\footnote{Note that if $A$ and $B$ are orthogonal matrices, columns of $A$ and $B$ are directly recovered by computing SVD of $T(I,I,\theta)$.} of $T(I,I,\theta)$. But, we show in the following theorem that if  we draw enough number of random vectors $\theta$ in the regime of $k=O(d)$, we can eventually provide good initialization vectors through SVD of $T(I,I,\theta)$. \\
Define
$$
g(L) := \sqrt{2 \ln(L)} - \frac{\ln(\ln(L))+c}{2\sqrt{2 \ln(L)}} - \sqrt{2 \ln(k)}.
$$

%\begin{lemma}
%Let $\theta \in \Rbb^d$ is a random Gaussian vector $\theta \sim \mathcal{N}(0,I)$. Then, $\lambdastar$ is a random Gaussian vector $\lambdastar \sim \mathcal{N}(0,\Sigma_\lambda)$ where $\Sigma_\lambda = \Diag(\wstar) \Cstar^\top \Cstar \Diag(\wstar)^\top$.
%\end{lemma}
%\bprf
%Vector $\lambdastar$ can be written as
%\begin{align*}
%\lambdastar = \Diag(\wstar) C^\top \theta.
%\end{align*}
%\eprf

%In the following propositions, we provide initialization results under two cases when $C$ is orthogonal and $C$ is non-orthogonal.

\begin{theorem} [SVD initialization when $k=O(d)$] \label{thm:initialization}
Consider tensor $\hT = T + \Psi$ where $T$ is a rank-$k$ tensor, and $\Psi$ is a perturbation tensor. Let Assumptions \ref{cond:rank-k}-\ref{cond:spectral bound} hold and $k=O(d)$. %Let $\rho := \max_{i \neq j} | \langle c_i,c_j \rangle |$ denote the incoherence parameter of matrix $C$.
Draw $L$ i.i.d. random vectors $\theta^{(j)} \sim \mathcal{N}(0,I_d), j \in [L]$.
Let $u_1^{(j)}$ and $v_1^{(j)}$ be the top left and right singular vectors of $\hT(I,I,\theta^{(j)})$. This is $L$ random runs of Procedure~\ref{algo:SVD init}. Suppose $L$ satisfies the bound
$$
g(L)
\geq \frac{w_{\max}(1+\mu)}{w_{\min}-\rho w_{\max}(1+\mu)} 4 \sqrt{\log k},
$$
with $\mu = \frac{2\mu_R +\tl{\mu} - 1}{1-\tl{\mu}} < \frac{w_{\min}}{w_{\max} \rho}-1$, for $\mu_R$ and $\mu_{\min}$ defined in \eqref{eqn:mu_E & m_1}, and some $0<\tl{\mu}<1$. Note that $\rho \leq \frac{\alpha}{\sqrt{d}}$ is also defined as the incoherence parameter in Assumption \ref{cond:incoherence}. Then, w.h.p., %with probability greater than %$\frac{3}{4} - k^{-7}$ $\frac{1}{2}$,
at  least one of the pairs $(u_1^{(j)},v_1^{(j)}), j \in [L]$, say $j^*$, satisfies
$$
\max \left\{ \dist \left(u_1^{(j^*)},a_1 \right), \dist \left(v_1^{(j^*)},b_1 \right) \right\}
%\max \left\{ \sqrt{1- \langle u_1,\astar_1 \rangle^2}, \sqrt{1- \langle v_1,\bstar_1 \rangle^2} \right\}
\leq \frac{4 w_{\max} \mu_{\min} (1+\rho) \sqrt{\log k} + \alpha_0 \sqrt{d} \psi}{w_{\min} \tl{\mu} g(L) - \alpha_0 \sqrt{d} \psi},
$$
where $\psi := \|\Psi\|$ is the spectral norm of perturbation tensor $\Psi$, and $\alpha_0>1$ is a constant.
%for $\alpha = \polylog(d)$.
\end{theorem}

%\begin{theorem} [SVD initialization in undercomplete and noiseless setting] \label{thm:initialization}Consider undercomplete ($k \leq d$) and noiseless tensor $T$ satisfying Assumptions \ref{cond:rank-k}-\ref{cond:spectral bound}. Let $\rho := \max_{i \neq j} | \langle c_i,c_j \rangle |$ denote the incoherence parameter of matrix $C$.
%Draw $L$ i.i.d. random vectors $\theta^{(j)} \sim \mathcal{N}(0,I_d), j \in [L]$.
%Let $u_1^{(j)}$ and $v_1^{(j)}$ be the top left and right singular vectors of $T(I,I,\theta^{(j)})$. This is $L$ random runs of Procedure~\ref{algo:SVD init}. Suppose $L$ satisfies the bound
%\begin{align*}
%\sqrt{\frac{\ln(L)}{8 \ln(k)}} \left( 1 - \frac{\ln(\ln(L))+c}{4\ln(L)} - \sqrt{\frac{\ln(8)}{\ln(L)}} \right)
%\geq \frac{w_{\max}(1+\mu)}{w_{\min}-\rho w_{\max}(1+\mu)},
%\end{align*}
%with $\mu = \mu_E+\mu_R+\tl{\mu} < \frac{w_{\min}}{w_{\max} \rho}-1$, for $\mu_E$ and $\mu_R$ defined in \eqref{eqn:mu_E & m_1}, and some $\tl{\mu}>0$. Then, with probability greater than %$\frac{3}{4} - k^{-7}$
%$\frac{1}{2}$, at  least one of the pairs $(u_1^{(j)},v_1^{(j)}), j \in [L]$, say $j^*$, satisfies
%\begin{align*}
%\max \left\{ \dist \left(u_1^{(j^*)},a_1 \right), \dist \left(v_1^{(j^*)},b_1 \right) \right\}
%%\max \left\{ \sqrt{1- \langle u_1,\astar_1 \rangle^2}, \sqrt{1- \langle v_1,\bstar_1 \rangle^2} \right\}
%\leq \frac{\mu_E}{\tl{\mu}}.
%\end{align*}
%\end{theorem}
%
%The result is proved by applying Lemmata \ref{lem:gap condition} and \ref{lem:initialization}.

\bprf
Let $\lambda^{(j)} := \Diag(w) \Cstar^\top \theta^{(j)} \in \Rbb^k$ and $\tl{\lambda}^{(j)} := \Cstar^\top \theta^{(j)} \in \Rbb^k$.
From Lemmata \ref{lem:gap condition} and \ref{lem:initialization}, there exists a $j^* \in [L]$ such that w.h.p.,
%with probability at least $1/2$,
we have
$$
\max \left\{ \dist \left( u_1^{(j^*)},a_1 \right), \dist \left( v_1^{(j^*)},b_1 \right) \right\}
%\max \left\{ \sqrt{1- \langle u_1,\astar_1 \rangle^2}, \sqrt{1- \langle v_1,\bstar_1 \rangle^2} \right\}
\leq \frac{\mu_{\min} \lambdastar_{(2)} + \| \Psi(I,I,\theta) \|}{\tl{\mu} \lambdastar_1 - \| \Psi(I,I,\theta) \|}.
$$
From \eqref{eqn:gap proof lower bound}, with probability at least $1-2k^{-1}$,  we have
$$
\lambda^{(j^*)}_1 \geq w_{\min} g(L).
$$
From \eqref{eqn:gap proof upper bound}, with probability at least $1-k^{-7}$, we have
$$
\lambda_{(2)}^{(j^*)} \leq w_{\max} \left( \rho \tl{\lambda}_1^{(j^*)} + 4 \sqrt{\log k} \right) \leq 4 w_{\max} (1+\rho) \sqrt{\log k},
$$
where in the last inequality, we also applied upper bound on $\tl{\lambda}_1^{(j^*)}$. Combining all above bounds and Lemma \ref{lem:Noise tensor norm} finishes the proof.
\eprf

\subsection{Auxiliary lemmata for initialization}

In the following Lemma, we show that the gap condition between the maximum and the second maximum of vector $\lambda$ required in Lemma \ref{lem:initialization} is satisfied under some number of random draws.

\begin{lemma}[Gap condition] \label{lem:gap condition}
Consider an arbitrary matrix $C \in \Rbb^{d \times k}$ with unit-norm columns which also satisfies incoherence condition $\max_{i \neq j} | \langle c_i,c_j \rangle | \leq \rho$ for some $\rho>0$. Let
$$
\lambdastar := \Diag(w) \Cstar^\top \theta \in \Rbb^k,
$$
denote the vector that captures correlation of $\theta \in \Rbb^d$ with columns of $C$. Without loss of generality, assume that $\lambdastar_1 = \max_i |\lambdastar_i|$, and let $\lambdastar_{(2)} := \max_{i \neq 1} |\lambdastar_i|$.
Draw $L$ i.i.d. random vectors $\theta^{(j)} \sim \mathcal{N}(0,I_d), j \in [L]$, and $\lambdastar^{(j)} := \Diag(w) \Cstar^\top \theta^{(j)}$. Suppose $L$ satisfies the bound
$$
\sqrt{\frac{\ln(L)}{8 \ln(k)}} \left( 1 - \frac{\ln(\ln(L))+c}{4\ln(L)} - \sqrt{\frac{\ln(k)}{\ln(L)}} \right)
\geq \frac{w_{\max}(1+\mu)}{w_{\min}-\rho w_{\max}(1+\mu)},
$$
for some $0<\mu< \frac{w_{\min}}{w_{\max} \rho}-1$.
Then, with probability at least $1 - 2k^{-1} - k^{-7}$, we have the following  gap condition for at least one draw, say $j^*$,
$$
%\frac{\lambdastar_1^{(j^*)}  - \lambdastar_{(2)}^{(j^*)}}{\lambdastar_{(2)}^{(j^*)}} \geq \mu. \\
\lambdastar_1^{(j^*)}   \geq (1 + \mu) \lambdastar_{(2)}^{(j^*)}.
$$
%for some $\mu>0$.
\end{lemma}
\bprf
%Without loss of generality assume that $w_1$ is the maximum among all weights $w_i,i \in [k]$.
Define $\tl{\lambda} := \Diag(w)^{-1} \lambda = C^\top \theta$. We have $\lambda_j = w_j \tl{\lambda}_j, j \in [k]$. \\
Each vector $\tl{\lambda}^{(j)}$ is a random Gaussian vector $\tl{\lambda}^{(j)} \sim \mathcal{N}(0,C^\top C)$.
Let $j^* := \argmax_{j \in [L]} \tl{\lambda}_1^{(j)}$. Since $\max_{j \in [L]} \tl{\lambda}^{(j)}_1,$ is a 1-Lipschitz function of $L$ independent $\mathcal{N}(0,1)$ random variables, similar to the analysis in Lemma B.1 of~\citet{AnandkumarEtal:tensor12}, we have
\begin{align} \label{eqn:gap proof lower bound}
\Pr \left[ \tl{\lambda}^{(j^*)}_1 \geq \sqrt{2 \ln(L)} - \frac{\ln(\ln(L))+c}{2\sqrt{2 \ln(L)}} - \sqrt{2 \ln(k)} \right] \geq 1 - \frac{2}{k}.
\end{align}
%We have $\tl{\lambda}_i := \langle \theta,\cstar_i \rangle, i \in [k]$.
Any vector $c_i, i \neq 1,$ can be decomposed to two components parallel and perpendicular to $c_1$ as $c_i = \langle c_i,c_1 \rangle c_1 + \Pc_{\perp_{c_1}}(c_i)$. Then, for any  $\tl{\lambda}_i, i \neq 1$, we have
$$
\tl{\lambda}_i := \langle \theta,\cstar_i \rangle
= \underbrace{\theta^\top \langle c_i,c_1 \rangle c_1}_{=: \tl{\lambda}_{i,\parallel}}
+ \underbrace{\theta^\top \Pc_{\perp_{c_1}}(c_i)}_{=: \tl{\lambda}_{i,\perp}}.
$$
Since $\Pc_{\perp_{c_1}}(c_i) \perp c_1, i \neq 1$, we have $\tl{\lambda}_{i,\perp}, i \neq 1,$ are independent of $\tl{\lambda}_1 := \theta^\top c_1$, and therefore, the following bound can be argued independent of bound in \eqref{eqn:gap proof lower bound}. From Lemma \ref{lem:Gaussian max bound}, we have
$$
\Pr \left[ \max_{i \neq 1} \tl{\lambda}^{(j^*)}_{i,\perp} \geq 4 \sqrt{\log k} \right] \leq k^{-7}.% \leq \frac{1}{4}.
$$
For $\tl{\lambda}_{i,\parallel}$, we have
$$
\tl{\lambda}_{i,\parallel} = \theta^\top \langle c_i,c_1 \rangle c_1 \leq  \rho \theta^\top c_1 = \rho \tl{\lambda}_1,
$$
where we also assumed that $\tl{\lambda}_1 := \theta^\top c_1>0$ which is true for large enough $L$, concluded from \eqref{eqn:gap proof lower bound}. By combining above two bounds, with probability at least $1-k^{-7}$, we have
\begin{align} \label{eqn:gap proof upper bound}
\tl{\lambda}_{(2)}^{(j^*)} \leq \rho \tl{\lambda}_1 + 4 \sqrt{\log k}.
\end{align}
From the given bound on $L$ in the lemma and inequalities \eqref{eqn:gap proof lower bound} and \eqref{eqn:gap proof upper bound}, with probability at least $1 - 2k^{-1} - k^{-7}$, we have
$$
\tl{\lambda}^{(j^*)}_1 \geq \frac{w_{\max}(1+\mu)}{w_{\min}-\rho w_{\max}(1+\mu)} \left( \tl{\lambda}_{(2)}^{(j^*)} - \rho \tl{\lambda}^{(j^*)}_1  \right).
$$
Simple calculations imply that
$$
w_{\min} \tl{\lambda}_1^{(j^*)} \geq (1+\mu) w_{\max} \tl{\lambda}_{(2)}^{(j^*)}.
$$
Incorporating inequalities $\lambda_1 \geq  w_{\min} \tl{\lambda}_1$ and $\lambda_{(2)} \leq w_{\max} \tl{\lambda}_{(2)}$ finishes the proof saying that the result of lemma is valid for the $j^*$-th draw.
\eprf

\vspace{1em}
In the following lemma, we show that if a vector $\theta \in \R^d$ is relatively more correlated with $c_1$ (comparing to $c_i,i \neq 1$), then dominant singular vectors of $\hT(I,I,\theta)$ provide good initialization vectors for $\astar_1$ and $\bstar_1$. %Then, we can do this with different random realizations of $\theta$ to complete the initialization matrices $\hA^{(0)}$ and $B^{(0)}$.

%When, the empirical tensor $\hT = T + \Psi$ is given, the following analysis needs to be revised. In this case, we have $\hT(I,I,\theta) = T(I,I,\theta) + \Psi(I,I,\theta)$. Therefore, the additional term $\Psi(I,I,\theta)$ shows up in the application of Wedin's perturbation theorem as $\hT(I,I,\theta) = M + E + \Psi(I,I,\theta)$. The effective perturbation bound is $E + \Psi(I,I,\theta)$, and therefore, we should also appropriately bound $\| \Psi(I,I,\theta) \|$ which adds to the numerator of bound provided by Wedin's theorem. We have $\vectorform( \Psi(I,I,\theta) ) = \mat(\Psi,3)^\top \theta$, where $\vectorform(\cdot)$ results the vectorized form of its argument. Hence,
%\begin{align*}
%\| \Psi(I,I,\theta) \| \leq \| \mat(\Psi,3)^\top \theta \| \leq \| \mat(\Psi,3)^\top \| \cdot \| \theta \| \leq \psi \| \theta \| \approx \psi \sqrt{d},
%\end{align*}
%where Assumption \ref{cond:perturbation bound} is exploited. It is more tricky to deal with $\sqrt{d}$ term to get a good bound.

%\begin{lemma}
%Let $u_1$ and $v_1$ be the top left and right singular vectors of $T(I,I,\theta)$. If $\tl{\mu}_\lambda \geq \tl{\mu} + \|E\| + \lambdastar_{(2)} \left(2\sqrt{\frac{k}{d}} + \frac{k}{d} \right)$ for some $\tl{\mu}>0$, then we have
%\begin{align*}
%\max \left\{ \sqrt{1- \langle u_1,\astar_1 \rangle^2}, \sqrt{1- \langle v_1,\bstar_1 \rangle^2} \right\}  \leq \frac{\| E \|}{\tl{\mu}}.
%\end{align*}
%\end{lemma}

Before proposing the lemma, we define
\begin{align} \label{eqn:mu_E & m_1}
\mu_E := \alpha \sqrt{\frac{k}{d}} \left( 2 + 2\alpha_0 \sqrt{\frac{k}{d}} + \frac{\alpha}{\sqrt{d}} \right),
\quad
\mu_R := \left( 1 + \alpha_0 \sqrt{\frac{k}{d}} \right)^2,
\quad
\mu_{\min} := \min\{\mu_E,\mu_R\}.
\end{align}
where $\alpha= \polylog(d)$, and $\alpha_0>0$ is a constant.

\begin{lemma} \label{lem:initialization}
Consider $\hT = T + \Psi$, where $T$ is a rank-$k$ tensor, and $\Psi$ is a perturbation tensor.
Let assumptions \ref{cond:rank-k}-\ref{cond:spectral bound} hold for $T$. Let $u_1$ and $v_1$ be the top left and right singular vectors of $\hT(I,I,\theta)$. Let
$$
\lambdastar := \Diag(\wstar) \Cstar^\top \theta \in \Rbb^k,
$$
denote the vector that captures correlation of $\theta$ with different $c_i, i \in[k]$, weighted by $w_i, i \in [k]$.
Without loss of generality, assume that $\lambdastar_1 = \max_i |\lambdastar_i|$, and let $\lambdastar_{(2)} := \max_{i \neq 1} |\lambdastar_i|$.
Suppose the relative gap condition
\begin{align} \label{eqn:gap condition}
%\frac{\lambdastar_1 - \lambdastar_{(2)}}{\lambdastar_{(2)}} \geq \mu_E+\mu_R+\tl{\mu},
\lambdastar_1 \geq (1+\mu)\lambdastar_{(2)},
\end{align}
is satisfied for some $\mu > \frac{\lambda_1}{\lambda_1 - \| \Psi(I,I,\theta) \|} 2 \mu_R - 1$, where $\mu_R$ and $\mu_{\min}$ are defined in \eqref{eqn:mu_E & m_1}.
%is satisfied for some $\tl{\mu}>\| \Psi(I,I,\theta) \|/\lambda_{(2)}$, and $\mu_E,\mu_R$ defined in \eqref{eqn:mu_E & m_1}.
Then, with high probability (w.h.p.),
$$
\max \{ \dist(u_1,a_1), \dist(v_1,b_1) \}
%\max \left\{ \sqrt{1- \langle u_1,\astar_1 \rangle^2}, \sqrt{1- \langle v_1,\bstar_1 \rangle^2} \right\}
\leq \frac{\mu_{\min} \lambdastar_{(2)} + \| \Psi(I,I,\theta) \|}{\tl{\mu} \lambdastar_1 - \| \Psi(I,I,\theta) \|},
$$
for $\| \Psi(I,I,\theta) \| / \lambda_1 < \tl{\mu}  <1$ defined as
$$
\tl{\mu} := \frac{1+\mu - 2\mu_R}{1+\mu}.
$$
\end{lemma}

\bprf
From Assumption \ref{cond:rank-k}, $T(I,I,\theta)$ can be written as equation \eqref{eqn:T(theta) expansion}, Expanded as
$$
T(I,I,\theta) = \lambdastar_1 \astar_1 \bstar_1^\top + \underbrace{\sum_{i \neq 1} \lambdastar_i \astar_i \bstar_i^\top}_{=: R}.
$$
From here, we prove the result in two cases. First when $\mu_E < \mu_R$ and therefore $\mu_{\min}=\mu_E$, and second when $\mu_E \geq \mu_R$ and therefore $\mu_{\min}=\mu_R$.

\ni {\bf Case 1 ($\mu_E < \mu_R$): }
According to the subspaces spanned by $\astar_1$ and $\bstar_1$, we decompose matrix $R$ to two components as $R = \Pc_{\perp}(R) + \Pc_{\parallel}(R)$. First term $\Pc_{\perp}(R)$ is the component with column space orthogonal to $\astar_1$ and row space orthogonal to $\bstar_1$, and $\Pc_{\parallel}(R)$ is the component with either the column space equal to $\astar_1$ or the row space equal to $\bstar_1$. We have
\begin{align*}
\Pc_{\perp}(R) & = (I-P_{\astar_1})R(I-P_{\bstar_1}), \\
\Pc_{\parallel}(R) & = P_{\astar_1} R + R P_{\bstar_1} - P_{\astar_1} R P_{\bstar_1},
\end{align*}
where $P_{\astar_1} = \astar_1 \astar_1^\top$ is the projection operator on the subspace in $\Rbb^d$ spanned by $\astar_1$, and similarly $P_{\bstar_1} = \bstar_1 \bstar_1^\top$ is the projection operator on the subspace in $\Rbb^d$ spanned by $\bstar_1$. Thus, for $\hT = T + \Psi$, we have
$$
\hT(I,I,\theta) = \underbrace{\lambdastar_1 \astar_1 \bstar_1^\top + \Pc_{\perp}(R)}_{=:M} + \underbrace{\Pc_{\parallel}(R)}_{=:E} + \Psi(I,I,\theta).
$$
Looking at $M$, it becomes more clear why we proposed the above decomposition for $R$. Since the column and row space of $\Pc_{\perp}(R)$ are orthogonal to $\astar_1$ and $\bstar_1$, respectively, the SVD of $M$ has $\astar_1$ and $\bstar_1$ as its left and right singular vectors, respectively. Hence, $M$ has the SVD form
\begin{align*}
M = [\astar_1 \ \tilde{U}_2]
\left[\begin{array}{cc}
\lambdastar_1 & 0 \\
0 & \tilde{\Sigma}_2
\end{array}\right]
[\bstar_1 \ \tilde{V}_2]^\top,
\end{align*}
where $\Pc_{\perp}(R) = \tilde{U}_2 \tilde{\Sigma}_2 \tilde{V}_2^\top$ is the SVD of $\Pc_{\perp}(R)$. Let $\tilde{\sigma}_2 := \max_i (\tilde{\Sigma}_2)_{ii}$. From gap condition \eqref{eqn:gap condition} assumed in the lemma and inequality \eqref{eqn:sigmatilde_2 bound}, we have $\lambdastar_1 \geq \tilde{\sigma}_2$, and therefore, $\astar_1$ and $\bstar_1$ are the top left and right singular vectors of $M$.
On the other hand, $\hT(I,I,\theta)$ has the corresponding SVD form
\begin{align*}
\hT(I,I,\theta) = [u_1 \ U_2]
\left[\begin{array}{cc}
\sigma_1 & 0 \\
0 & \Sigma_2
\end{array}\right]
[v_1 \ V_2]^\top,
\end{align*}
where $u_1$ and $v_1$ are its top left and right singular vectors. We have
\begin{align} \label{eqn:sigmatilde_2 bound}
\tilde{\sigma}_2 =  \| \Pc_{\perp}(R) \| & \leq \| R \|  \nn \\
&= \left\| \sum_{i = 2}^k \lambdastar_i \astar_i \bstar_i^\top \right\|  \nn \\
&\leq \lambdastar_{(2)} \left\|A_{\setminus 1}\right\| \left\|B_{\setminus 1}^\top\right\|  \nn \\
&\leq \lambdastar_{(2)} \left\|A\right\| \left\|B^\top\right\|  \nn \\
&\leq \left( 1+\alpha_0 \sqrt{\frac{k}{d}} \right)^2 \lambdastar_{(2)} =: \mu_R \lambdastar_{(2)},
\end{align}
where the sub-multiplicative property of spectral norm is used in the second inequality, and the last inequality is from Assumption \ref{cond:spectral bound}.
From Weyl's theorem, we have
\begin{align}
| \sigma_1 - \lambdastar_1 |
& \leq \|E\| + \| \Psi(I,I,\theta) \| \nn \\
& \leq \lambdastar_{(2)} \alpha \sqrt{\frac{k}{d}} \left( 2 + 2\alpha_0 \sqrt{\frac{k}{d}} + \frac{\alpha}{\sqrt{d}} \right) + \| \Psi(I,I,\theta) \| \nn \\
& =: \mu_E  \lambdastar_{(2)} + \| \Psi(I,I,\theta) \|, \label{eqn:Weyl's result}
\end{align}
where \eqref{eqn:Initialization error norm bound} is used in the second inequality.
Therefore, we have
\begin{align*}
\sigma_1 - \tilde{\sigma}_2 &= \sigma_1 - \lambdastar_1 + \lambdastar_1 - \tilde{\sigma}_2 \\
&\geq -\mu_E \lambdastar_{(2)}  - \| \Psi(I,I,\theta) \| + \lambdastar_1 - \mu_R \lambdastar_{(2)} \\
& \geq \left( 1-\frac{\mu_E+\mu_R}{1+\mu} \right) \lambdastar_1 - \| \Psi(I,I,\theta) \|, \\
& =: \tl{\mu}_1 \lambdastar_1 - \| \Psi(I,I,\theta) \| =: \nu,
\end{align*}
where bounds \eqref{eqn:sigmatilde_2 bound} and \eqref{eqn:Weyl's result} are used in the first inequality, and the second inequality is concluded from the gap condition \eqref{eqn:gap condition} assumed in the lemma. Therefore, since $\sigma_1 \geq \beta + \nu$ and $\tilde{\sigma}_2 \leq \beta$ for some $\beta>0$, Wedin's theorem is applied to the equality $\hT(I,I,\theta)=M+E+\Psi(I,I,\theta)$,  which implies that
%\begin{align*}
%\max \left\{ | \sin( \angle (u_1,\astar_1)) |, | \sin( \angle (v_1,\bstar_1)) | \right\}  \leq \frac{\| E \|_2}{\nu},
%\end{align*}
%or equivalently
\begin{align*}
\max \left\{ \sqrt{1- \langle u_1,\astar_1 \rangle^2}, \sqrt{1- \langle v_1,\bstar_1 \rangle^2} \right\}
& \leq \frac{\| E + \Psi(I,I,\theta) \|}{\nu} \\
& \leq \frac{\mu_E \lambdastar_{(2)} + \| \Psi(I,I,\theta) \|}{\tl{\mu}_1 \lambdastar_1 - \| \Psi(I,I,\theta) \|} \\
& \leq \frac{\mu_{\min} \lambdastar_{(2)} + \| \Psi(I,I,\theta) \|}{\tl{\mu} \lambdastar_1 - \| \Psi(I,I,\theta) \|},
\end{align*}
where we used $\mu_{\min} = \mu_E$ and $\tl{\mu}_1 > \tl{\mu}$ in the last inequality when $\mu_E < \mu_R$. Since $\dist^2(u_1,a_1) + \langle u_1,a_1 \rangle^2 = 1$, the proof is complete for this case.
%or equivalently, if $\mu_E \leq \tl{\mu}$,
%\begin{align*}
%\min \left\{ |\langle u_1,\astar_1 \rangle|, |\langle v_1,\bstar_1 \rangle| \right\}  \geq \sqrt{1 - \left( \frac{\mu_E}{\tl{\mu}} \right)^2}.
%\end{align*}

\ni  \textbf{Bounding the spectral norm of $E$}:
For any $i \neq j$, let $\rho_{ij}^{(a)} :=| \inner{\astar_i,\astar_j}|$ and $\rho_{ij}^{(b)} :=|\inner{\bstar_i,\bstar_j}|$. We have
\begin{align*}
E := \Pc_{\parallel}(R) &= P_{\astar_1} R + R P_{\bstar_1} - P_{\astar_1} R P_{\bstar_1}, \\
&= \astar_1 \astar_1^\top R + R \bstar_1 \bstar_1^\top - \astar_1 \astar_1^\top R \bstar_1 \bstar_1^\top \\
&= \sum_{i \neq 1} \lambdastar_i \astar_1 \astar_1^\top \astar_i \bstar_i^\top
+ \sum_{i \neq 1} \lambdastar_i \astar_i \bstar_i^\top \bstar_1 \bstar_1^\top
- \sum_{i \neq 1} \lambdastar_i \astar_1 \astar_1^\top \astar_i \bstar_i^\top \bstar_1 \bstar_1^\top \\
&= \sum_{i \neq 1} \lambdastar_i \rho_{1i}^{(a)} \astar_1 \bstar_i^\top
+ \sum_{i \neq 1} \lambdastar_i \rho_{1i}^{(b)} \astar_i \bstar_1^\top
- \sum_{i \neq 1} \lambdastar_i \rho_{1i}^{(a)} \rho_{1i}^{(b)} \astar_1 \bstar_1^\top \\
&= \underbrace{\Astar_{(1)} \Diag(\lambdastar_{(a)}) \Bstar_{\setminus 1}^\top}_{E_1}
+ \underbrace{\Astar_{\setminus 1} \Diag(\lambdastar_{(b)}) \Bstar_{(1)}^\top}_{E_2}
- \underbrace{\Astar_{(1)} \Diag(\lambdastar_{(a,b)}) \Bstar_{(1)}^\top}_{E_3},
\end{align*}
where $\Astar_{(1)} := \Bigl[ \overbrace{\astar_1 | \astar_1 | \dotsb | \astar_1}^{k-1 \ \operatorname{times}} \Bigr] \in \Rbb^{d \times (k-1)}$, $\Bstar_{\setminus 1} := [\bstar_2|\bstar_3|\dotsb|\bstar_k] \in \Rbb^{d \times (k-1)}$, and $\lambdastar_{(a)} := [\lambdastar_i \rho_{1i}^{(a)}]_{i \neq 1} \in \Rbb^{k-1}$. The other notations are similarly defined. \\
For $E_1$, we have
\begin{align*}
\|E_1\| &\leq \| \Astar_{(1)} \Diag(\lambdastar_{(a)}) \| \| \Bstar_{\setminus 1}^\top \| \\
& = \| \lambdastar_{(a)} \| \| \astar_1 \| \| \Bstar_{\setminus 1}^\top \| \\
& \leq \sqrt{k} \lambdastar_{(2)} \rho \| \Bstar^\top \| \\
& \leq \lambdastar_{(2)} \alpha \sqrt{\frac{k}{d}} \left(1+\alpha_0 \sqrt{\frac{k}{d}}\right).
\end{align*}
Where the first equality is concluded from Lemma \ref{lem:spectral norm 1}, and Assumptions \ref{cond:incoherence} and \ref{cond:spectral bound} are exploited in the last inequality. Similarly, for $E_2$ and $E_3$, we have
\begin{align*}
\|E_2\| &\leq \lambdastar_{(2)} \alpha \sqrt{\frac{k}{d}} \left(1+\alpha_0 \sqrt{\frac{k}{d}}\right), \\
\|E_3\| &\leq \lambdastar_{(2)} \alpha^2 \frac{\sqrt{k}}{d}.
\end{align*}
Therefore, we have
\begin{align} \label{eqn:Initialization error norm bound}
\| E \| \leq \lambdastar_{(2)} \alpha \sqrt{\frac{k}{d}} \left( 2 + 2\alpha_0 \sqrt{\frac{k}{d}} + \frac{\alpha}{\sqrt{d}} \right).
\end{align}
%where $\Xc_1 \subset \Rbb^{d \times d}$ is the subspace spanned by $\astar_1$ and $\bstar_1$.

\ni  {\bf Case 2 ($\mu_R \leq \mu_E$): }
The result can be similarly achieved when $\mu_R \leq \mu_E$. Here we directly apply Wedin's theorem to $\hT(I,I,\theta) = \lambda_1 a_1 b_1^\top + R + \Psi(I,I,\theta)$, treating $R + \Psi(I,I,\theta)$ as the error term. From Weyl's theorem, we have
$$
\sigma_1 \geq \lambda_1 - \| R \| - \| \Psi(I,I,\theta) \|
\geq \underbrace{\left( 1 - \frac{\mu_R}{1+\mu} \right)}_{=: \tl{\mu}_2} \lambda_1 - \| \Psi(I,I,\theta) \|,
$$
where \eqref{eqn:sigmatilde_2 bound} and gap condition \eqref{eqn:gap condition} are used in the second inequality. Since $\tl{\sigma}_2 = 0$, by Wedin's theorem, we have
\begin{align*}
\max \left\{ \sqrt{1- \langle u_1,\astar_1 \rangle^2}, \sqrt{1- \langle v_1,\bstar_1 \rangle^2} \right\}
& \leq \frac{\mu_R \lambda_{(2)} + \| \Psi(I,I,\theta) \|}{\tl{\mu}_2 \lambda_1 - \| \Psi(I,I,\theta) \|} \\
& \leq \frac{\mu_{\min} \lambdastar_{(2)} + \| \Psi(I,I,\theta) \|}{\tl{\mu} \lambdastar_1 - \| \Psi(I,I,\theta) \|},
\end{align*}
where we used $\mu_{\min} = \mu_R$ and $\tl{\mu}_2 \geq \tl{\mu}$ in the last inequality when $\mu_R \leq \mu_E$. Since $\dist^2(u_1,a_1) + \langle u_1,a_1 \rangle^2 = 1$, the proof is complete for this case.
\eprf

The above lemma concludes the proof for initialization procedure, except for a few auxiliary lemmata that we prove next.

First we use Gaussian tail bounds to prove that the largest entry of a Gaussian vector can be quite large with inverse polynomial probability:

%\section{Auxiliary Lemmata}

\begin{lemma} \label{lem:Gaussian tail bound}
Let $x \sim \mathcal{N}(0,\sigma)$ be a Gaussian random variable with mean zero and variance $\sigma^2$. Then, for any $t>0$, we have
$$
\left( \frac{\sigma}{t} - \frac{\sigma^3}{t^3} \right) f(t/\sigma) \leq \Pr[x \geq t] \leq \frac{\sigma}{t} f(t/\sigma),
$$
where $f(t) = \frac{1}{\sqrt{2\pi}} e^{-t^2/2}$.
\end{lemma}
\bprf
Let $z = \frac{x}{\sigma}$, where $z\sim \mathcal{N}(0,1)$ is a standard Gaussian random variable. Then, we have $\Pr [x \geq t] = \Pr [z \geq t/\sigma]$, and therefore, the result is proved by using standard tail bounds for Gaussian random variable.
\eprf

\begin{lemma} \label{lem:Gaussian max bound}
Consider $r=[r_1,r_2,\dotsc,r_k]^\top \in \Rbb^k$ as a $k$-dimensional random Gaussian vector with zero mean and covariance $\Sigma$, i.e., $r \sim \mathcal{N}(0,\Sigma)$. For any $k \geq 2$, we have
$$
\Pr \left[ r_{(1)} \geq 4 \sigma_{\max} \sqrt{\log k} \right] \leq k^{-7}.
$$
\end{lemma}
\bprf
From Lemma \ref{lem:Gaussian tail bound}, for any $i \in [k]$, we have
$$
\Pr \left[ |r_i| \geq 4 \sigma_{\max} \sqrt{\log k} \right] \leq \frac{1}{2\sqrt{2 \pi \log k}} k^{-8} \leq k^{-8},
$$
where the last inequality is concluded from the fact that $k \geq 2$. The result is then proved by taking a union bound.
\eprf

Next we prove a basic fact about spectral norm that is used in the proof of Lemma~\ref{lem:initialization}.

\begin{lemma} \label{lem:spectral norm 1}
Given $h \in \Rbb^m$ and $v \in \Rbb^n$, let $H = [h| h| \dotsb | h] \Diag(v) \in \Rbb^{m \times n}$. Then, $\|H\| = \|h\| \|v\|$.
\end{lemma}
\bprf
By definition
$$
\|H\| = \sup_{\|x\|=1} \|Hx\|.
$$
We have $Hx = \langle v,x \rangle h$, and therefore, $\|Hx\| = |\langle v,x \rangle| \|h\|$. This is maximized by $x=v/\|v\|$, and this finishes the proof.
\eprf

Finally, we show that noise matrix $\Psi(I, I, \theta)$ has bounded norm with high probability which is useful for initialization argument in Theorem~\ref{thm:initialization}.

\begin{lemma} \label{lem:Noise tensor norm}
Let $\theta \in \R^d$ be standard multivariate Gaussian as $\mathcal{N}(0,I_d)$. Then, for any $\alpha_0>1$, we have
$$
\Pr \left[ \| \Psi(I, I, \theta) \| \leq \alpha_0 \sqrt{d} \psi \right] \geq 1-e^{-(\alpha_0-1)^2d/2},
$$
where $\psi := \| \Psi \|$ is the spectral norm of error tensor $\Psi$.
\end{lemma}

\bprf
Let $\theta_n := \frac{1}{\|\theta\|} \theta$ denote the normalized version of $\theta$. Then, we have
$$
\| \Psi(I, I, \theta) \| = \|\theta\| \cdot \| \Psi(I, I, \theta_n) \| \leq \|\theta\| \psi,
$$
where the last inequality is from the definition of tensor spectral norm.
Applying the bound on $\|\theta\|$ in Lemma~\ref{lem:ChiSquareTail} finishes the proof.
\eprf

The following lemma provides concentration bound for the norm of standard Gaussian vector which is basically a tail bound for the chi-squared random variable.

\begin{lemma}[Lemma 15 of \citet{ConcentrationProjections}] \label{lem:ChiSquareTail}
Let the random vector $\theta$ is distributed as $\mathcal{N}(0,I_d)$. Then, for any $\alpha_0>1$, we have
$$
\Pr \left[ \|\theta\| \geq \alpha_0 \sqrt{d} \right] \leq e^{-(\alpha_0-1)^2d/2}.
$$
\end{lemma}

%\begin{lemma} \label{lem:Noise tensor norm}
%Let $\theta \in \R^d$ be standard multivariate Gaussian. Then, with probability at least $1-\delta$, we have
%\begin{align*}
%\| \Psi(I, I, \theta) \| \leq C_1 \|\Psi\|_F \sqrt{\log \frac{1}{\delta}},
%\end{align*}
%for some constant $C_1>0$.
%\end{lemma}
%\bprf
%We have
%\begin{align*}
%\| \Psi(I, I, \theta) \| \leq \| \Psi(I, I, \theta) \|_F = \| \mat(\Psi,3)^\top \theta \|,
%\end{align*}
%where the equality is concluded from the fact that $\vectorform \left( \Psi(I,I,\theta)^\top \right) = \mat(\Psi,3)^\top \theta$, where $\vectorform(\cdot)$ results the vectorized form of its argument. Then, by applying Lemma \ref{lem:Gaussian mixture upper bound}, the result can be proved. Let $M := \mat(\Psi,3)^\top$, and $\Sigma = M^\top M$. Then,
%\begin{align*}
%& \tr(\Sigma) = \|M\|_F^2 = \|\Psi\|_F^2, \\
%& \tr(\Sigma^2) \le \tr(\Sigma)^2 = \|\Psi\|_F^4, \\
%& \| \Sigma \| = \| M^\top M \| \leq \|M\|^2 \leq \| \Psi \|_F^2.
%\end{align*}
%Finally, the result is proved by applying Lemma \ref{lem:Gaussian mixture upper bound}.
%\eprf
%
%For Gaussian random vectors, we know the following fact.
%\begin{lemma} \label{lem:Gaussian mixture upper bound}
%If $x \in \R^d$ is standard multivariate Gaussian, $M$ be an arbitrary matrix, and $\Sigma = M^\top M$, then
%$$
%\Pr[\|Ax\|_2^2 > \tr(\Sigma)+2\sqrt{\tr{\Sigma^2}t}+2\|\Sigma\|t] \le e^{-t}.
%$$
%\end{lemma}
%
%%\textcolor{red}{The bound is usually not tight, because as long as $\Sigma$ is not rank 1, $\tr(\Sigma^2) \ll \tr(\Sigma)^2$, and $\|\Sigma\| \le \Sigma^2$.}

\section{Clustering Process} \label{sec:clustering}
In the last step of main algorithm, we need to cluster the generated 4-tuples into $k$ clusters. Theoretically, we only have convergence guarantees when the initialization vectors are good enough, while the other initializations can potentially generate arbitrary 4-tuples. In the worst case, these arbitrary 4-tuples can make the clustering process hard, and therefore, we provide specific Procedure~\ref{alg:cluster} for which the output properties are provided in Lemma~\ref{lem:clustering}.

Note that the key observation for the algorithm is if $T(\ha,\hb,\hc)$ is large for some $(\ha,\hb,\hc)$, then these vectors are close to $(a_i, b_i, c_i)$ for some $i \in [k]$.

For simplicity, we only prove this when the initialization procedure in Theorem~\ref{thm:global convergence} takes polynomial time, namely $k = O(d)$ and $w_{\max}/w_{\min} = O(1)$. Without loss of generality, we also assume $w_{\max} = w_1 \ge w_2\ge\cdots \ge w_k = w_{\min}$. In this case, we choose the threshold $\epsilon$ in the following lemmata to be some small constant
%\mjcomment{Is it a constant or should be e.g. proportional with $1/\sqrt{d}$?!}
%\rgcomment{No, it is a constant that depends the ratios $k/d$ and $w_{max}/w_{min}$. We know (from the convergence proof) that we have vectors that are really close to the columns. However, it is not clear whether these vectors will always have tensor value $T(u,v,w)$ much larger than those other vectors who are only constant close. In order to get a significant gap we use $T(u,v,w)$ to distinguish between the case when $u,v,w$ are constant close, and the case when they are farther away.}
depending on $k/d$ and $w_{\max}/w_{\min}$. Also, we work in the case when noise $\Psi = 0$, however the proof still works when the noise $\psi = \|\Psi\| = o(1)$.

%We also need a new Assumption
%
%Assumption (??), the $2->p$ norm of $A$, $B$, $C$ are bounded by $1+o(1)$ for some constant $2.5 < p < 3$.
%
%This is satisfied by random matrices (again by ~\cite{guedon2007lp, adamczak2011chevet})

\begin{lemma}
\label{lem:largewisgood}
Suppose
\[\max\{|\langle a_i,\ha\rangle|,|\langle b_i,\hb\rangle|,|\langle c_i,\hc\rangle|\} \le \epsilon, \quad \forall i \in [t-1],\]
for some $t \in [k]$. Let $\delta := O \left( \frac{w_{\max}}{w_{\min}} \epsilon^{3-p} \right)$, and assume $|T(\ha,\hb,\hc)| \ge (1-\delta) w_t$. Then, there exists some $j$ such that
\[\max\{\dist(\ha, a_j),\dist(\hb, b_j),\dist(\hc, c_j)\} < \frac{w_{\min}}{10w_{\max}}.\]
\end{lemma}

\bprf
Partition tensor $T = \sum_{i \in [k]} w_i a_i \otimes b_i \otimes c_i$ to $T_1+T_2$, where $T_1$ contains all the terms indexed from $1$ to $t-1$, and $T_2$ contains the remaining terms.
From Corollary~\ref{cor:holder}, we have
\[|T_1(\ha,\hb,\hc)| \le w_{\max} \left\| A_{[t-1]}^\top \ha \right\|_3 \cdot \left\|B_{[t-1]}^\top \hb\right\|_3 \cdot \left\| C_{[t-1]}^\top \hc \right\|_3,\]
where $A_{[t-1]} \in \Rbb^{d \times (t-1)}$ denotes the first $t-1$ columns of $A$, and similarly for $B_{[t-1]}$ and $C_{[t-1]}$.
We also have
\[\left\| A_{[t-1]}^\top \ha \right\|_3^3 \le \left\|A_{[t-1]}^\top \ha \right\|_p^p \cdot \max_{i \in [t-1]} |\langle a_i, \ha \rangle|^{3-p} = O \left(\epsilon^{3-p} \right),\]
where Assumption \ref{cond:2-to-pnorm} and the assumption in the lemma are exploited in the last step. Similar arguments hold for $b$ and $c$.
Combining with the earliest inequality, we have
\[|T_1(\ha,\hb,\hc)| \le w_{\max} O \left( \epsilon^{3-p} \right) \le w_t\delta,\]
where the definition of $\delta$ is exploited in the last inequality. Applying assumption $|T(\ha,\hb,\hc)| \ge (1-\delta) w_t$ to the above bound, we have
\begin{align} \label{eqn:T2LowerBound}
|T_2(\ha,\hb,\hc)| \ge (1-2\delta)w_t.
\end{align}

On the other hand, from Corollary~\ref{cor:holder},
$$|T_2(\ha,\hb,\hc)|\le w_t \|A^\top \ha\|_3\|B^\top \hb\|_3\|C^\top \hc\|_3.$$
%\mjcomment{Not clear to me from here.}
%\rgcomment{Can you be more specific? The above inequality is just Corollary 3, and the 3-norms are upperbounded by the $2\to 3$ norms of matrices $A$, $B$, $C$ which we assumed (we actually assumed even stronger $2\to p$ norm for $p < 3$). If one of them is smaller than say $1-100\delta$, it is impossible to get the product to be larger than $(1-2\delta)w_t$}
Since all the 3-norms are bounded by $1+o(1)$, each of them must be at least $1-O(\delta)$ to let inequality~\eqref{eqn:T2LowerBound} hold.
Now we have
$$1-O(\delta) \le \sum_{j=1}^k |\langle a_j, \ha\rangle|^3 \le \max\{|\langle a_j, \ha\rangle|\}^{3-p} \sum_{t=1}^k |\langle a_j, \ha\rangle|^p \le (1+o(1))\max\{|\langle a_j, \ha\rangle|\}^{3-p},$$
where the last inequality is from Assumption~\ref{cond:2-to-pnorm}.
This implies $\max\{|\langle a_j, \ha\rangle|\} = 1-O(\delta)$, which in turn implies there exists a $j$ such that
$$\dist(\ha, a_j) < w_{\min}/10w_{\max}$$
when $\epsilon$ and $\delta$ are small enough.

%\rgcomment{The inequality above should be clear, right? It uses the fact that $A$ has $2\to p$ norm $1+o(1)$}

By symmetry we know there is also a $j'$ such that $\dist(\hb, b_{j'}) < w_{\min}/10w_{\max}$.
If $j \ne j'$, then it is easy to check $T_2(\ha,\hb,\hc)$ cannot be large. Hence, $j= j'$ and the Lemma is correct.
\eprf

On the other hand, we know if there is a good initialization, the largest $T(\ha,\hb,\hc)$ must be large.

\begin{lemma}
\label{lem:existlargew}
Suppose there exists a good initialization (see initialization condition \eqref{eqn:good init} in the local convergence theorem) for some column $t \in [k]$, and
\[\max\{|\langle a_i,\ha^{(0)}\rangle|,|\langle b_i,\hb^{(0)}\rangle|,|\langle c_i,\hc^{(0)}\rangle|\} \le \epsilon, \quad \forall i \neq t.\]
Let $\delta := O \left( \frac{w_{\max}}{w_{\min}} \epsilon^{3-p} \right)$.
Then the corresponding output of iterations in Algorithm \ref{algo:Power method form} denoted by $(\ha,\hb,\hc)$ satisfy
\[|T(\ha,\hb,\hc)| > (1-\delta)w_t.\]
Furthermore, for any $i \ne t$, $\max\{|\langle \ha, a_i \rangle|,|\langle \hb, b_i \rangle|,|\langle \hc, c_i \rangle|\} \leq o(\epsilon)$.
\end{lemma}

\bprf
Similar to the proof of Lemma \ref{lem:largewisgood}, partition tensor $T  = \sum_{i \in [k]} w_i a_i \otimes b_i \otimes c_i$ to $T_2 = w_t a_t\otimes b_t\otimes c_t$ and $T_1 = T-T_2$. Since the initialization is good, by the local convergence result in Theorem \ref{thm:local convergence}, we have
\[\dist(\ha, a_t) \leq \tilde{O} \left( \frac{w_{\max}}{w_{\min}} \frac{\sqrt{k}}{d} \right) \leq o(\delta),\]
where the incoherence condition and $p > 2$ are exploited in the last step.
%\mjcomment{This is the reason, right?}
%\rgcomment{I don't think we used incoherence property and $p>2$ here. This step is just saying, if we had good initialization then we converge to something that's close to columns of $A$. This is exactly what we proved in Theorem 4. We are not using anything else.}
Therefore, $|T_2(\ha,\hb,\hc)| \ge (1-\delta/2) w_t$.

Similar to Lemma \ref{lem:largewisgood}, by using Corollary~\ref{cor:holder}, we have $|T_1(\ha,\hb,\hc)| \le w_t\delta/2$. Applying these bounds, we have
\[|T(\ha,\hb,\hc)| \ge |T_2(\ha,\hb,\hc)| - |T_1(\ha,\hb,\hc)| \ge (1-\delta)w_t.\]

The last part of the Lemma is trivial because $\dist(\ha, a_t)$ is small and $\langle a_i, a_t\rangle$ is small by incoherence.
\eprf

Finally we prove the clustering process succeeds.

\begin{lemma} \label{lem:clustering}
%There are at least one successful initialization for every component (which happens w.h.p.).
Procedure~\ref{alg:cluster} outputs $k$ cluster centers that are $\tilde{O} \left(\frac{w_{\max}}{w_{\min}} \frac{\sqrt{k}}{d} \right)$ close to the true components of the tensor.
\end{lemma}

\bprf
We prove by induction to show that every step of the algorithm correctly computes one component.

Suppose all previously found 4-tuples are $\tilde{O}(w_{\max}\sqrt{k}/w_{\min}d)$ close to some $(a_i,b_i,c_i)$ (notice that this is true at the beginning when no components are found). Let $t$ be the smallest index that has not been found. Then all the remaining 4-tuples satisfy 
$$\max\{|\langle a_i,\ha\rangle|,|\langle b_i,\hb\rangle|,|\langle c_i,\hc\rangle|\} \le \epsilon, \quad \forall i < t.$$ By Lemma~\ref{lem:existlargew} we know there must be a 4-tuple with $|T(\ha,\hb,\hc)| > w_t(1-\delta)$. On the other hand, by Lemma~\ref{lem:largewisgood} we know the 4-tuple we found must satisfy $\max\{\dist(\ha, a_j),\dist(\hb, b_j)\} < w_{\min}/10w_{\max}$ for some $j$ (and this cannot be some $j$ that has already been found). This tuple then satisfies the conditions of the local convergence Theorem~\ref{thm:local convergence}. Hence, after $N$ iterations it must have converged to $(a_j,b_j,c_j)$. At this step the algorithm successfully found a new component of the tensor.
%\mjcomment{Why do we need to do $N$ more iterations?}
%\rgcomment{As I explained before, it is not easy to tell a really close vector from a constant-close vector just by looking at $T(u,v,w)$. The procedures only allow us to find vectors that are constant-close, so they are good initial points. From these initial points we use more iterations to make sure they converge.
%Please refer to the old tensor power method paper for this part (there we needed to do more iterations for the same reason).}

\eprf

\end{document}